\documentclass[11pt, letterpaper]{article}

\usepackage[left=1in, right=1in, top=1in, bottom=1in]{geometry}
\usepackage[utf8]{inputenc}
\usepackage[T1]{fontenc}
\usepackage{bm}
\usepackage{type1cm}
\usepackage{lettrine}
\usepackage{amsmath,amssymb,amsthm}
\usepackage{moreverb}
\usepackage{siunitx}
\usepackage{mathtools}
\usepackage{amsmath}
\usepackage{amssymb}
\usepackage{algorithmic}
\usepackage{graphics}
\usepackage{graphicx}
\usepackage{subfigure}
\usepackage{caption}
\usepackage{extarrows}
\usepackage{color}
\usepackage{framed}
\usepackage{wrapfig}
\usepackage{bm}
\usepackage{mathrsfs}
\usepackage{mathabx}
\usepackage{multirow}
\usepackage{longtable}
\usepackage{hyperref}
\usepackage{paralist}
\usepackage{indentfirst}
\usepackage{relsize}
\usepackage{extarrows}
\usepackage{upgreek}
\usepackage{bm}
\usepackage{mwe}
\usepackage[dvipsnames,table]{xcolor}
\usepackage{booktabs}
\usepackage{authblk}
\usepackage{lettrine}
\usepackage{type1cm}
\usepackage{threeparttable}
\usepackage[sort&compress,numbers]{natbib}
\usepackage[figurename=Figure]{caption}
\usepackage[normalem]{ulem}

\graphicspath{ {./Figure/} }
\usepackage[font=footnotesize,labelfont=bf]{caption}

\providecommand{\keywords}[1]{\textbf{\textit{Keywords: }} #1}

\newcommand{\EQ}{\begin{equation}}
\newcommand{\EN}{\end{equation}}
\newcommand{\EQA}{\begin{eqnarray}}
\newcommand{\ENA}{\end{eqnarray}}

\hypersetup{
bookmarks=true,
bookmarksopen=true,
bookmarksnumbered=true,
unicode=false,
pdftoolbar=true,
pdfmenubar=true,
pdffitwindow=false,
pdfstartview={FitH},
pdftitle={My title},
pdfauthor={Author},
pdfsubject={Subject},
pdfcreator={Creator},
pdfproducer={Producer},
pdfkeywords={keywords},
pdfnewwindow=true,
colorlinks=true,
linkcolor=blue,
citecolor=blue,
filecolor=blue,
urlcolor=blue
}

\begin{document}

\title{\textbf{Benchmarking neural surrogates on realistic spatiotemporal multiphysics flows}}

\author[1,$^\dag$]{Runze Mao}
\author[2,$^\dag$]{Rui Zhang}
\author[3]{Xuan Bai}
\author[3]{Tianhao Wu}
\author[3]{Teng Zhang}
\author[1]{Zhenyi Chen}
\author[1]{\\Minqi Lin}
\author[2]{Bocheng Zeng}
\author[1]{Yangchen Xu}
\author[1]{Yingxuan Xiang}
\author[1]{Haoze Zhang}
\author[4]{\\Shubham Goswami}
\author[4]{Pierre A. Dawe}
\author[1]{Yifan Xu}
\author[5]{Zhenhua An}
\author[2]{Mengtao Yan}
\author[6]{\\Xiaoyi Lu}
\author[6]{Yi Wang}
\author[7]{Rongbo Bai}
\author[8]{Haobu Gao}
\author[4]{Xiaohang Fang}
\author[1,3]{Han Li}
\author[2,*]{\\Hao Sun}
\author[1,3,*]{Zhi X. Chen}

\affil[1]{\small State Key Laboratory for Turbulence and Complex Systems, School of Mechanics and Engineering Science, Peking University, Beijing, China}
\affil[2]{\small Gaoling School of Artificial Intelligence, Renmin University of China, Beijing, China}
\affil[3]{\small AI for Science Institute, Beijing, China}
\affil[4]{\small University of Calgary, Calgary, Canada}
\affil[5]{\small Kyoto University, Kyoto, Japan}
\affil[6]{\small FM Global, Boston, United States}
\affil[7]{\small LandSpace Technology Corporation Ltd., Beijing, China}
\affil[8]{\small Aero Engine Academy of China, Beijing, China \vspace{12pt}}

\affil[$\dag$]{Equal contribution}
\affil[*]{Corresponding authors}

\date{}

\maketitle

\vspace{-26pt} 
\begin{abstract}
\small
Predicting multiphysics dynamics is computationally expensive and challenging due to the severe coupling of multi-scale, heterogeneous physical processes. While neural surrogates promise a paradigm shift, the field currently suffers from an ``illusion of mastery'', as repeatedly emphasized in top-tier commentaries~\cite{ref1}
: existing evaluations overly rely on simplified, low-dimensional proxies, which fail to expose the models' inherent fragility in realistic regimes. To bridge this critical gap, we present \textbf{REALM} (\textbf{RE}alistic \textbf{A}I \textbf{L}earning for \textbf{M}ultiphysics), a rigorous benchmarking framework designed to test neural surrogates on challenging, application-driven reactive flows. REALM features 11 high-fidelity datasets spanning from canonical multiphysics problems to complex propulsion and fire safety scenarios, alongside a standardized end-to-end training and evaluation protocol that incorporates multiphysics-aware preprocessing and a robust rollout strategy. Using this framework, we systematically benchmark over a dozen representative surrogate model families, including spectral operators, convolutional models, Transformers, pointwise operators, and graph/mesh networks, and identify three robust trends: (i) a scaling barrier governed jointly by dimensionality, stiffness, and mesh irregularity, leading to rapidly growing rollout errors; (ii) performance primarily controlled by architectural inductive biases rather than parameter count; and (iii) a persistent gap between nominal accuracy metrics and physically trustworthy behavior, where models with high correlations still miss key transient structures and integral quantities. Taken together, REALM exposes the limits of current neural surrogates on realistic multiphysics flows and offers a rigorous testbed to drive the development of next-generation physics-aware architectures.
\end{abstract}


\vspace{12pt} 

\section*{Introduction}

Multiphysics flows, in which fluid mass, momentum, and energy transport are coupled with additional physics such as chemical reactions~\cite{kim2024learning}, radiation~\cite{kuranz2018high}, or electromagnetism~\cite{chatterjee2017magnetic}, are ubiquitous in science and technology, spanning scales from microfluidic semiconductors to combustion engines~\cite{borghi1988turbulent} and astrophysical plasmas~\cite{zhou2004colloquium}. Accurately predicting spatiotemporal dynamics in these systems is essential for scientific discovery and engineering applications. Yet despite many decades of computational efforts, such prediction remains extremely challenging due to the strong coupling, inherent stiffness, and high dimensionality~\cite{keyes2013multiphysics,poinsot2005theoretical}.

A particularly challenging class of problems arises in reactive multiphysics flow systems, such as biochemical transport in blood flow, reacting flows in energy-conversion and propulsion devices, fire and explosion hazards, and magnetized astrophysical plasma flows with nuclear reactions. Such systems typically involve the tight coupling of fluid dynamics and chemical kinetics, mathematically governed by nonlinear partial differential equations (PDEs) and stiff ordinary differential equations (ODEs)~\cite{poinsot2005theoretical,oran1987numerical}. These processes are inherently multiscale and high-dimensional: many-species chemical ODEs can evolve on microsecond to nanosecond time scales, several orders of magnitude faster than the surrounding fluid dynamics~\cite{pope2013small,lu2009toward}. This stiffness-induced disparity imposes severe constraints on traditional computational fluid dynamics (CFD), where time integration becomes prohibitively expensive due to the very small stable time steps required~\cite{gear1971numerical,strang1968construction}. Moreover, realistic chemical mechanisms (reaction network models) can involve hundreds of species, reactions, and additional couplings, further increasing the effective dimensionality and making high-fidelity simulation of even modest 3D domains extremely costly~\cite{chen2011petascale}.

Over the past decade, rapid progress in machine learning (ML) has motivated extensive efforts to build purely data-driven surrogates for complex multiphysics systems. Early work largely followed the reduced-order modeling (ROM) paradigm~\cite{rowley2017model}, where high-dimensional fields are first projected onto a low-dimensional subspace using statistical techniques such as proper orthogonal decomposition (POD)~\cite{sirovich1987turbulence} or dynamic mode decomposition (DMD)~\cite{schmid2010dynamic}, and the resulting modal coefficients are advanced in time using simple regression models. More recent studies have replaced these shallow regressors with nonlinear sequence models, including long short-term memory (LSTM) networks~\cite{hochreiter1997long,shi2015convolutional}, convolutional or recurrent autoencoders~\cite{usandivaras2024deep} that directly learn the temporal evolution of latent or physical variables~\cite{geneva2022transformers}. Beyond discrete vector representations, neural operator frameworks~\cite{azizzadenesheli2024neural}, such as DeepONet~\cite{lu2021learning,kontolati2024learning,kopanivcakova2025deeponet} and the Fourier neural operator (FNO)~\cite{li2021fourier,tran2023factorized}, aim to learn mappings between infinite-dimensional function spaces, for example from initial fields to full spatiotemporal solutions, thereby bypassing explicit time marching once the model is trained. Although these data-driven approaches have shown promise in simplified settings, they often struggle with robustness, extrapolation, and generalization when confronted with strongly multiscale and high-dimensional dynamics.

To overcome the limitations of purely data-driven methods, physics-aware ML approaches have emerged recently for modeling spatiotemporal systems. A prominent example is physics-informed neural networks (PINNs)~\cite{raissi2019physics,raissi2020hidden}, which use neural networks to map spatiotemporal coordinates to PDE solutions while embedding the governing equations as soft constraints within the loss function. This mesh-free formulation has demonstrated impressive results on selected inverse problems and high-dimensional settings. However, training PINNs for forward simulation often suffers from optimization difficulties, sensitivity to loss balancing and domain decomposition, and substantial retraining costs when the geometry, parameters, or boundary conditions change~\cite{krishnapriyan2021characterizing,grossmann2024can}. Beyond soft constraint, physics-encoded approaches hard-wire numerical structure into the network, such as coupling learnable or non-learnable numerical solvers within the model~\cite{long2019pde,rao2023encoding,wang2025multipdenet,yan2025learnable,franco2023mesh}, or adding explicit correction layers to enforce conservation laws~\cite{richter2022neural,wang2025enforcing}. These physics-encoded methods exhibit strong generalization and robustness, while providing only limited acceleration and are restricted to relatively simple settings, because they usually retain an underlying traditional numerical solver within the architecture.

Despite their promising potential, the intrinsic capability and generalizability of these neural surrogates remain uncertain, as most evaluations are still restricted to simplified, low-dimensional settings. Recent community analyses have warned of overoptimism in the field, highlighting that many reported speedups rely on comparisons against weak numerical baselines rather than optimized solvers~\cite{ref1,brandstetter2025envisioning,mcgreivy2024weak}. This issue is compounded by a lack of rigorous testbeds: while the field abounds with \textit{machine learning solutions looking for PDE problems}, it lacks \textit{CASP-like\footnote{The Critical Assessment of protein Structure Prediction (CASP) is a long-running blind community challenge in protein-structure prediction that has acted as a \textit{de facto} benchmark in structural biology~\cite{moult1999critical}, similar to ImageNet~\cite{deng2009imagenet} in computer vision and ERA5~\cite{hersbach2020era5} reanalysis in weather and climate modeling.} complex and challenging benchmark problems} to validate them. Current benchmark efforts, while valuable, still have notable limitations. Datasets like PDEBench~\cite{takamoto2022pdebench} and The Well~\cite{ohana2024well} have advanced standardized testing, but are largely composed of canonical settings with simplified physics and modest dimensionality. 
Similarly, industrial-scale datasets like DrivAerNet~\cite{elrefaie2024drivaernet++} focus on realistic aerodynamics, but are predominantly steady-state and lack the complexity of multiphysics dynamics, while more recent efforts such as RealPDEBench~\cite{hu2026realpdebench} move toward complex physical systems with real-world data but are not specifically designed around the realistic engine configurations we target here.
Consequently, a critical gap remains: the community lacks benchmarks that are both (1) \textit{grounded in real-world applications} and (2) \textit{computationally complex enough that traditional solvers struggle}. Evaluating neural surrogates on such realistic and complex benchmarks is essential for demonstrating the models’ advantages and practical utility on real-world scientific and engineering tasks.

To bridge this gap, we present REALM (\textbf{RE}alistic \textbf{AI} \textbf{L}earning for \textbf{M}ultiphysics), a benchmarking framework that casts real-world multiphysics flows governed by coupled PDE--ODE systems into explicit, reproducible challenges for neural surrogates. Rather than relying on idealized problems, REALM is built around a suite of application-driven reactive multiphysics scenarios, including flame ignition in shear-driven turbulence, building fire hazards, cryogenic rocket combustors, and shock-driven detonations, which span over two dimensional (2D) and three dimensional (3D) domains, regular and irregular meshes, and chemical mechanisms with varying stiffness. Each trajectory in REALM is generated by high-fidelity CFD or direct numerical simulation and typically requires hundreds to thousands of CPU/GPU hours on large-scale HPC clusters, placing the benchmark squarely in the regime where acceleration is practically valuable. Beyond the dataset itself, REALM provides an end-to-end evaluation protocol with standardized preprocessing and unified normalization, clearly specified forecasting and rollout tasks, and capacity-aligned model configurations that enable fair and transparent comparisons across architectures. Using REALM, we systematically evaluate a set of representative neural surrogates, providing a comprehensive comparative assessment of their capabilities and failure modes, and an extensive study on realistic multiphysics flows rather than on simplified fluid benchmarks. As such, REALM is not merely another dataset, but a rigorous, application-driven testbed for probing the limits of current neural surrogates and guiding the development of next-generation machine-learning models for multiphysics flow simulation.

\section*{Results}
\subsection*{Mathematical Formulation}

We study multiphysics flows in which transport and chemistry are tightly coupled. The dynamics are governed by a conservation-law system with transport and a stiff reactive source~\cite{poinsot2005theoretical},
\begin{equation}
  \frac{\partial \mathbf{q}}{\partial t}
  + \boldsymbol{\nabla}\!\cdot \mathcal{F}(\mathbf{q})
  - \boldsymbol{\nabla}\!\cdot \mathcal{D}(\mathbf{q},\,\boldsymbol{\nabla} \mathbf{q})
  + \mathcal{S}(\mathbf{q}) \;=\; 0,
\label{eq:ns}
\end{equation}
where
\begin{equation}
\mathbf{q}=\big[\,\rho,\;\rho\mathbf{u},\;\rho e,\;\rho Y_1,\ldots,\rho Y_{N_s}\,\big]^\top
\end{equation}
collects the conserved variables: $\rho$ is density, $\mathbf{u}\!\in\!\mathbb{R}^d$ velocity, $e$ total specific energy, and $Y_k$ species mass fractions with $\sum_{k=1}^{N_s}Y_k=1$. Here $\mathcal{F}$ and $\mathcal{D}$ are the convective and diffusive fluxes, and $\mathcal{S}(\mathbf{q})$ is the chemical source. In $d$ spatial dimensions, we denote the gradient operator by
\begin{equation}
  \boldsymbol{\nabla}
  = \bigg[\,\frac{\partial}{\partial x_1},\,\ldots,\,\frac{\partial}{\partial x_d}\,\bigg]^\top,
\end{equation}
and $\boldsymbol{\nabla}\!\cdot$ is the associated divergence.
The latter acts in the species block and follows stiff kinetics,
\begin{equation}
  \mathcal{S}(\mathbf{q})=\frac{d(\rho \mathbf{Y})}{dt}=\dot{\omega}(T,\mathbf{Y})=\sum_{r=1}^{N_r}\nu_{r}\,R_{r}(T,\mathbf{Y}),
\label{eq:chem}
\end{equation}
with temperature $T$, species vector $\mathbf{Y}$, stoichiometric coefficients $\nu_r$, and elementary rates $R_r$. These elementary rates are governed by the specific chemical reaction mechanism and are computed via the Arrhenius law, which models the kinetics as a highly nonlinear function of temperature $T$ and species $\mathbf{Y}$ involving products of polynomial and exponential terms~\cite{law2010combustion}.

\begin{figure}[t!]
  \centering
   \includegraphics[width=\linewidth]{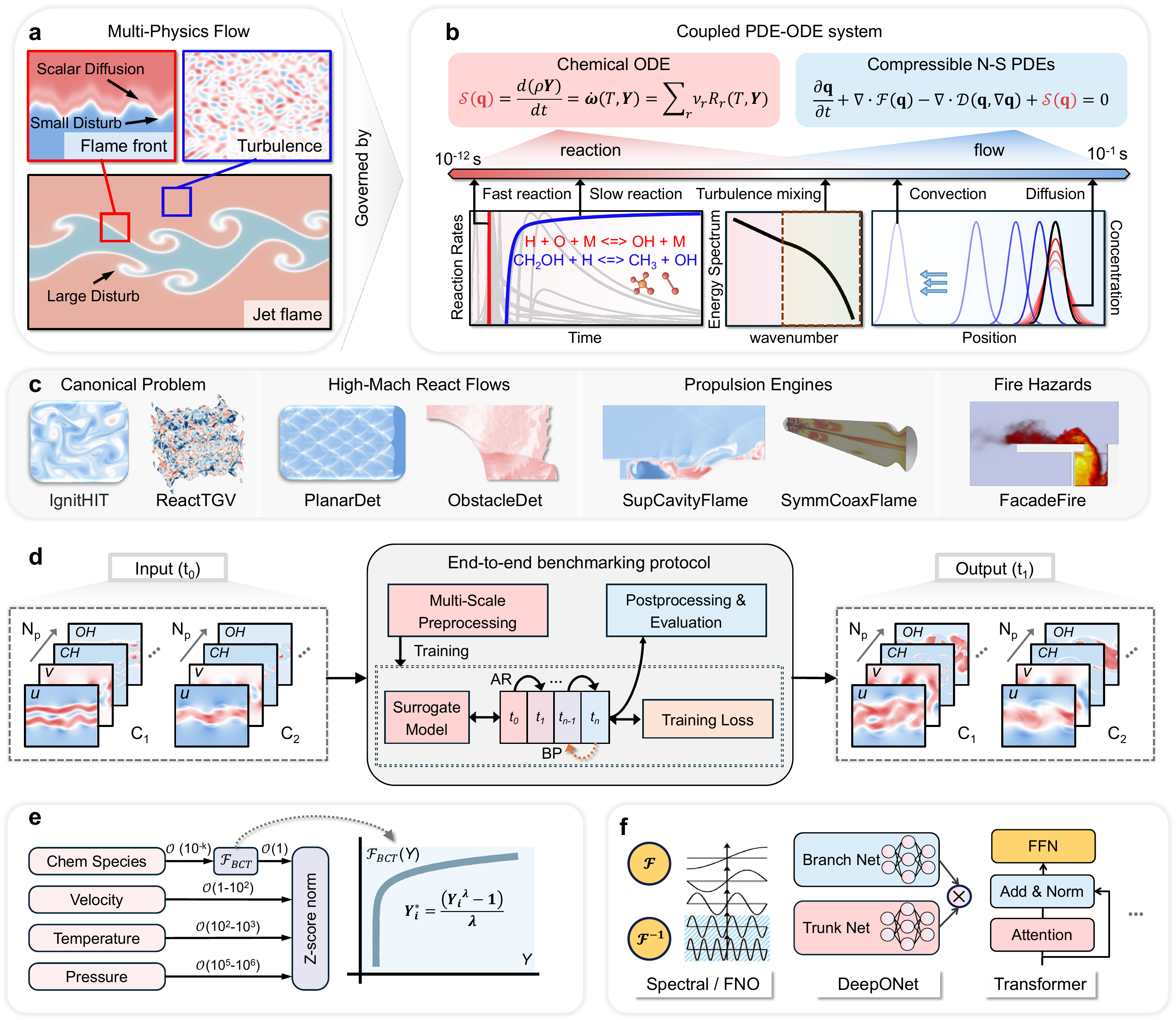}
   \caption{\textbf{Overview of REALM benchmark and problem setting.} \textbf{a}, Multiphysics reactive flow. Illustration of a jet-flame configuration where large-scale coherent motions interact with small-scale eddies; at the flame front, scalar diffusion regularizes steep gradients. 
   \textbf{b}, Coupled PDE--ODE dynamics. The reactive flow follows the compressible Navier--Stokes system with diffusion and chemistry (as sketched in the panel). Here, $\mathbf{q}$ denotes the conserved state vector; $\mathcal{F}(\mathbf{q})$ are the convective fluxes; $\mathcal{D}(\mathbf{q},\nabla\mathbf{q})$ collects diffusive contributions; and $\mathcal{S}(\mathbf{q})$ is the source obtained from the chemical ODE system. The panel highlights severe scale separation (chemistry $10^{-12}-10^{-9}$\,s vs.\ flow $\mathcal{O}(10^{-1})$\,s), the coexistence of fast and slow pathways, turbulence mixing across wavenumbers, and the contrasting signatures of convection and diffusion in concentration profiles.
   \textbf{c}, Dataset examples used in REALM, spanning canonical problems, high-Mach reactive flows, propulsion-engine scenarios, and fire-hazard cases. 
   \textbf{d}, REALM training and evaluation protocol: multi-scale preprocessing and training with autoregressive. Inputs/outputs are shown for multiple operating conditions $\{C_1,C_2,\ldots\}$ with $N_p$ fields per state.
   \textbf{e}, REALM multi-scale preprocessing: species mass fractions undergo a box-cox-type transform $\mathcal{F}_{\mathrm{BCT}}$ to compress dynamic range from $\mathcal{O}(10^{-k})$ to $\mathcal{O}(1)$, followed by $z$-score normalization for all variables.
   \textbf{f}, Surrogate model families supported in REALM: the framework is model-agnostic and applies the same protocol across operator families, including spectral operators, convolutional backbones, transformer-style models, pointwise models, and mesh/graph or point-cloud models.}
   \label{fig:framework}
\end{figure}

Building on Eqs.~(\ref{eq:ns})-(\ref{eq:chem}), the resulting reactive-flow system exhibits three defining characteristics (Fig.~\ref{fig:framework}\textbf{a,b}): (i) \emph{severe space-time scale separation}: spatial coupling of large- and small-scale disturbances and a temporal gap between chemistry ($10^{-12}-10^{-9}$\,s) and flow ($\mathcal{O}(10^{-1})$\,s); (ii) \emph{strong stiffness} concentrated in the reactive source $\mathcal{S}(\mathbf{q})$ due to coexisting fast and slow pathways; and (iii) \emph{high-dimensionality} arising from advection-diffusion tightly coupled with multi-species kinetics and transport. Together, these traits render the problem intrinsically multiscale and ill-conditioned for time integration, posing a stringent target for predictive modeling.

\subsection*{Learning Framework}

The benchmark’s primary learning task is neural surrogate modeling for multiphysics flows: we learn a time-advancing neural surrogate that maps function-valued inputs to function-valued outputs on the native mesh. The task is illustrated in Fig.~\ref{fig:framework}\textbf{d} and instantiated in a representative scenario (the 2D jet flame in Fig.~\ref{fig:framework}\textbf{c}). For a physical condition $C \sim \mathcal{C}$, where $\mathcal{C}$ denotes the distribution over possible conditions, and with associated computational mesh $\mathcal{M}_C$, let $\mathbf{U}_C(\mathbf{x},t)\in\mathbb{R}^{N_p}$ denote the vector of $N_p$ physical fields (e.g., $\rho,\,T,\,\mathbf{u},\,\mathbf{Y}$) at $\mathbf{x}\in\mathcal{M}_C$ and time $t$. 
Given snapshots $\{t_h\}_{h=0}^{H}$ with step $\Delta t$, we target continuous multi-step forecasting from $\mathbf{U}_C(\mathbf{x},t_0)$ to $\mathbf{U}_C(\mathbf{x},T)$ with $T=t_0+H\Delta t$ in an autoregressive manner:
\begin{equation}
\widehat{\mathbf{U}}_{C,h+1} \;=\; \mathcal{G}_\theta\!\big(\mathbf{U}_{C,h};\,\Delta t,\,C\big).
\end{equation}

To train neural surrogates effectively on tightly coupled PDE--ODE dynamics, we adopt a unified training pipeline in REALM that standardizes multi-scale preprocessing and temporal rollout (Fig.~\ref{fig:framework}\textbf{d}-\textbf{f}). In particular, we apply a channel-consistent preprocessing strategy: species mass fractions, which span several orders of magnitude, are first mapped to an $\mathcal{O}(1)$ range via a monotone Box--Cox transformation $\mathcal{F}_{\mathrm{BCT}}$~\cite{box1964analysis} and then jointly normalized with all other physical fields, mitigating cross-channel scale disparity (Fig.~\ref{fig:framework}\textbf{e}). Time evolution is learned with a short-horizon autoregressive scheme and backpropagation only through the final step, improving long-horizon stability under stiff reaction-transport coupling while keeping training efficient. The training pipeline is architecture-agnostic (Fig.~\ref{fig:framework}\textbf{f}), allowing spectral/Fourier surrogates, convolutional backbones, transformer-style surrogates, pointwise models, and mesh/graph surrogates to be trained under the same protocol, enabling fair and reproducible comparisons across surrogate model families.

\subsection*{Datasets}
\begin{figure}[t!]
  \centering
   \includegraphics[width=\linewidth]{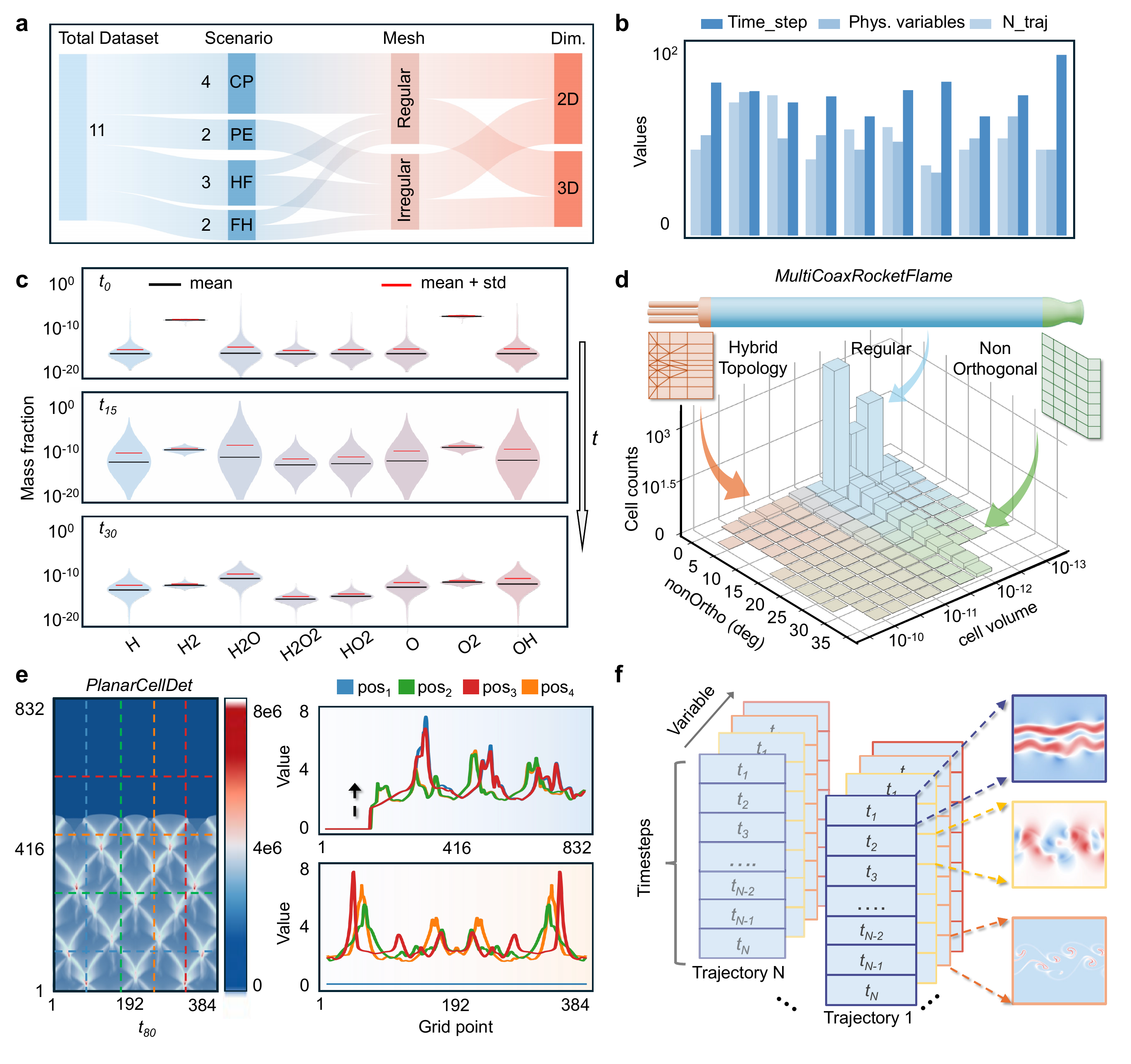}
   \caption{\textbf{Dataset statistics and packaging.} \textbf{a}, Taxonomy of all cases in the suite, grouped by scenario [Canonical Problems (CP), High-Mach Reacting Flows (HF), Propulsion Engines (PE), and Fire Hazards (FH)], by mesh type (regular vs.\ irregular), and by dimensionality (2D/3D).
   \textbf{b}, Global scale of the suite: per-case bars show the number of trajectories ($N_{\mathrm{traj}}$), the number of physical variables, and the total data volume.
   \textbf{c}, Dynamic range of key species in the \emph{EvolveJet} case, shown as violin plots of  mass fraction at three representative times. Dots and red bars mark the mean and mean $\pm$ standard deviation.
   \textbf{d}, Cell-quality landscape for a representative irregular case (\emph{MultiCoaxFlame}). A 2D histogram over cell non-orthogonality and cell volume shows hybrid topology with regular and non-orthogonal regions, illustrating geometric heterogeneity faced by irregular-mesh surrogates.
   \textbf{e}, Spatial variability in a 2D detonation example. Left: spatial distribution of the peak pressure $p_{\max}$ at a selected time. Right: spatial profiles along four horizontal and vertical sample lines at the same time, revealing intermittent peaks and strongly nonstationary structure across the grid.
   \textbf{f}, Data organization. Each case contains multiple trajectories; each trajectory consists of a sequence of time steps; each time step stores a stack of physical variables. Thumbnails on the right depict typical fields for reference.
   }
   \label{fig:data}
\end{figure}

Building on the governing equations and the multi-step forecasting task, the dataset suite in the REALM benchmark instantiates these settings in practice and enables within-scenario generalization. The datasets comprise 11 cases across four subsets: Canonical Problems (CP), High-Mach Reacting Flows (HF), Propulsion Engines (PE), and Fire Hazards (FH). Representative examples are shown in Fig.~\ref{fig:framework}\textbf{c}. As summarized in Fig.~\ref{fig:data}\textbf{a},\textbf{b}, domains span 2D and 3D on both regular and irregular meshes, with cell counts from $\sim 2\times 10^{4}$ to $\sim 1.2\times 10^{7}$. Channel counts range from $N_p=6$ to $40$ and include $\rho$, $T$, $\mathbf{u}$, $p$, and species $\mathbf{Y}$. Each case defines a fixed native mesh and a controlled set of physical conditions $C$, and provides uniformly sampled trajectories with $20$-$50$ time steps. All data are standardized into per-trajectory tensors on the native mesh $\mathcal{M}_C$ with axes \emph{timesteps} ($t$) and \emph{channels} ($N_p$), and we publish fixed train/val/test splits by trajectory index (cf. Fig.~\ref{fig:data}\textbf{f}). 

The chosen cases are representative of both idealized scenario and engineering applications, emphasizing regimes where multiphysics modeling is most challenging. First, the data are high-dimensional with strong cross-variable scale disparity and pronounced temporal nonstationarity, as illustrated by the jet-flame example in Fig.~\ref{fig:data}\textbf{c} where channel-wise ranges span orders of magnitude and also dramatically evolve over time. Second, spatial intermittency is prominent in high-Mach scenarios (Fig.~\ref{fig:data}\textbf{e}), where combustion waves and shocks form cellular structures and flame fronts with sharp interfaces and steep pressure or composition gradients. Third, realistic propulsion and large-scale fire cases employ irregular meshes with heterogeneous cell volumes and aspect ratios and multi-patch boundaries (Fig.~\ref{fig:data}\textbf{d}), which introduce complex geometries and irregular stencils. All datasets are generated from high-fidelity simulations tailored to the intended operating regimes. Case-specific descriptions are provided in \textcolor{blue}{Methods}, while the data-generation procedures and computational settings are documented in \textcolor{blue}{Supplementary Note A}. In addition, three illustrative case videos are included in \textcolor{blue}{Supplementary Note C}.

\subsection*{2D Regular Cases}

We begin with 2D problems on a regular mesh. The suite comprises three cases: ignition in homogeneous isotropic turbulence (\emph{IgnitHIT}), an evolving jet flame (\emph{EvolveJet}), and a high Mach number detonation with strong spatial discontinuities (\emph{PlanarDet}). Together, these span two combustion archetypes: premixed flame propagation and detonation-driven reactive shocks. All three include many species, evolve rapidly in space and time, and exhibit pronounced multiscale behavior. We evaluate five spectral models (\textbf{FNO}~\cite{li2021fourier}, \textbf{FFNO}~\cite{tran2023factorized}, \textbf{CROP}~\cite{gao2025discretization}, \textbf{DPOT}~\cite{hao2024dpot}, \textbf{UNO}~\cite{rahmanu}), one convolutional model (\textbf{CNext}~\cite{liu2022convnet}), four transformer style models (\textbf{FactFormer}~\cite{li2023scalable}, \textbf{Transolver}~\cite{wu2024transolver}, \textbf{ONO}~\cite{xiao2023improved}, \textbf{GNOT}~\cite{hao2023gnot}), and one pointwise neural PDE solver (\textbf{DeepONet}~\cite{lu2021learning}). DeepONet is used in a time-conditioned form; all other models realize the time-advancing operator \(\mathcal{G}_\theta\) in an autoregressive setting. All models share an identical training pipeline. The results shown here correspond to the best configurations selected under our tuning protocol; a complete search procedure is provided in \textcolor{blue}{Supplementary Note B}.

\begin{figure}[t!]
  \centering
  \includegraphics[width=\linewidth]{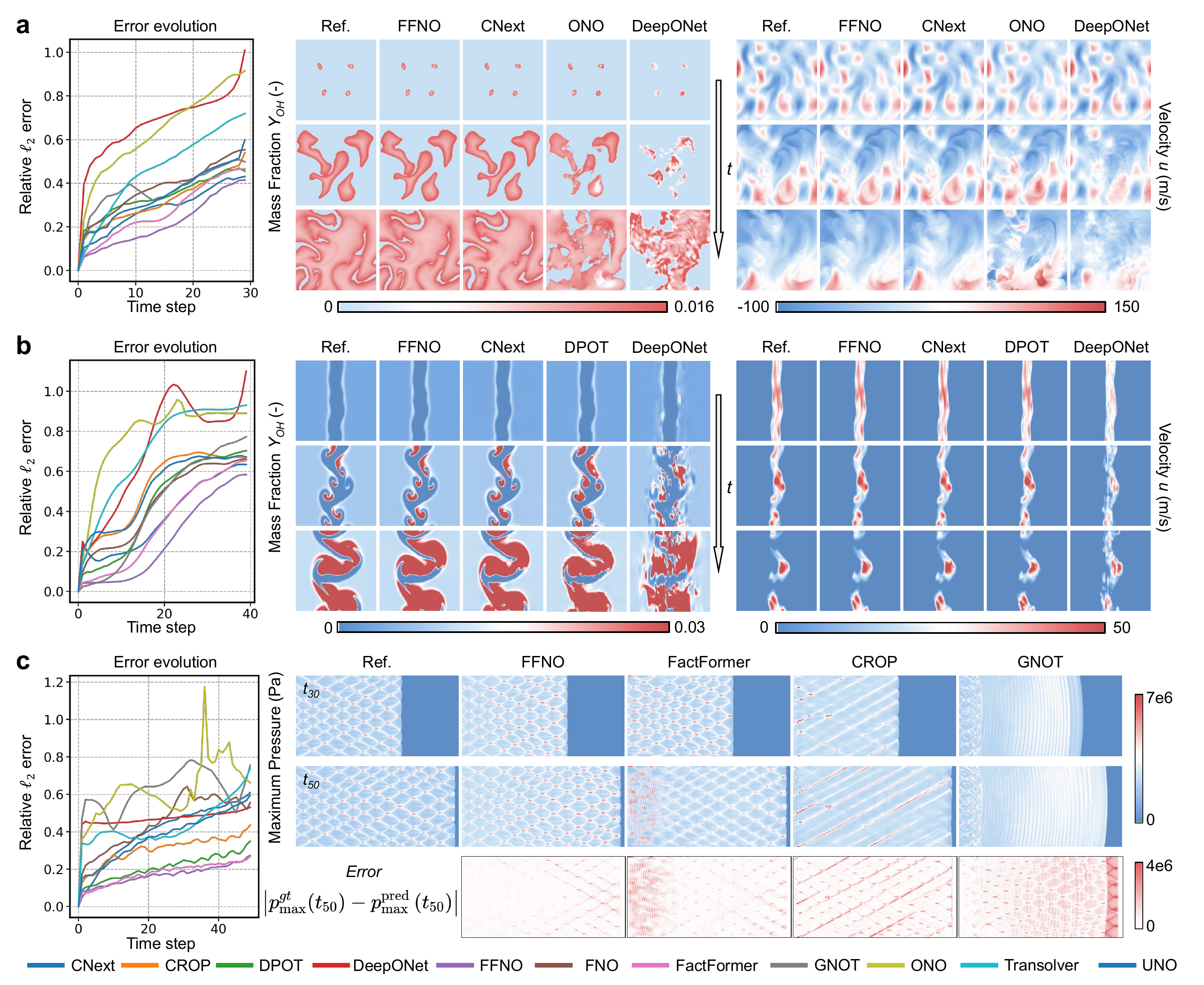}
   \caption{\textbf{2D regular cases: quantitative errors and visual comparisons.}
   \textbf{a}, \emph{IgnitHIT.} Left: rollout error evolution (relative $\ell_2$) over time. 
   Right: snapshots of OH mass fraction $Y_{\mathrm{OH}}$ and streamwise velocity $u$ for the reference and representative surrogates; arrows indicate increasing time.
   \textbf{b}, \emph{EvolveJet.} Left: rollout error evolution (relative $\ell_2$). 
   Right: snapshots of $Y_{\mathrm{OH}}$ and $u$ for the reference and representative surrogates. 
   \textbf{c}, \emph{PlanarDet.} Left: rollout error evolution (relative $\ell_2$). 
   Right: maximum pressure fields $p_{\max}$ at representative times ($t_{30}$, $t_{50}$) for the reference and representative surrogates, together with error maps at $t_{50}$ showing $\big|p_{\max}^{\mathrm{gt}}(t_{50}) - p_{\max}^{\mathrm{pred}}(t_{50})\big|$.
}
   \label{fig:2Dstructured}
\end{figure}

The three cases from Fig.~\ref{fig:2Dstructured}\textbf{a} to~\ref{fig:2Dstructured}\textbf{c} correspond to increasing difficulty, progressing from smoother dynamics to finer multiscale structure and finally to discontinuities that challenge surrogate stability and fidelity. Overall, FFNO consistently shows the lowest error growth and preserves coherent structures over long horizons; in the detonation case, it is the only model among those tested that reliably maintains the cellular pattern and the propagation speed. The convolutional CNext is the next most reliable, performing competitively on the first two cases and remaining reasonably accurate on the detonation case. Among transformer-style models, FactFormer performs moderately well, whereas other variants (Transolver, ONO, GNOT) degrade faster when shocks or thin reaction zones dominate. Classical FNO and CROP lag behind FFNO on long horizons, and the pointwise DeepONet struggles on all three cases.

In the \emph{IgnitHIT} case (cf. Fig.~\ref{fig:2Dstructured}\textbf{a}) and the \emph{EvolveJet} case (cf. Fig.~\ref{fig:2Dstructured}\textbf{b}), we visualize the OH mass fraction and the streamwise velocity $u$ the same timesteps. Under the REALM protocol, different physical fields exhibit consistent model behavior rather than one field succeeding while another fails. For example, ONO and DeepONet in Fig.~\ref{fig:2Dstructured}\textbf{a} and \ref{fig:2Dstructured}\textbf{b} show similar biases in both $\mathrm{OH}$ and velocity, characterized by field discontinuities and a loss of fine scale structures. For both cases, CNext recovers the large scale topology and plume break up, although high gradient regions display saw tooth artifacts that reduce smoothness. FFNO best preserves the global roll up and fine filaments and shows the slowest error growth. Compared with the \emph{IgnitHIT} case, the \emph{EvolveJet} case is more challenging, and the average $\ell_2$ loss grows faster for all models. On the one hand, the long rollout horizon increases the difficulty; on the other hand, periodic boundaries further introduce errors near the domain edges where structures fail to wrap cleanly. DPOT produces a scaly, patch-like artifact pattern suggestive of over-regularized spectral mixing. The structural-similarity curve for the $\mathrm{OH}$ field decays over time and closely tracks the average relative $\ell_2$ error, confirming that quantitative loss growth aligns with degradation of coherent structure. Results for the other models on the $\mathrm{OH}$ field are provided in Extended Fig.~\ref{fig:extendfig1}.

Figure~\ref{fig:2Dstructured}\textbf{c} and Extended Fig.~\ref{fig:extendfig1}\textbf{c} presents the planar detonation case, which is a classical reacting flow challenge that requires accurate wavefront propagation and recovery of fine cellular patterns. Motivated by the strong cross field consistency observed in Fig.~\ref{fig:2Dstructured}\textbf{a} and \ref{fig:2Dstructured}\textbf{b}, we show predictions of a single diagnostic field, the maximum pressure, at $t_{30}$ and $t_{50}$ together with the corresponding error map at $t_{50}$. This case is the most demanding in the suite, since it requires capturing both the propagating primary wavefront and the cellular microstructure, each of which depends on accurately resolving sharp spatial gradients. As shown in Extended Fig.~\ref{fig:extendfig1}\textbf{d}, many models successfully track the primary wavefront, likely because phase or position errors on the leading shock incur large loss penalties and therefore receive stronger gradient signals during training. In contrast, only a few models are able to reproduce the cellular microstructure. FFNO and CNext retain the overall cell topology and shock curvature, with most discrepancies appearing as amplitude bias and small phase offsets. FactFormer performs competitively but develops ripple like oscillations in the upstream region that suggest overactive long range mixing. CROP and GNOT miss parts of the cellular lattice, either merging neighboring cells or fragmenting them into irregular patches.

\subsection*{3D Regular Cases}

Next, we consider three problems in three dimensions on regular mesh: a reacting Taylor-Green vortex (\emph{ReactTGV}), a buoyancy driven pool fire (\emph{PoolFire}), and a propagating premixed flame against planar flow superimposed with homogeneous isotropic turbulence (\emph{PropHIT}). These cases push surrogate models into a high-dimensional regime, with meshes up to $\mathcal{O}(10^{7})$ cells and fully 3D coherent spatiotemporal structures, including vortex tubes and reacting sheets that must be preserved over long rollouts. We evaluate essentially the same model families as in the 2D regular study; three transformer style variants (FactFormer, ONO, GNOT) are omitted here due to training memory limits, because the number of grid cells directly controls the size of intermediate feature tensors that must be stored during backpropagation. All results reported below correspond to the best configurations selected under our capacity aligned tuning protocol (\textcolor{blue}{Supplementary Note B}).

\begin{figure}[t!]
  \centering
  \includegraphics[width=\linewidth]{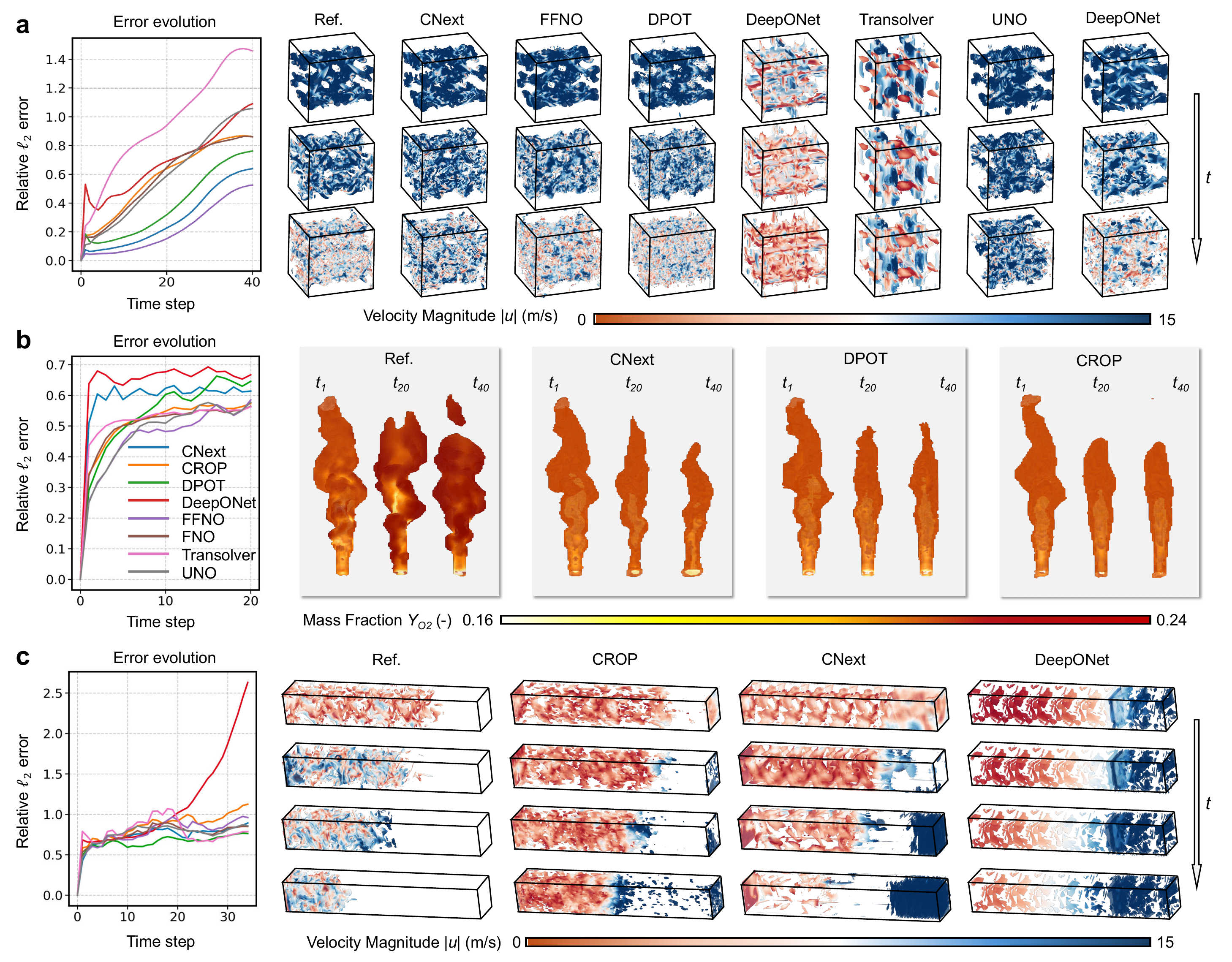}
   \caption{\textbf{3D regular cases: quantitative errors and visual comparisons.}
   \textbf{a}, \emph{ReactTGV.} Left: rollout error evolution (relative $\ell_{2}$) over time. 
   Right: vorticity isosurfaces colored by velocity magnitude $|u|$ for the reference and representative neural surrogates; arrows indicate increasing time.
   \textbf{b}, \emph{PoolFire.} Left: rollout error evolution (relative $\ell_{2}$). 
   Right: temperature isosurfaces at three representative times ($t_{1}, t_{20}, t_{40}$), colored by oxygen mass fraction $Y_{\mathrm{O_2}}$, for the reference and representative neural surrogates.
   \textbf{c}, \emph{PropHIT.} Left: rollout error evolution (relative $\ell_{2}$). 
   Right: vorticity isosurfaces colored by velocity magnitude $|u|$ for the reference and representative neural surrogates.
}
   \label{fig:3Dstructured}
\end{figure}

Training in 3D regular settings is markedly more challenging than in two dimensions. Across all three cases, the neural surrogates struggle to preserve fine scale structure, and the accumulated discrepancy becomes pronounced over longer rollouts. This is also reflected in the rollout $\ell_2$ error curves, which grow much faster than in the 2D cases. Overall, FFNO and DPOT deliver the strongest performance, consistent with the 2D results, while DeepONet and Transolver tend to lag behind. In \emph{ReactTGV} (Fig.~\ref{fig:3Dstructured}\textbf{a}), FFNO and DPOT capture the progressive sharpening of vortical structures and the large scale phase evolution, although the smallest filaments remain under resolved. DeepONet and Transolver fail to reproduce the temporal cascade and exhibit noticeable shape discrepancies. In \emph{PoolFire} (Fig.~\ref{fig:3Dstructured}\textbf{b}), temperature isosurfaces reveal the 3D flame morphology; none of the models accurately reproduce the buoyancy driven pulsation and skirt dynamics, and the shedding frequency is often biased. In \emph{PropHIT} (Fig.~\ref{fig:3Dstructured}\textbf{c}), the propagating front induces a clear vorticity asymmetry between burned and unburned regions due to thermal expansion. Most neural surrogates capture this one-sided vorticity signature and the bulk propagation, but they miss fine corrugations and underpredict sheet curvature. Additional visualizations and per-family comparisons are provided in the Extended Fig.~\ref{fig:extendfig2}.

\subsection*{Irregular Cases}
We finally evaluate neural surrogates on irregular meshes, covering three 2D cases, a supersonic cavity flame (\emph{SupCavityFlame}), an obstacle detonation (\emph{ObstacleDet}), and a symmetric flame (\emph{SymmCoaxFlame}), and two 3D cases, a multi coaxial rocket flame (\emph{MultiCoaxFlame}) and a building facade fire (\emph{FacadeFire}). These scenarios correspond to practical applications, including screeching jets, rocket engines, and building fire safety. They feature irregular geometries, nonorthogonal cells, and strong mesh nonuniformity, all of which increase the difficulty of training. For these cases we benchmark one spectral operator (\textbf{LSM}~\cite{wu2023solving}), one transformer model (\textbf{Transolver}), two pointwise models (\textbf{DeepONet}, \textbf{PointNet}~\cite{qi2017pointnet}), and three mesh/graph models (\textbf{GraphSAGE}~\cite{hamilton2017inductive}, \textbf{GraphUNet}~\cite{gao2019graph}, \textbf{MGN}~\cite{pfaff2020learning}). All models follow the REALM protocol with identical data splits and normalization. The results reported below correspond to the best configurations selected under our capacity aligned tuning procedure (\textcolor{blue}{Supplementary Note B}).

\begin{figure}[t!]
  \centering
  \includegraphics[width=\linewidth]{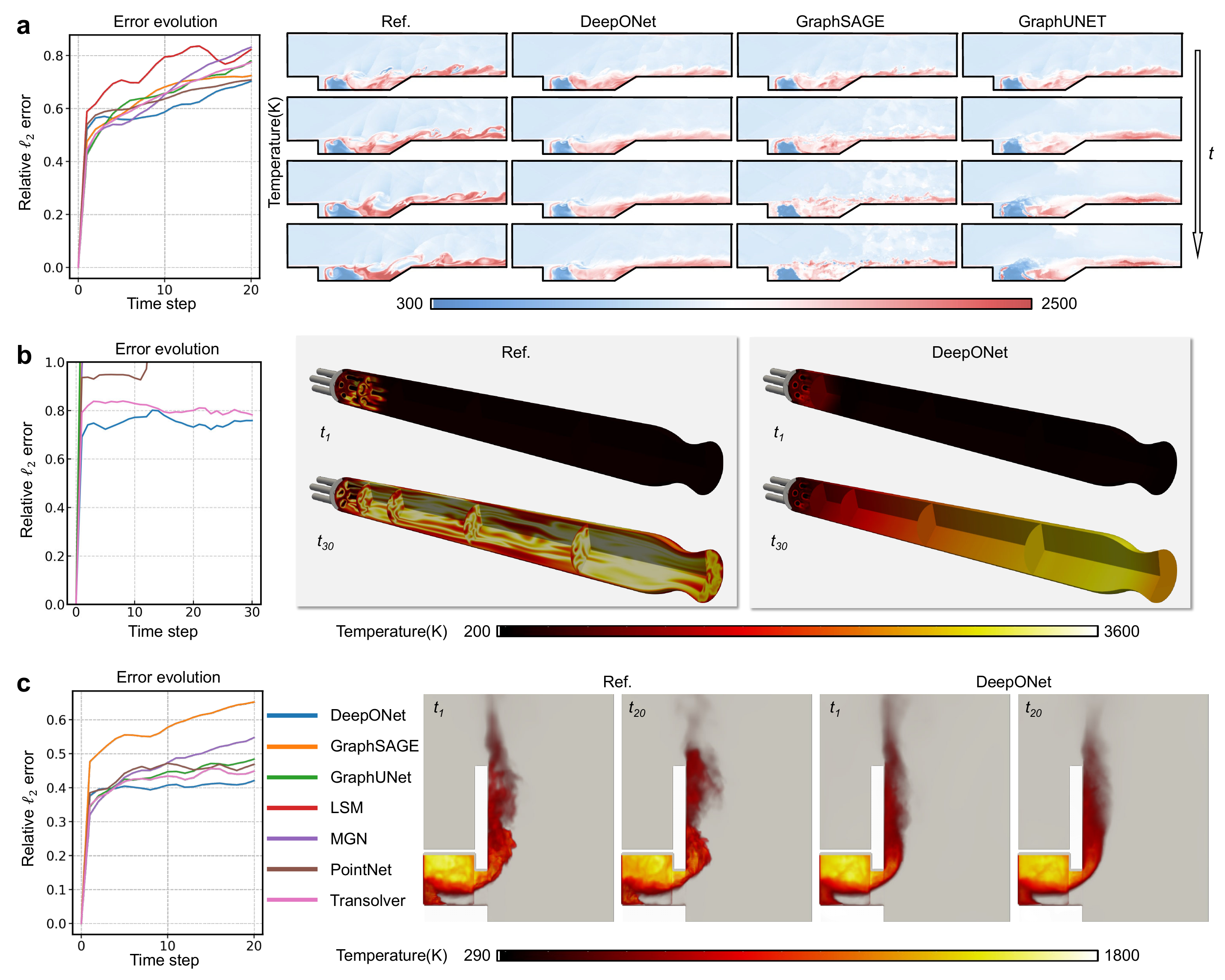}
   \caption{\textbf{Irregular cases: quantitative errors and visual comparisons.}
   \textbf{a}, \emph{SupCavityFlame.} Left: rollout error evolution (relative $\ell_{2}$) over time. Right: snapshots of the temperature field for the reference and representative neural surrogates.
   \textbf{b}, \emph{MultiCoaxFlame.} Left: rollout error evolution (relative $\ell_{2}$). Right: snapshots of the temperature field for the reference and representative neural surrogates.
   \textbf{c}, \emph{FacadeFire.} Left: error evolution (relative $\ell_{2}$). Right: volume rendering of the temperature field for the reference and representative neural surrogates.}
   \label{fig:unstructured}
\end{figure}

Figure~\ref{fig:unstructured}, together with Extended Fig.~\ref{fig:extendfig3}, summarizes performance on all irregular cases and highlights the sharp increase in difficulty relative to regular settings: most neural surrogates capture only coarse patterns while missing fine scale features, which leads to faster error growth. Overall, in contrast to its weaker performance on regular cases, DeepONet attains the lowest losses across all irregular cases, suggesting that its pointwise operator representation is comparatively robust to irregular meshes and nonuniform sampling. By contrast, graph based models tend to underperform, which we attribute to training instability and over smoothing in deep message passing on highly anisotropic, multi resolution meshes. In \emph{SupCavityFlame} (Fig.~\ref{fig:unstructured}\textbf{a} and Extended Fig.~\ref{fig:extendfig3}\textbf{a}), the macroscopic state is quasi-steady across time, so the target is the evolution of small-scale flow and flame structures and the shock system; most models fail to reproduce these dynamics, and LSM and Transolver exhibit noticeable spatial averaging and over-smoothing. In \emph{ObstacleDet} (Extended Fig.~\ref{fig:extendfig3}\textbf{b}), which couples large-scale wave propagation with intermittent micro-scale discontinuities, many neural surrogates effectively low-pass filter the fine features; only DeepONet and GraphUNet recover the global propagation and arrival times, albeit with smoothed fronts. In \emph{SymmCoaxFlame} (Extended Fig.~\ref{fig:extendfig3}\textbf{c}), DeepONet and LSM qualitatively track the large-scale jet penetration and cavity filling but strongly smooth out the reacting shear layers, whereas MGN rapidly departs from the reference and collapses into noisy, over-diffused temperature fields.

The difficulty further escalates in the 3D irregular cases, \emph{MultiCoaxFlame} (Fig.~\ref{fig:unstructured}\textbf{b}) and \emph{FacadeFire} (Fig.~\ref{fig:unstructured}\textbf{c}). In the rocket flame, most neural surrogates incur almost unit relative $\ell_2$ error within only a few steps and completely lose the cellular combustion structures, while DeepONet remains the best of the group but still produces an over-diffused, overly hot plume along the tube. In \emph{FacadeFire}, all methods accumulate substantial error during rollout; DeepONet again maintains the lowest error and visually plausible flame and smoke plumes, whereas graph-based models such as GraphSAGE degrade fastest and fail to track the rising fire front.

\begin{figure}[t!]
  \centering
  \includegraphics[width=\linewidth]{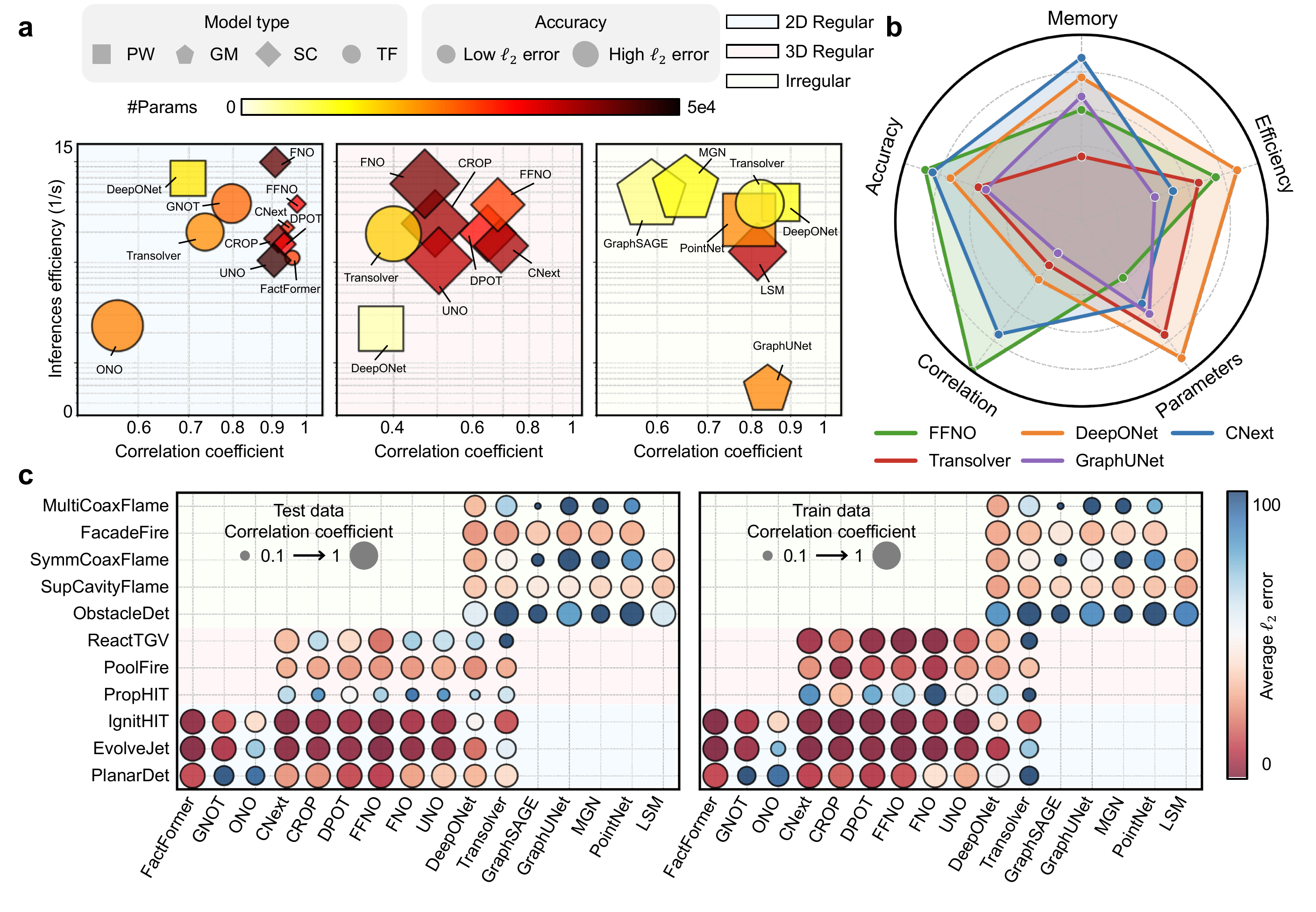}
   \caption{\textbf{Cross-benchmark model comparison.}
   \textbf{a}, \emph{Per-category scatter plots} summarizing average surrogate performance on 2D-regular (left), 3D-regular (middle), and irregular (right) cases. The horizontal axis shows the correlation coefficient with the reference solution (higher is better), and the vertical axis shows inference efficiency in rollouts per second (higher is better). Marker \emph{area} encodes the average relative $\ell_2$ error (larger markers indicate lower error). Marker \emph{color} encodes the number of trainable parameters (\#Params; light to dark indicates larger models). Marker \emph{shape} distinguishes model families: pointwise (PW, circle), graph/mesh (GM, pentagon), spectral / convolutional (SC, diamond), and transformer style (TF, square).
   \textbf{b}, \emph{Radar chart} of the mean performance of five representative models over all cases, comparing accuracy (relative $\ell_2$ error), correlation, efficiency (inference speed), parameter count, and memory footprint. A larger enclosed area indicates a more favorable overall trade-off.
   \textbf{c}, \emph{Per case performance summary} for all selected models under the capacity aligned protocol. Left: test data; right: train data. Each dot corresponds to one model-case pair; marker \emph{area} encodes the correlation coefficient (larger circles indicate higher correlation), and marker \emph{color} encodes the rollout averaged relative $\ell_2$ error. Training errors are computed with the same multi step evaluation protocol used for testing.}
   \label{fig:summary}
\end{figure}

\section*{Discussion}
This paper introduces REALM, a unified benchmark and protocol framework designed to evaluate neural surrogates for coupled PDE--ODE multiphysics systems under realistic spatiotemporal scenarios. The target problems are of clear science and engineering significance with extraordinary computational difficulty: stiff chemistry coupled with compressible transport, wide ranges of time and length scales, and strong intermittency that arise in real devices and hazards. REALM provides a comprehensive suite of eleven datasets, spanning canonical reacting flows to high-Mach propulsion configurations and large-scale fire scenarios, across both regular and irregular meshes in 2D and 3D. These cases naturally embed intricate chemistry and flow coupling, broad operating envelopes, and complex geometries, which force models to confront genuine scale separation and nonstationary dynamics. Furthermore, REALM establishes a model-agnostic, end-to-end framework that standardizes preprocessing and provides capacity-aligned model presets.

Based on REALM, we evaluate over a dozen representative surrogate model families across all cases and observe three conclusions that hold robustly across dimensionality, physics, and discretization. \textbf{(1). A scaling barrier governed jointly by dimensionality, stiffness, and mesh regularity.} As we move from 2D to 3D, and from moderately stiff problems to strongly coupled, highly stiff multiphysics regimes, rollout errors grow markedly faster, and correlations drop. Even in 2D regular settings that resemble prior benchmarks, reliability is not guaranteed, where detonation cases with steep pressure and composition gradients already exhibit rapid error accumulation and loss of cellular structure. 3D irregular discretizations further exacerbate this difficulty. In these settings, irregular stencils, non-orthogonality, and non-uniform resolution result in instability, indicating that current neural surrogates struggle to scale to realistic engineering demands.instability. \textbf{(2). Architectural inductive biases govern performance, rather than the parameter count.} On regular meshes, spectral and convolutional operators dominate because they encode resolution or translation invariance and long-range multi-scale coupling. Transformers are parameter-efficient but memory-limited at realistic 3D resolutions, and pointwise operators underperform due to weak spatial priors. Conversely, on irregular meshes, translation-invariant operators degrade while pointwise models such as DeepONet become comparatively robust, and graph/mesh networks frequently over-smooth due to anisotropic message passing. These results highlight that matching the model's inductive bias to the underlying discretization is more critical than raw expressivity. \textbf{(3). A persistent gap between nominal accuracy metrics and physically trustworthy behavior on realistic tasks.} For instance, on the 2D PlanarDet case, several models achieve correlation coefficients above 0.8 with the reference, yet still fail to reproduce the temporal evolution of the mean detonation cell size (Extended Data Fig.~\ref{fig:extendfig1}\textbf{d}), which is crucial for assessing detailed detonation dynamics. Furthermore, across irregular cases such as SupCavityFlame and FacadeFire, models that attain seemingly reasonable testing correlations still lose a large fraction of fine-scale structures, distort key integral quantities, or develop spurious diffusion and oscillations over rollout, resulting in over-smoothing fields. These trade-offs are quantitatively summarized in the accuracy and efficiency plots in Fig.~\ref{fig:summary}\textbf{a}, the multidimensional radar chart comparing five representative models in Fig.~\ref{fig:summary}\textbf{b}, and the per case training and test correlation based accuracy distributions in Fig.~\ref{fig:summary}\textbf{c}, where systematic gaps between training and test correlations reveal nontrivial overfitting in several cases.

Synthesizing these empirical findings offers broader insights into the future trajectory of AI-driven scientific computing. The three trends identified here, including scaling barriers tied to dimensionality, stiffness, and mesh irregularity, the primacy of architectural inductive biases, and the mismatch between nominal accuracy metrics and physically trustworthy behaviour, point to concrete priorities for next-generation neural surrogates. Addressing the scaling barrier will require architectures and training schemes that explicitly confront stiff, multiscale PDE--ODE couplings, for example, through multi-rate integration, physics-encoded solvers, or curriculum learning strategies that progressively increase problem complexity. Second, the central role of inductive bias calls for geometry-aware operators. In particular, a key direction is to enhance model expressivity for 3D problems on irregular meshes, while incorporating stronger physical priors to mitigate overfitting. At the same time, the observed accuracy-physics gap highlights the need for evaluation protocols that emphasize conservation properties, integral quantities, and long-horizon statistics rather than relying solely on pointwise errors or correlations, which in turn motivate new loss functions and training objectives aligned with physical reliability. The transition from learning smooth, canonical flows to modelling stiff, chemically reacting environments thus represents a necessary leap for the community. By anchoring evaluation in these rigorous, application-driven scenarios, REALM aims to steer the development of neural surrogates away from overfitting simplified proxies and toward solving the “grand challenges” of computational engineering. We hope that this benchmark will serve as a foundational testbed, catalyzing a new generation of neural solvers capable of handling the complex, multiscale, and high-dimensional nature of real physical systems.

\section*{Methods}
We herein introduce the REALM benchmark methodology and provide case-wise dataset descriptions aligned with the training-evaluation protocol in Fig.~\ref{fig:framework}.

\subsection*{Per-case dataset descriptions}

\paragraph*{IgnitHIT$^{2d}_{\Box}$ (2D-regular mesh ignition kernel in homogeneous isotropic turbulence)\footnote{Superscripts $^{2d}$ and $^{3d}$ denote 2D and 3D problems, respectively, and subscripts $_{\Box}$ and $_{\triangle}$ denote regular and irregular meshes.}}
This case models the early-time evolution of localized hydrogen ignition kernels immersed in a statistically stationary turbulent flow.
More specifically, we consider hydrogen ignition kernels embedded in homogeneous isotropic turbulence (HIT), a canonical turbulent combustion configuration used to quantify turbulent flame propagation speed~\cite{peters1999turbulent}, characterize detailed flow-flame interaction~\cite{chaudhuri2012flame}, and probe the transition from laminar kernel growth to a self-sustained turbulent flame. Spherical (or near-spherical) kernels ignite and expand while interacting with turbulent straining, leading to wrinkling and kernel-kernel merging. The simulation runs in a \(0.05\times0.05~\mathrm{m}^2\) domain on a 2D regular mesh of \(1024\times1024\) cells. The suite comprises 36 trajectories with initial conditions differing in (i) the number of ignition kernels, (ii) 3 initial kernel geometries, and (iii) multiple turbulence intensities. From an operator-learning perspective, \textit{IgnitHIT} is highly sensitive to initial conditions: small differences in kernel size/shape or local turbulence phase lead to qualitatively different ignition progress. Thin flame reaction zones and strong turbulence impose multiscale demands; rapid topological changes stress models that assume smooth, slowly varying fields. Accurate surrogates should preserve sharp gradients and cross-channel coupling while remaining stable over rollouts, meanwhile generalizing across large variations in initial conditions.

\paragraph*{EvolveJet$^{2d}_{\Box}$ (2D-regular mesh time-evolving shear jet flame).}
This case models a methane -oxygen jet flame developing in a high-shear mixing layer, where thin reacting layers are continuously wrinkled, merged, and convected by the surrounding turbulence. In combustion terms, this corresponds to a canonical shear-layer flame that exhibits mixing-driven stabilization, strain-induced local extinction and reignition, and a sustained coupling among scalar dissipation, heat release, and vortex dynamics~\cite{karami2015mechanisms}. The simulation runs on a 2D regular mesh of \(800\times550\) cells. The suite comprises 30 trajectories spanning 3 equivalence ratios \(\phi\) (from fuel lean to rich) and 10 Reynolds numbers based on the jet core velocity and the shear-layer thickness; for 3 higher Reynolds-number settings a dual-jet variant is included to intensify shear-layer interaction. From an operator-learning perspective, \textit{EvolveJet} is convection dominated and multiscale, with stiff chemistry and strong cross-channel coupling. Variations in \(\phi\), Reynolds number, and single versus dual-jet topology induce condition shifts in heat release, flame thickness, and shear intensity, making within-scenario generalization challenging. The chemistry is comparatively rich in this case, involving 36 species, and the resulting inter-species interactions and temporal variability further increase prediction difficulty.

\paragraph*{ReactTGV$^{3d}_{\Box}$ (3D-regular mesh reacting Taylor-Green vortex).}
This case models a nonpremixed hydrogen flame embedded in a 3D vortical flow within a triply periodic box.
Specifically, it is a canonical flame-vortex interaction configuration for studying local extinction and reignition, as well as the coupled dynamics of scalar mixing, heat release, and vortex structures~\cite{abdelsamie2021taylor}. Under intense straining and vortex stretching, the thin reacting sheet exhibits intermittent extinction and reignition. In our benchmark it is run on a regular grid of \(256\times256\times256\) (about \(1.6\times10^{7}\) cells) over a millimetre-scale domain. We generate 16 trajectories by varying the Reynolds number and the characteristic mixing length of the initial composition distribution, and each trajectory provides 40 uniformly spaced snapshots on the native mesh. From an operator-learning perspective the periodic domain provides no inflow or outflow anchors, so the model must capture intrinsic phase dynamics and small phase errors can grow over long rollouts. The high resolution captures filamentary reacting sheets and sharp gradients, which makes preservation of fine-scale structure essential, and the variation across Reynolds number and mixing length changes the ratio of chemical and flow time scales, increasing the difficulty of within-scenario generalization. In addition, the large 3D mesh size is also challenging for surrogate models based on transformer and graph networks.

\paragraph*{PropHIT$^{3d}_{\Box}$ (3D-regular mesh propagating flame in homogeneous isotropic turbulence).}

This case examines a premixed hydrogen-air flame evolving in statistically stationary, homogeneous-isotropic turbulence. This is the \textit{de facto} configuration for studying turbulence-chemistry interactions~\cite{you2020modelling}. The computational domain is a rectangular box with side lengths $(L_x,L_y,L_z)$, where $L_y=L_z=5.3\,\delta_L$ (laminar flame thickness) and $L_x=8L_y=42.4\,\delta_L$. The domain is discretized on a uniform grid with $N_x\times N_y\times N_z=12N\times N\times N$ points; we fix $N=128$ for all runs, i.e., $1536\times128\times128$.  
We conduct 8 cases by crossing pressure $p\in\{2,5\}\,\mathrm{atm}$ with turbulence intensities $u'/S_L\in\{2,5,10,20\}$, where $S_L$ is the laminar flame thickness. From an operator-learning perspective, the thin, highly corrugated reaction zones undergo broadband straining and curvature, producing intermittent local extinction/reignition; the absence of external anchoring and the large domain aspect ratio make phase accuracy over long rollouts critical, and variation in pressure and turbulence intensities shifts chemical-eddy time-scale separation, tightening the generalization challenge within this scenario.

\paragraph*{PlanarDet$^{2d}_{\Box}$ (2D-regular mesh planar cellular detonation).}

This case models a planar detonation, a canonical supersonic reacting flow where the leading shock couples with auto-igniting reaction wave behind it forming a self-sustained detonation wave~\cite{oran2015understanding,powers2006review}. Transverse waves and their intersection points travel along the main shock front and collide at regular intervals, forming the characteristic cell-like patterns in 2D detonations. The simulation is run in a rectangular domain on a regular grid of \(840\times400\) cells, capturing the coupled dynamics of shock curvature, combustion chemistry, and cell formation on the native mesh. The suite comprises 9 trajectories spanning 3 equivalence ratios \(\phi\) and 3 initial temperatures \(T_0\). From an operator-learning standpoint the task is demanding: sharp discontinuities coexist with extremely thin reaction zones; accurate surrogates must preserve discontinuities without spurious smoothing, maintain the correct detonation speed, and reproduce cell size and periodicity. Spectral or convolutional priors can exhibit ringing near discontinuities, so models must balance stability with low numerical dissipation while retaining multi-species coupling in the reaction layer. The relatively sparse data points (3 conditions in \((\phi, T_0)\), respectively) further elevate the difficulty for model generalization.

\paragraph*{SupCavityFlame$^{2d}_{\triangle}$ (2D-irregular mesh supersonic cavity flame).}

This case models hydrogen injection into a supersonic air stream over an open cavity, representative of scramjet combustors.
In propulsion terms, it is a cavity-based supersonic combustion configuration, where a high-enthalpy air stream interacts with transverse fuel jets and a recirculating cavity flow that traps hot products and stabilizes the flame, leading to coupled shock-shear-flame dynamics~\cite{lin2025direct,gruber2001fundamental}. The configuration leads to coupled interactions between shocks, shear layers and the flame: the flow separates and reattaches over the cavity, pressure waves strike the reacting shear layer, and an unsteady recirculation region keeps the flame stable by triggering repeated ignition and controlling its growth through mixing. The geometry includes cavity steps and injectors, so the domain is non-uniform and non-rectilinear; accordingly, it is discretized on a 2D irregular mesh of \(\sim3\times10^{6}\) cells. The suite comprises 9 trajectories obtained by combining 3 fuel-injection velocities with 3 injector locations. \textit{SupCavityFlame} is challenging for surrogate modeling because shock-induced discontinuities coexist with thin reacting layers and abundant small-scale structure from shear-layer roll-up and shock-flame interaction. The irregular mesh introduces irregular connectivity, nonuniform cell sizes, and anisotropy, breaking translational invariance and complicating spectral or convolutional priors. Robust surrogates must therefore  preserve discontinuities and cross-channel coupling, and maintain gradient fidelity and stability over long rollouts.

\paragraph*{ObstacleDet$^{2d}_{\triangle}$ (2D-irregular obstacle attenuated detonation).}

This case models a detonation wave propagating through a 2D channel and interacting with a solid blockage.
In detonation dynamics, it is a canonical detonation-obstacle configuration: the lead shock diffracts and weakens past the blockage, the shock-reaction zone can decouple and partially quench, and re-initiation may occur after a delay or only beyond a critical threshold, capturing safety-critical transitions in high-speed reactive flows~\cite{bao2021experimental} and providing a challenging testbed for modeling shock-reaction coupling and regime changes. The computational domain is a 2D rectangle with a semicircular cutout, discretized on an irregular mesh of \(\sim3.1\times10^{5}\) cells with curved boundary conformity. The suite comprises six trajectories formed by varying the obstacle blockage height with the effective mixture reactivity. For neural surrogate modeling, this case poses great challenges in preserving sharp discontinuities without artificial smoothing, predicting phase-accurate wave speeds and cell scales over long rollouts where small shock-position errors accumulate, and handling topology changes during diffraction and re-initiation. The irregular mesh and curved wall introduce complex connectivity and nonuniform cell sizes. Robust surrogates must remain stable on the native mesh, respect cross-channel coupling, and maintain gradient fidelity near curved boundaries and reacting layers.

\paragraph*{SymmCoaxFlame$^{2d}_{\triangle}$ (2D-irregular coaxial symmetric rocket flame).} 

This case models a methane-oxygen rocket combustor with a single-element shear-coaxial injector~\cite{roth2017experimental}. A high-momentum oxygen core issues through the inner post and mixes with a surrounding methane stream, forming a strongly sheared interface where thin reaction zones convect, wrinkle, and intermittently merge under hydrodynamic instabilities. The computational domain is 2D-axisymmetric and discretized on a complex irregular mesh with 294{,}900 cells, which captures injector lip geometry, near-wall gradients, and downstream nozzle acceleration. The suite comprises 12 trajectories obtained by varying inlet temperature boundary conditions and fuel-oxidizer mixture ratio. This case is nonperiodic with multiple boundary types (inlet, wall, nozzle exit) and features pressure-heat-release coupling, density contrast across the mixing layer, and thin reacting sheets. All of these are challenging for existing machine learning methods. The irregular mesh also introduces connectivity and nonuniform cell issues, breaking translational invariance and reducing the direct applicability of spectral or standard convolutional priors. Accurate surrogates must preserve sharp gradients and cross-channel coupling, and remain stable over long rollouts.

\paragraph*{MultiCoaxFlame$^{3d}_{\triangle}$ (3D-irregular multi-coaxial rocket flame).}

This case extends ``SymmCoaxFlame'' to a more realistic 3D configuration with a seven-element methane-oxygen shear-coaxial injector array in a rocket combustor~\cite{silvestri2016characterization}. Each oxygen core issues within a surrounding methane annulus, forming strongly sheared mixing layers and thin reacting sheets. Close injector spacing introduces jet-jet interactions, vortex merging between adjacent shear layers, and enhanced 3D turbulent mixing throughout the chamber and into the nozzle, yielding markedly unsteady combustion. The domain is discretized on a 3D irregular mesh of approximately 13.5 million hybrid cells, resolving injector lips, recirculation zones, and nozzle acceleration. The suite comprises 6 trajectories formed by two mixture ratios crossed with three thrust levels.
From an operator-learning perspective, the seven-element array introduces multi-source coupling and broadband interactions across injector, chamber, and nozzle scales. The irregular grid’s irregular connectivity and nonuniform cell sizes break translational invariance and complicate spectral or standard convolutional priors, and the \(\mathcal{O}(10^7)\) degrees of freedom impose significant memory and throughput demands on surrogates. The limited number of operating conditions further increases the difficulty of condition generalization. Effective surrogates must  preserve sharp gradients and cross-channel coupling, and remain stable over long horizons.

\paragraph*{PoolFire$^{3d}_{\Box}$ (3D-regular pool fire).} 
This case models a buoyancy-driven methane-air pool fire and its rising plume. In fire science, this is a canonical configuration used to study flame spread, plume entrainment and mixing, and the coupling among heat release, soot formation, and large-scale buoyant motion~\cite{modak1981burning}. The flow is set by variable-density, buoyancy-driven turbulence with diffusion-limited combustion and exhibits the classic McCaffrey regimes~\cite{McCaffrey1979}: a continuous flame zone near the pool, an intermittent transition region with frequent local extinction/reignition, and an overlying plume region dominated by entrainment and large coherent structures. The simulation is conducted in a \(3\,\mathrm{m}\times3\,\mathrm{m}\times3\,\mathrm{m}\) domain to capture integral-length-scale dynamics such as puffing, plume meander, and far-field entrainment. The suite comprises 15 trajectories formed by crossing five heat-release rates with three inlet sizes. \textit{PoolFire} spans a wide separation of scales, from thin reacting layers to meter-scale buoyant eddies, with strong coupling among heat release, density stratification, and vertical acceleration. The plume’s low-frequency puffing and lateral meander introduce phase variability that amplifies over long rollouts; nonperiodicity and strong intermittency further stress stability. Accurate surrogates must preserve cross-channel coupling and gradients while remaining robust to large coherent motions and slow drifts in phase.

\paragraph*{FacadeFire$^{3d}_{\triangle}$ (3D-irregular building facade fire).} 
This case models a compartment fire that vents through a window and generates an upward flame along a non-combustible exterior facade.
In building fire safety, it is a canonical facade fire scenario that features vented flame attachment/detachment, buoyancy-driven plume formation, and facade-guided entrainment of ambient air~\cite{klopovic2001comprehensive,lu2025numerical}. The domain comprises a cubic combustion chamber, a square window opening, and an adjacent vertical facade, discretized on a 3D irregular mesh with \(\sim1.8\times10^{5}\) cells (overall size \(5\times6\times6~\mathrm{m}^3\)). The suite contains 9 trajectories obtained by varying the heat-release rate from \(400\,\mathrm{kW}\) to \(1200\,\mathrm{kW}\) in \(100\,\mathrm{kW}\) increments. From an operator-learning perspective, \textit{FacadeFire} spans wide spatial and temporal scales: thin reacting layers and near-wall shear coexist with meter-scale buoyant eddies, intermittent venting, and plume meander. Complex, nonperiodic boundary conditions and density stratification drive large coherent motions. The irregular mesh introduces irregular connectivity and nonuniform cell sizes, weakening translational invariance and complicating spectral or standard convolutional priors. Effective surrogates must be stable under buoyancy-induced large-scale unsteadiness.

\subsection*{Models}

We benchmark over a dozen neural surrogates and evaluate them under a standardized protocol (fixed data splits, normalization, and rollout horizon $H$). All models are trained to approximate the time-advancing operator $\mathcal{G}_\theta$ via either autoregressive or time-conditioned schemes. To ensure a rigorous comparison while preserving original inductive biases, we meticulously tuned key architectural hyperparameters, including depth, width, and the number of layers or attention heads. We establish three capacity tiers for each family: \emph{small} ($\sim$10$^6$ parameters), \emph{medium} ($\sim$10$^7$), and \emph{large} ($\sim$10$^8$). Detailed hyperparameter tuning studies are provided in \textcolor{blue}{Supplementary Notes B}. Models are categorized by their \emph{spatial inductive bias}, including pointwise, convolutional, spectral, transformer-based, and mesh/graph message-passing. A brief introduction to each family follows.

\paragraph*{Pointwise models.}
This family learns a per-point map conditioned on spatiotemporal coordinates, using shared multilayer perceptrons without explicit neighborhood aggregation. It is lightweight and geometry-agnostic, naturally applicable to both regular and irregular meshes, but provides a weaker inductive bias for long-range spatial coupling. In our benchmark, we include (i) \textbf{DeepONet}~\cite{lu2021learning}, which uses a branch encoding of the input field with a trunk encoding of the spatiotemporal coordinates $[\textbf{x},t]$ to predict the target at the corresponding location, and (ii) \textbf{PointNet}~\cite{qi2017pointnet} applies shared MLPs to $[u(\textbf{x}),\textbf{x}]$, forms a permutation-invariant global descriptor via pooling, and refines pointwise outputs with this global context. 

\paragraph*{Spectral models.}
Spectral models perform global mixing in the frequency domain, typically by projecting features to Fourier space, applying learnable filters on a truncated set of modes, and transforming back. This yields an efficient inductive bias for long-range and multi-scale coupling with relatively few parameters, and often improves stability over deep rollouts. In our benchmark, we include 
(i) \textbf{FNO}~\cite{li2021fourier} (Fourier Neural Operator), a classic spectral operator that learns mappings between functions in the frequency domain and has been shown to have consistent approximation properties,
(ii) \textbf{FFNO}~\cite{tran2023factorized} (Factorized Fourier Neural Operator), an improved version of FNO, which processes each spatial dimension separately, effectively reducing the parameter scale,
(iii) \textbf{CROP}~\cite{gao2025discretization}, a cross-resolution operator-learning pipeline that is free of aliasing and discretization mismatch errors, enabling efficient cross-resolution and multi-scale learning,
(iv) \textbf{DPOT}~\cite{hao2024dpot}, a classic pre-trained PDE foundation model, which designs a new architecture consisting of a temporal aggregation layer and multiple Fourier attention layers, 
(v) \textbf{UNO} (U-shaped Neural Operator)~\cite{rahmanu}, which implements the operator through an encoder-decoder pyramid with skip connections while maintaining discretization invariance.
(vi) \textbf{LSM} (Latent Spectral Model)~\cite{wu2023solving}, which lifts fields into a compact latent grid via an encoder, applies Fourier mixing in the latent space, and decodes back to the physical domain. Through decoupling spectral resolution from the input mesh, LSM can reduce memory/compute while retaining global spectral coupling.
This class is most straightforward to apply to regular data; accordingly, we use it only on regular cases in this benchmark.

\paragraph*{Transformer-style models.}
Attention-based models conduct the nonlocal interactions directly in the spatial domain, using positional/mesh encodings to inject geometry and sparsity/factorization to keep costs tractable. This family is flexible across discretizations and effective for long-range coupling, though memory usage can grow without locality constraints. In our benchmark, we include (i) \textbf{FactFormer}~\cite{li2023scalable}, which employs factorized/sparse attention to scale to large spatial tokens; (ii) \textbf{Transolver}~\cite{wu2024transolver}, a time-conditioned transformer that aggregates spatio-temporal dependencies for multi-step forecasting; (iii) \textbf{ONO}~\cite{xiao2023improved}, which stacks self/linear with MLP blocks and a learned feature whitening step for stable token mixing; and (iv) \textbf{GNOT}~\cite{hao2023gnot}, a linear-attention operator over flattened mesh tokens augmented with coordinate features.

\paragraph*{Convolutional models.}
CNN-based surrogate model local transport and diffusion with shared spatial filters and multi-resolution pyramids, providing a strong inductive bias for locality and scale separation at low computational cost. U-shaped decoders further enable fine-scale reconstruction after coarse-scale mixing. In our benchmark, we include \textbf{CNext}~\cite{liu2022convnet}, a lightweight multi-scale CNN backbone with residual blocks for efficient feature mixing. 

\paragraph*{Graph models.}
Graph models conduct message passing to capture interactions among meshes or particles. In our benchmark, we include (i) \textbf{MGN} (\textsc{MeshGraphNet})~\cite{pfaff2020learning}, which aggregate edge-aware features (e.g., metric tensors, face normals) for geometry-informed updates; (2) \textbf{GraphUNet}~\cite{gao2019graph}, which follows a U-shaped pooling-unpooling hierarchy on graphs to mix global context while recovering fine structures at high resolution; (3) \textbf{GraphSAGE}~\cite{hamilton2017inductive}, which performs inductive neighborhood aggregation with learnable aggregation functions to scale to large, irregular meshes.

\subsection*{Training pipeline}

\paragraph*{Preprocessing.}
Reactive flow dynamics typically involve chemical species varying across vast orders of magnitude (e.g., $10^{-12}$ to $10^{-1}$). Directly employing such raw data to formulate loss functions inevitably results in an ill-conditioned optimization landscape, thereby hindering numerical convergence. To mitigate this, we adopt a two-stage, channel-consistent normalization. First, species mass fractions are transformed by a channel-wise Box-Cox map~\cite{box1964analysis}:
\begin{equation}
\mathrm{BCT}_\lambda(y)=
\begin{cases}
\dfrac{y^{\lambda}-1}{\lambda}, & \lambda\neq 0,\\[6pt]
\log y, & \lambda=0,
\end{cases}
\qquad \text{with } \lambda=0.1 \text{ in all experiments,}
\end{equation}
so that small-magnitude mass fractions (often \(10^{-k}\)) are mapped to \(\mathcal{O}(1)\) values and become comparable in scale to the other fields. Second, all channels are standardized using training-set statistics computed after the species transform:
\begin{equation}
\widetilde{U}=\frac{U-\mu}{\sigma},
\end{equation}
where \(\mu\) and \(\sigma\) are per-channel means and standard deviations aggregated over batch, time, and space. At evaluation, we apply the inverse map to report physical units:
\begin{equation}
U=\sigma\,\widetilde{U}+\mu,\qquad
y=\mathrm{BCT}^{-1}_{\lambda}(y^\ast)=\bigl(\lambda\,y^\ast+1\bigr)^{\!1/\lambda},
\end{equation}
where \(y^\ast\) denotes the transformed species prior to inversion.

\paragraph*{Training Protocol and Objective.}
We train the surrogate model using a multistep prediction strategy to enhance stability over long horizons. Instead of standard teacher forcing, we perform a forward rollout for a fixed horizon and compute gradients solely based on the prediction error at the final step. The training objective is computed directly on the preprocessed variables $\widetilde{\mathbf{U}}$. Let $\widetilde{\mathbf{U}}_\alpha$ and $\widetilde{{\mathbf{U}}}^*_\alpha$ denote the ground-truth and predicted normalized tensors for a feature group $\alpha \in \{\mathrm{chem},\,T,\,\rho,\,\textbf{u},\,p\}$. Therefore, the total loss is defined as:
\begin{equation}
\mathcal{L} = \sum_{\alpha\in\mathcal{A}} \ell_\alpha({\widetilde{\mathbf{U}}}^*_\alpha, \widetilde{\mathbf{U}}_\alpha),
\end{equation}
where the loss $\ell_\alpha$ is the Mean Squared Error (MSE) averaged over all spatial and channel dimensions within that group. We optimize the parameters using Adam~\cite{kingma2015adam} with weight decay and a One-Cycle learning rate schedule~\cite{smith2019super}. The batch size $B$ and initial learning rate are treated as hyperparameters; their sensitivity analysis is provided in \textcolor{blue}{Supplementary Note B}.

\subsection*{Evaluation metrics}
\paragraph*{Normalized Prediction Error.} We evaluate the model using the same channel grouping $\alpha \in \{\mathrm{chem},\,T,\,\rho,\,\textbf{u},\,p\}$ as the training objective. The evaluation metric is computed directly on the preprocessed variables $\widetilde{\mathbf{U}}$ and averaged over the rollout horizon $H$ on the test data. Specifically, for each rollout step $h$, the error is the sum of the MSE for each group:
\begin{equation}
\mathcal{L}^{(h)} = \sum_{\alpha \in \mathcal{A}} \frac{1}{N_\alpha} \left\| \widetilde{\mathbf{U}}^*_{\alpha, h} - \widetilde{\mathbf{U}}_{\alpha, h} \right\|_2,
\end{equation}
where $\widetilde{\mathbf{U}}^*_{\alpha, h}$ and $\widetilde{\mathbf{U}}_{\alpha, h}$ denote the predicted and ground-truth normalized tensors for group $\alpha$ at step $h$, and $N_\alpha$ is the total number of elements in that group. The reported normalized prediction error is the temporal average $
\overline{\mathcal{L}} = \frac{1}{H}\sum_{h=1}^{H}\mathcal{L}^{(h)}$.

\paragraph{Structural metrics.}
In addition to the normalized prediction error, we assess the correlation between predicted and reference fields. For each rollout step \(h\in\{1,\ldots,H\}\) and for each physical channel \(c\in\{1,\ldots,N_p\}\), we take the decoded fields \(\mathbf{U}^{*}_{c,h}\) and \(\widehat{\mathbf{U}}_{c,h}\), flatten them to vectors on the mesh, and compute the Pearson correlation: 
\begin{equation}
  r^{(h)}_{c}\;=\;\mathrm{corr}\!\big(\mathbf{U}^{*}_{c,h},\,\widehat{\mathbf{U}}_{c,h}(\cdot,t_h)\big)
  \;=\;
  \frac{\big\langle
    \delta \mathbf{U}^{*}_{c,h},\,
    \delta \widehat{\mathbf{U}}_{c,h}
  \big\rangle}
  {\big\|\delta \mathbf{U}^{*}_{c,h}\big\|_2\,
   \big\|\delta \widehat{\mathbf{U}}_{c,h}\big\|_2},
\end{equation}
where \(\delta \mathbf{U}^{*}_{c,h}\) and \(\delta \widehat{\mathbf{U}}_{c,h}\) denote the corresponding fields with their spatial mean subtracted, and \(\langle\cdot,\cdot\rangle\) is the standard Euclidean inner product. The time-evolving correlation curve reported in the main text is the channel-averaged value
\begin{equation}
  r^{(h)} \;=\; \frac{1}{N_p}\sum_{c=1}^{N_p} r^{(h)}_{c},
\end{equation}
and we occasionally further summarize it by the temporal average \(\overline{r} = H^{-1}\sum_{h=1}^{H} r^{(h)}\).

\paragraph{Efficiency.}
We report three hardware-dependent measures under a fixed device, precision, batch size, and rollout horizon $H$: (i) \emph{parameter count} (millions), obtained from trainable parameters of the loaded checkpoint; (ii) \emph{inference time}, measured as per-rollout wall-clock by summing step-level intervals around the model call (synchronizing the device before/after timing); the per-step time is the per-rollout time divided by $H$; and (iii) \emph{peak inference memory} (MB, CUDA only), obtained by resetting the runtime peak counter prior to the rollout and reading the maximum allocated bytes after completion. Metric computation and I/O are excluded from timing; we repeat three times and report the median.

\section*{Data availability} 
All the used datasets are available on GitHub at \url{https://github.com/deepflame-ai/REALM}.  A companion website, \url{https://realm-bench.org}, provides centralized access to the data, detailed descriptions of each benchmark case, and a continuously updated leaderboard of model performance.

\section*{Code availability} 
All the source codes to reproduce the results in this study are available on GitHub at \url{https://github.com/deepflame-ai/REALM}.

\bibliographystyle{unsrt}
\bibliography{reference}

\vspace{36pt}
\noindent\textbf{Acknowledgement:}
The work is supported by the National Natural Science Foundation of China (92270203, 52441603, 523B2062, 52276096, 62276269, 6250636, 92270118). R.Z. would like to acknowledge the support from the China Postdoctoral Science Foundation under grant number 2025M771582 and the Postdoctoral Fellowship Program of CPSF under Grant Number GZB20250408.  \\

\noindent\textbf{Author contributions:} R.M., R.Z., H.S. and Z.X.C. conceived and designed the study. R.M., R.Z., X.B. and B.Z. developed the AI framework, implemented the training pipeline and carried out the surrogate-model experiments. T.Z., Z.C., M.L., T.W., Y.X., Y.X., H.Z., S.G., P.A.D., Z.A., X.L., R.B., H.G. generated and curated the datasets. H.S. and Z.X.C. supervised the project. R.M. and R.Z. drafted the manuscript; all authors contributed to research discussions and to writing and revising the paper. \\

\noindent\textbf{Correspondence to:} Hao Sun (\url{haosun@ruc.edu.cn}), Zhi X. Chen (\url{chenzhi@pku.edu.cn})\\

\noindent\textbf{Competing interests:}
The authors declare no competing interests.\\

\noindent\textbf{Supplementary information:}
The supplementary information is attached.


\clearpage
\setcounter{figure}{0}
\renewcommand{\figurename}{Extended Data Figure}
\setcounter{table}{0}
\renewcommand{\tablename}{Extended Data Table}

\begin{figure}[t!]
  \centering
   \includegraphics[width=1\linewidth]{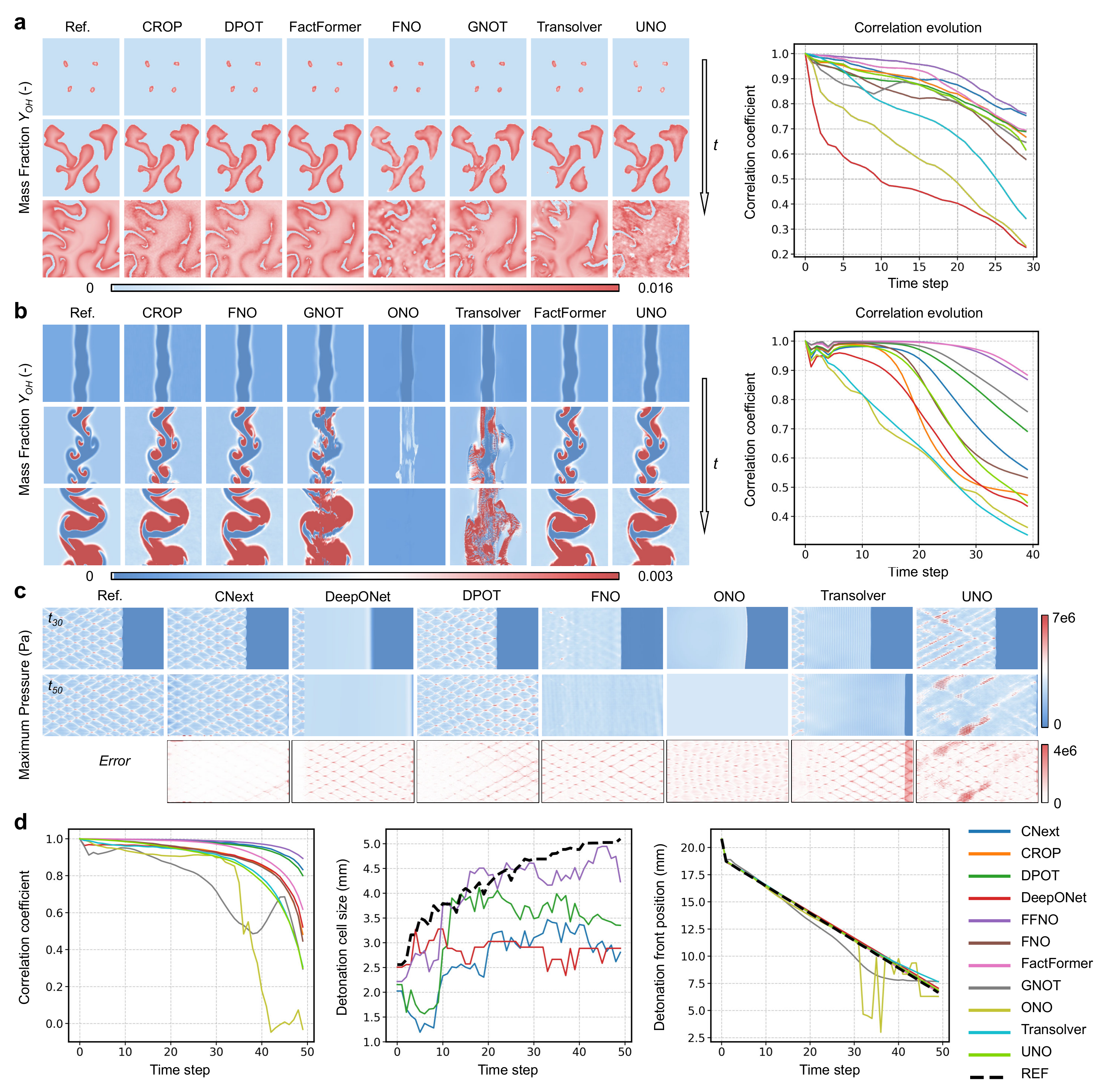}
   \caption{\textbf{2D regular cases: quantitative errors and visual comparisons.} \textbf{a}, \emph{IgnitHIT.} Left: snapshots of OH mass fraction $Y_{\mathrm{OH}}$ for the reference and the remaining surrogates; arrows indicate increasing time. Right: temporal evolution of the averaged correlation coefficients between the predicted fields and the reference. 
   \textbf{b}, \emph{EvolveJet.} Left: snapshots of $Y_{\mathrm{OH}}$ for the reference and the remaining surrogates. Right: temporal evolution of the averaged correlation coefficients between the predicted fields and the reference. 
   \textbf{c}, \emph{PlanarDet.} Maximum pressure fields $p_{\max}$ at representative times ($t_{30}$, $t_{50}$) for the reference and the surrogates, together with error maps at $t_{50}$ showing $\big|p_{\max}^{\mathrm{gt}}(t_{50}) - p_{\max}^{\mathrm{pred}}(t_{50})\big|$.
   \textbf{d}, \emph{PlanarDet.} Left: temporal evolution of the averaged correlation coefficients between the predicted fields and the reference. Middle: temporal evolution of the mean detonation cell size for the reference and the surrogates. Right: temporal evolution of the detonation front location for the reference and the surrogates.}
   \label{fig:extendfig1}
\end{figure}

\begin{figure}[t!]
  \centering
   \includegraphics[width=1\linewidth]{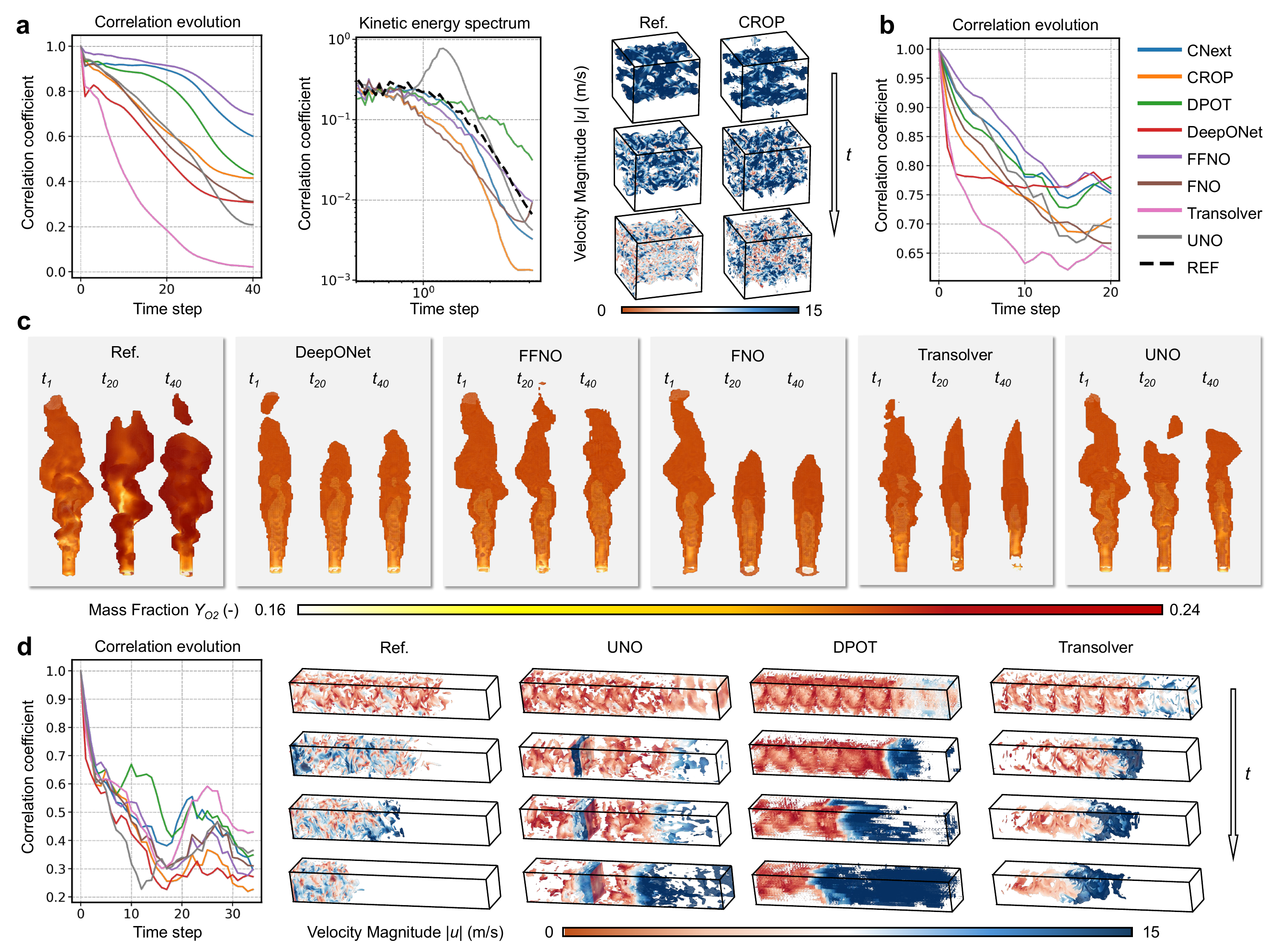}
   \caption{\textbf{3D regular cases: quantitative errors and visual comparisons.} \textbf{a}, \emph{ReactTGV.} Left: temporal evolution of the averaged cross-correlation coefficients and, at the final time, a comparison of the turbulent energy spectra between the predicted fields and the reference. Right: vorticity isosurfaces colored by velocity magnitude $|u|$ for the reference and for the predictions given by CROP; arrows indicate increasing time.
   \textbf{b}, \emph{PoolFire.} Temporal evolution of the averaged cross correlation coefficients between the predicted fields and the reference. 
   \textbf{c}, \emph{PoolFire.} Temperature isosurfaces at three representative times ($t_{1}, t_{20}, t_{40}$), colored by oxygen mass fraction $Y_{\mathrm{O_2}}$, for the reference and surrogates.
   \textbf{d}, \emph{PropHIT.} Left: temporal evolution of the averaged cross correlation coefficients between the predicted fields and the reference. Right: vorticity isosurfaces colored by velocity magnitude $|u|$ for the reference and surrogates.}
   \label{fig:extendfig2}
\end{figure}

\begin{figure}[t!]
  \centering
   \includegraphics[width=1\linewidth]{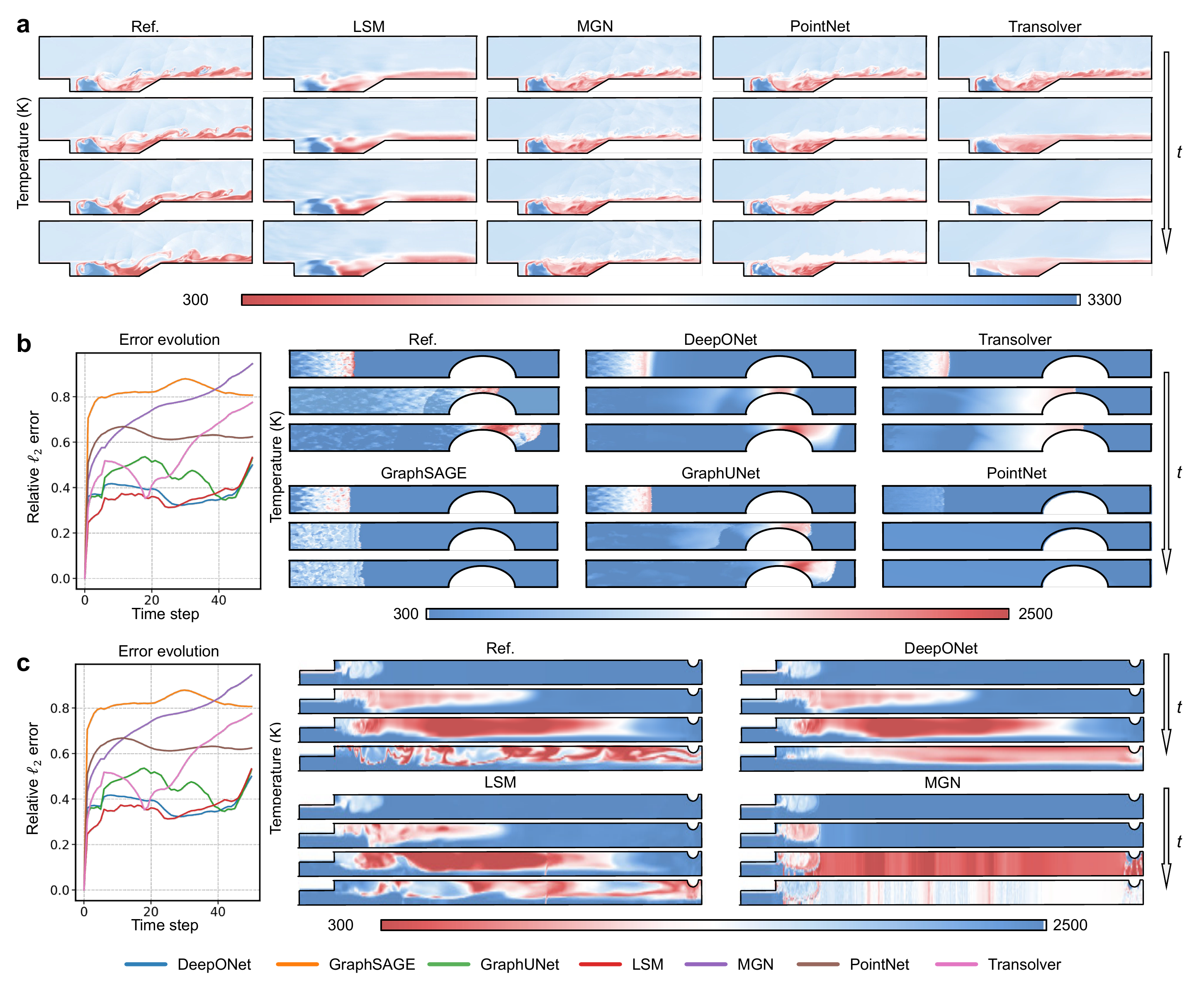}
   \caption{\textbf{irregular cases: quantitative errors and visual comparisons.} \textbf{a}, \emph{SupCavityFlame.} Snapshots of the temperature field for the reference and surrogates. 
   \textbf{b}, \emph{ObstacleDet.} Left: rollout error evolution (relative $\ell_{2}$). Right: snapshots of the temperature field for the reference and surrogates.
   \textbf{c}, \emph{SymmCoaxFlame.} Left: rollout error evolution (relative $\ell_{2}$). Right: snapshots of the temperature field for the reference and surrogates.}
   \label{fig:extendfig3}
\end{figure}

\clearpage

\renewcommand{\thefigure}{S.\arabic{figure}}
\renewcommand{\thetable}{S.\arabic{table}}

\renewcommand\thesection{\Alph{section}}
\renewcommand\thesubsection{\thesection.\arabic{subsection}}
\renewcommand\thesubsubsection{\thesubsection.\arabic{subsubsection}}

\noindent This supplement provides detailed, case-by-case descriptions of the datasets in the REALM benchmark, including their physical setups, numerical configurations, and released data. It also documents the complete training setup for all models, covering architectures, hyperparameters, and evaluation protocols.

\section{Data generation}

\subsection{Dataset summary}

Table~\ref{tab:data_summary} provides an overview of all datasets in REALM, including the raw simulation resolution, number of physical quantities, number of trajectories, and number of time steps. The last column reports the spatial and/or temporal downsampling factors applied when constructing the training sets; all other columns refer to the original high-resolution data.

\begin{table}[h!]
  \centering
  \small
  \caption{Summary of all datasets in REALM. “\# $N$” denotes the grid size, “\# $N_p$” the number of physical variables, “\# $N_t$” the number of time steps, and “\# $N_{\mathrm{traj}}$” the number of trajectories. Note that “\# $N$” and “\# $N_t$” correspond to the original high-resolution simulations. “Train downsampling” indicates the spatial and/or temporal subsampling factors used for training (T: time, S: space). “T$\times k$” denotes retaining every $k$-th snapshot in time; for structured grids, “S$\times k$” denotes retaining every $k$-th grid point per dimension, and for unstructured meshes it denotes retaining approximately one out of $k$ cells.}
  \label{tab:data_summary}
  \begin{tabular}{lcccccc}
    \toprule
    Case name & \# $N$ & \# $N_p$ & \# $N_{\mathrm{traj}}$ & \# $N_t$ & Train downsampling \\
    \midrule
    IgnitHIT$^{2d}_{\Box}$          & $128 \times 128$      & $12$ & $36$ & $30$ & T$\times 1$, S$\times 1$ \\
    EvolveJet$^{2d}_{\Box}$ & $256 \times 256$    & $39$ & $30$ & $40$ & T$\times 1$, S$\times 1$ \\
    ReactTGV$^{3d}_{\Box}$    & $256 \times 256 \times 256$      & $11$ & $16$ & $41$ & T$\times 1$, S$\times 8$ \\
    PropHIT$^{3d}_{\Box}$    & $129 \times 129 \times 1025$      & $13$ & $7$ & $35$ & T$\times 1$, S$\times 8$ \\
    PlanarDet$^{2d}_{\Box}$    & $832 \times 384$      & $13$ & $9$ & $50$ & T$\times 1$, S$\times 1$ \\
    SupCavityFlame$^{2d}_{\triangle}$    & $2988480$      & $12$ & $9$ & $101$ & T$\times 5$, S$\times 10$ \\
    ObstacleDet$^{2d}_{\triangle}$    & $8996482$      & $5$ & $6$ & $51$ & T$\times 1$, S$\times 29$ \\
    SymmCoaxFlame$^{2d}_{\triangle}$    & $294900$      & $21$ & $12$ & $36$ & T$\times 1$, S$\times 1$ \\
    MultiCoaxFlame$^{3d}_{\triangle}$    & $13478444$      & $22$ & $5$ & $91$ & T$\times 3$, S$\times 45$ \\
    PoolFire$^{3d}_{\Box}$    & $80 \times 80 \times 200$      & $9$ & $15$ & $101$ & T$\times 5$, S$\times 1$ \\
    FacadeFire$^{2d}_{\triangle}$    & $2524423$      & $9$ & $9$ & $101$ & T$\times 5$, S$\times 10$ \\
    \bottomrule
  \end{tabular}
\end{table}

\subsection{Case-by-case data generation}

In the following, we describe the data-generation procedure for each case in detail, including the physical configuration, numerical setup, and provided data. 

\subsubsection*{IgnitHIT$^{2d}_{\Box}$ (2D-regular mesh ignition kernel in homogeneous isotropic turbulence).}
\paragraph*{Physical description.}

\begin{figure}[t!]
  \centering
   \includegraphics[width=10cm]{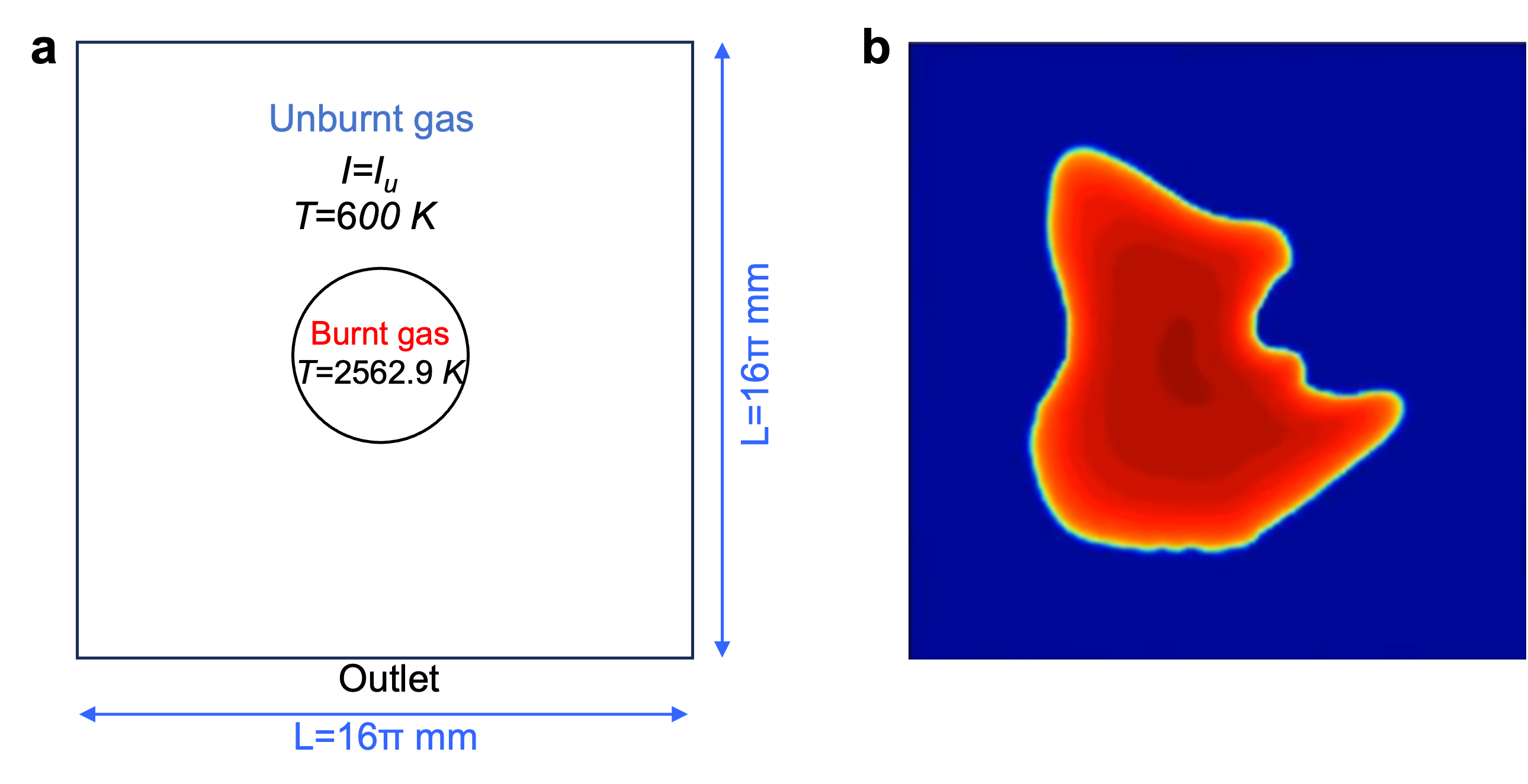}
   \caption{\textbf{IgnitHIT configuration and representative evolution.}
   \textbf{a}, Computational domain and initial condition.
   \textbf{b}, Representative temperature field at a later time.}
   \label{fig:1app}
\end{figure}

As shown in Fig.~\ref{fig:1app}, we consider a premixed, 2D \(\mathrm{H}_2/\mathrm{O}_2\) flame evolving in a square domain of size \(16\pi\times16\pi\,\mathrm{mm}^2\), discretized on a uniform \(1024\times1024\) grid. Initially, the domain is filled with quiescent unburnt gas at temperature \(T_u=600\,\mathrm{K}\), except for a circular ignition kernel at the domain center that contains fully burnt products at \(T_b=2562.9\,\mathrm{K}\). Both regions correspond to a stoichiometric mixture with equivalence ratio \(\phi=1\), but with different thermodynamic states. The velocity field is initialized with a homogeneous isotropic turbulent state generated in Fourier space and mapped to the computational grid, with prescribed rms intensity \(I_u\); this turbulence wrinkles the flame front and drives surface growth and localized flame-flow interactions. Zero gradient (Neumann) boundary conditions are applied to temperature and species mass fractions, while pressure and velocity use nonreflecting, wave transmissive boundaries so that large scale flow and reaction structures can exit the domain without artificial reflection.

\paragraph*{Simulation method.} 

Solutions are computed with \textit{DeepFlame}~\cite{mao2023deepflame,mao2024integrated,mao2025deepflame} on the native mesh. \textit{DeepFlame} is an open-source reactive-flow simulation platform widely used in the combustion community, built on a finite-volume formulation with implicit chemical integration and supporting high-fidelity simulations on both CPU and GPU architectures. The governing PDEs are advanced implicitly at each step, and stiff chemical source terms \(\dot{\omega}(T,Y)\) are integrated with a CVODE-based routine~\cite{cohen1994cvode}. The grid resolution resolves flame fronts and small-scale turbulent structures, so no subgrid model is used and the simulations are DNS. Transport properties follow a mixture-averaged model with temperature-dependent viscosity and species diffusivities. \emph{To avoid repetition, we refer to this configuration as the “DeepFlame-based implicit DNS setup” in subsequent cases.} Time integration uses a dynamic step with Courant number 0.5; snapshots are saved at a uniform cadence to yield 30 frames per trajectory on the native mesh. Chemistry follows a hydrogen–oxygen mechanism with 9 species and 23 elementary reactions~\cite{burke2012comprehensive}. The average compute cost is 60 core-hours per case, for a total of 2160 core-hours across the suite.

\paragraph*{Varied physical parameters.}
We construct a suite of 36 ignition-to-propagation trajectories by crossing three independent physical knobs that control flame initiation morphology and turbulent mixing. Specifically, the number of ignition kernels is varied as \(N_{\mathrm{k}}\in\{1,2,3,4\}\), the kernel geometry is selected from three canonical shapes (circle, square, triangular), and the turbulence intensity in the inflow is set to \(u'\in\{5,\,10,\,15\}\,\mathrm{m/s}\). Increasing \(N_{\mathrm{k}}\) modifies the initial flame surface area and thus the early heat release rate, while the kernel shape regulates curvature-induced burning variation. Adjusting \(u'\) tunes the characteristic eddy turnover scale and controls the strength of flame wrinkling and mixing. Crossing these factors yields systematically diverse transient flame evolution patterns under controlled parametric variation.

\paragraph*{Provided data.}
Each snapshot includes temperature \(T\), pressure \(p\), density \(\rho\), velocity components \((u,v)\), and species mass fractions \(\mathbf{Y}\) for the hydrogen-oxygen mechanism used in this case. The original data was downsampled on the $128\times128$ mesh, with new coordinates generated accordingly, and we provide a fixed trajectory-level split into training/validation/test sets with 26/5/5 trajectories, respectively.

\subsubsection*{EvolveJet$^{2d}_{\Box}$ (2D-regular mesh time-evolving shear jet flame).}

\begin{figure}[t!]
  \centering
   \includegraphics[width=13cm]{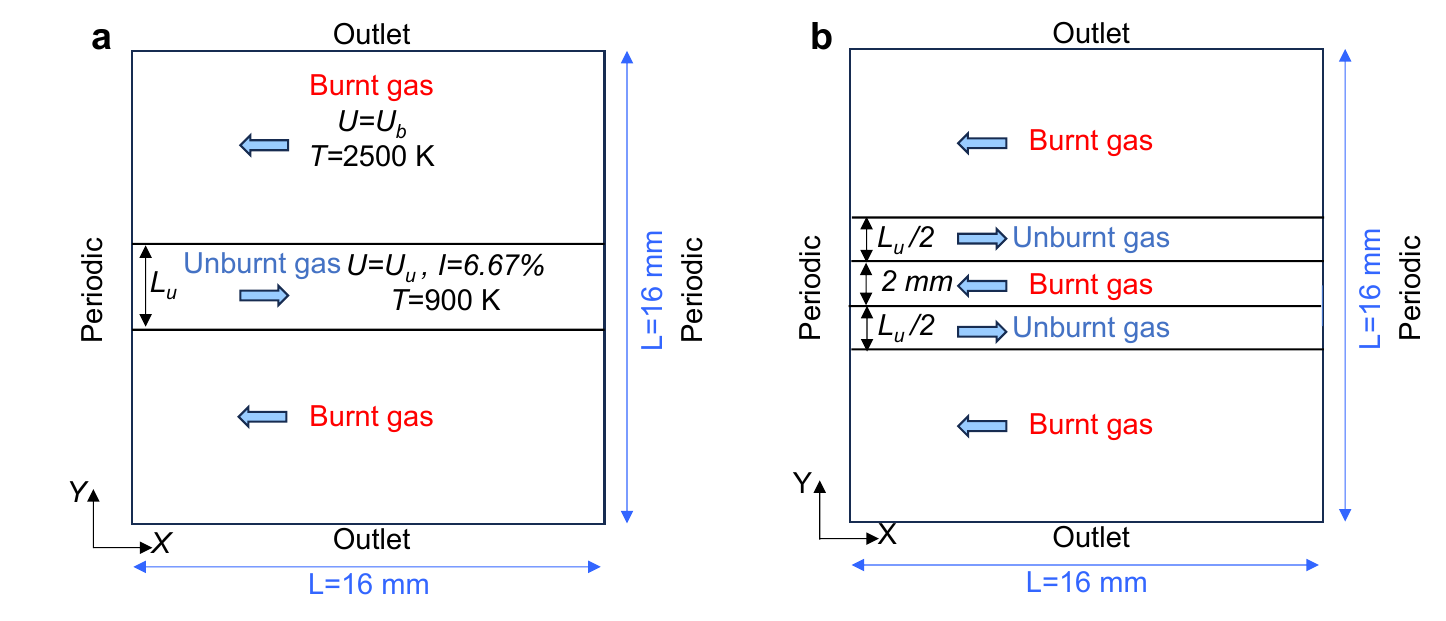}
   \caption{\textbf{Evolvejet configurations.}
   \textbf{a}, Single jet configuration.
   \textbf{b}, Dual jet variant.}
   \label{fig:2app}
\end{figure}

\paragraph*{Physical description.}
We consider two configurations of a 2D, shear driven premixed \(\mathrm{CH_4}/\mathrm{O_2}\) jet flame in a rectangular domain of size \(16\times16\,\mathrm{mm}^2\), discretized on a regular grid of \(800\times550\) cells. In both setups, the domain is periodic in the transverse (\(y\)) direction and open in the streamwise (\(x\)) direction, with nonreflecting outlet conditions at the left and right boundaries. In the single jet configuration (Fig.~\ref{fig:2app}\textbf{a}), a central horizontal layer of thickness \(L_u\) contains fresh, stoichiometric reactants (\(\phi=1\)) at \(T_u = 900\,\mathrm{K}\) convecting with mean velocity \(U_u\), while the regions above and below are filled with burnt products at \(T_b \approx 2500\,\mathrm{K}\) moving with velocity \(U_b\). The resulting velocity contrast across the interfaces generates strong shear layers in which the flame front wrinkles, merges, and convects downstream. In the dual jet variant (Fig.~\ref{fig:2app}\textbf{b}), the central layer is split into three sublayers: two unburnt streams of thickness \(L_u/2\) separated by a thin \(2\,\mathrm{mm}\) layer of burnt products. The unburnt streams and the burnt layer are assigned opposite streamwise velocities, producing a counter flowing shear configuration that further enhances mixing and flame-shear interactions.

\paragraph*{Simulation method.} 
Solutions are computed with the DeepFlame-based implicit DNS setup on the native mesh. Surrounding burned products are initialized from thermochemical equilibrium at the target pressure, and the central jet inflow is seeded with a precomputed turbulent field (turbulence intensity \(\approx 6.67\%\)) mapped to the jet band. Time integration uses a fixed step \(\Delta t=\langle5\times10^{-8}\rangle\,\mathrm{s}\); snapshots are saved at a uniform cadence to yield 40 frames per trajectory on the native mesh. Chemistry follows the methane-oxygen GRI-Mech 3.0 mechanism (53 species, 325 reactions)~\cite{gri30}. 

\paragraph*{Varied physical parameters.}
The suite contains 30 trajectories obtained by crossing three equivalence ratios \(\phi\in\{0.5,\,1.0,\,1.5\}\) with seven jet-flow configurations, parameterized by the jet-core speed \(U\) and the central-band thickness \(L\). The Reynolds number is defined as \(\mathrm{Re}=UL/\nu\) (SI units). Taking a representative \(\nu=1.5\times10^{-5}\,\mathrm{m^2\,s^{-1}}\), the resulting \(\mathrm{Re}\) for each \((U,L)\) is listed in table~\ref{tab:evojf_re}. Note that for three higher-\(\mathrm{Re}\) settings (31.5@1 mm, 27@2 mm, and 36@2 mm) we also include a dual-jet variant by splitting the central pipe into two \(1\,\mathrm{mm}\) streams separated by a \(2\,\mathrm{mm}\) fresh-mixture strip, which intensifies shear-layer interaction. Variation in \(\phi\) controls heat release and flame thickness, while variation in \(\mathrm{Re}\) sets shear intensity and turbulence level, yielding a broad range of physical conditions within one scenario.

\begin{table}[t!]
\centering
\caption{Jet-flow settings and Reynolds numbers for case \textit{EvolveJet}} 
\label{tab:evojf_re}
\begin{tabular}{llc}
\toprule
\(U\) (m/s) & \(L\) (mm) & \(\mathrm{Re}\) \\
\midrule
18.0 & 1 & \(1.20\times10^{3}\) \\
22.5 & 1 & \(1.50\times10^{3}\) \\
27.0 & 1 & \(1.80\times10^{3}\) \\
31.5 & 1 & \(2.10\times10^{3}\) \\
\midrule
18.0 & 2 & \(2.40\times10^{3}\) \\
27.0 & 2 & \(3.60\times10^{3}\) \\
36.0 & 2 & \(4.80\times10^{3}\) \\
\bottomrule
\end{tabular}
\end{table}

\paragraph*{Provided data.}
Each snapshot includes temperature \(T\), pressure \(p\), density \(\rho\), velocity components \((u,v)\), and species mass fractions \(\mathbf{Y}\) for the methane-oxygen mechanism used in this case. Data are released on the native \(800\times550\) mesh with corresponding spatial coordinates, together with a fixed trajectory-level split of 24/3/3 for training, validation, and test.

\subsubsection*{ReactTGV$^{3d}_{\Box}$ (3D-regular mesh reacting Taylor-Green vortex).}

\begin{figure}[t!]
  \centering
   \includegraphics[width=13cm]{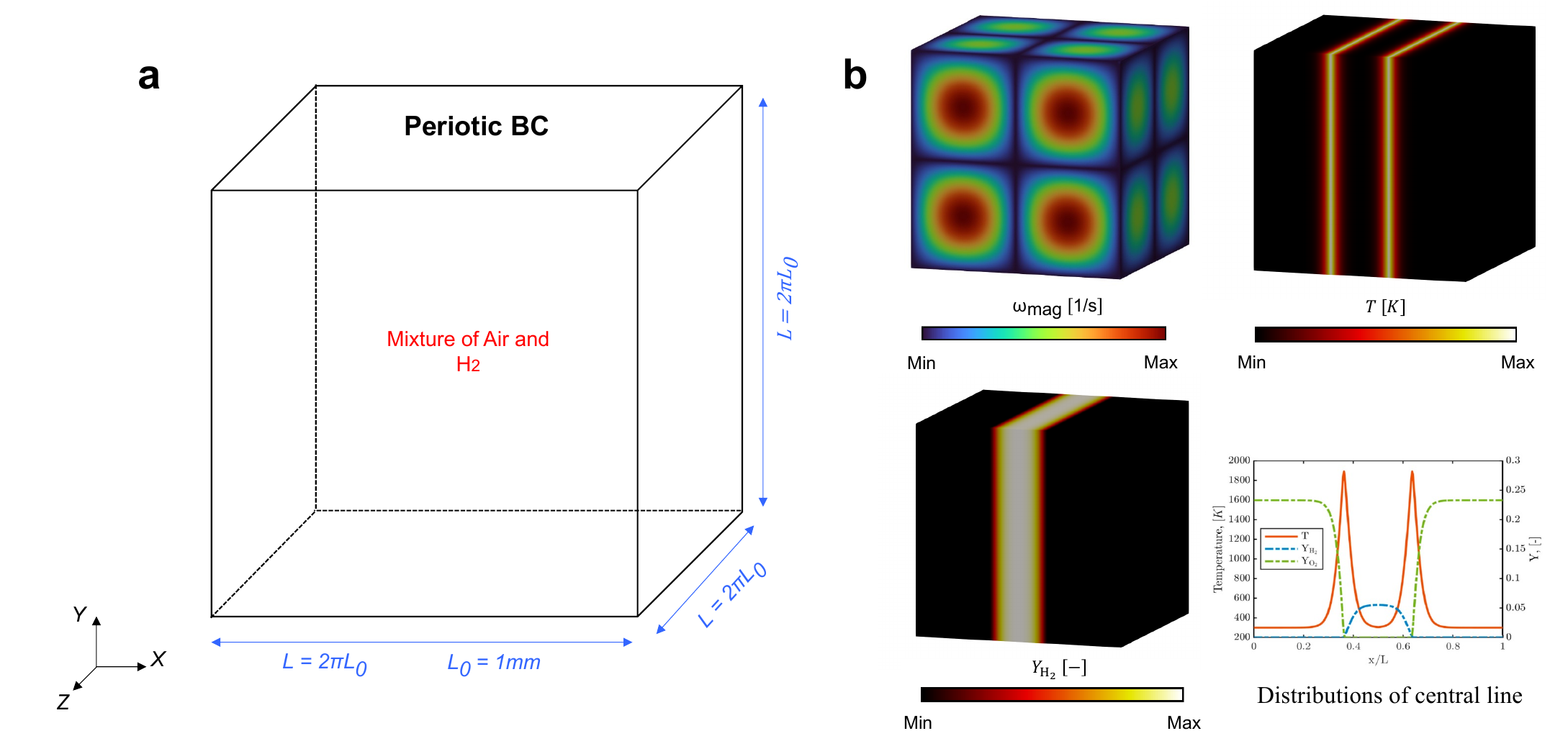}
   \caption{\textbf{ReactTGV computational setup.}
   \textbf{a}, ReactTGV configuration.
   \textbf{b}, Visualization of the initial fields setting.}
   \label{fig:3app}
\end{figure}

\paragraph*{Physical description.}

As shown in Fig.~\ref{fig:3app}, we consider a periodic cubic domain $\Omega=[0,L]^3$ discretized on a regular grid of $256\times256\times256$ cells, with side length $L=2\pi\,\mathrm{mm}$ and uniform spacing $\Delta x=L/256$. The fundamental wavenumber is $k=2\pi/L$. The initial velocity follows the Taylor-Green form:
\begin{equation}
u(\textbf{x},0)=U_0\sin(kx)\cos(ky)\cos(kz)\:,
\end{equation}
\begin{equation}
v(\textbf{x},0)=-U_0\cos(kx)\sin(ky)\cos(kz)\:,
\end{equation}
\begin{equation}
w(\textbf{x},0)=0\:,  
\end{equation}
with Reynolds number $\mathrm{Re}=U_0 L/\nu(T_0)$ based on the kinematic viscosity $\nu$ at the reference temperature $T_0=300\,\mathrm{K}$. The composition is non-premixed and blended by a smooth mask:
\begin{equation}
\psi(\textbf{x})=\tfrac12\Big[1+\tanh\!\Big(c\,\frac{R_d(\textbf{x})-R}{R}\Big)\Big],
\end{equation}
where $R_d(\textbf{x})$ is the distance to the prescribed mixing centerline or surface, $R$ sets the characteristic half-width, and $c$ controls interface sharpness (here fixed as $c=3$). Following the reference setup, we prescribe side values $Y_{\mathrm{H_2}}^{0}=0.0556$ on the fuel side and $Y_{\mathrm{O_2}}^{0}=0.233$ on the oxidizer side, with nitrogen complement elsewhere. The initial species fields are:
\begin{equation}
Y_{\mathrm{H_2}}(\textbf{x},0)=Y_{\mathrm{H_2}}^{0}\,[1-\psi(\textbf{x})]\:,
\end{equation}
\begin{equation}
Y_{\mathrm{O_2}}(\textbf{x},0)=Y_{\mathrm{O_2}}^{0}\,\psi(\textbf{x})\:,
\end{equation}
\begin{equation}
Y_{\mathrm{N_2}}(\textbf{x},0)=1-Y_{\mathrm{H_2}}(\textbf{x},0)-Y_{\mathrm{O_2}}(\textbf{x},0),
\end{equation}
with all other species initially zero, and the pressure uniform at $p_0$. After blending, each grid point is brought to a thermochemical equilibrium state by solving a local equilibrium at $T_0=300\,\mathrm{K}$ and $p_0$, yielding the final initial temperature and species fields used to start the simulation.

\paragraph*{Simulation method.}
Solutions are computed with the DeepFlame-based implicit DNS setup on the native mesh. The time step is fixed at \(\Delta t=10^{-6}\,\mathrm{s}\). Snapshots are written every 10 time steps (that is, every \(10\Delta t\)), and each trajectory contains 40 uniformly spaced frames on the native mesh. The reaction mechanism is a reduced H\(_2\)/air scheme with nine species and twelve reversible reactions~\cite{boivin2011explicit}.

\paragraph*{Varied physical parameters.}
We generate 16 trajectories by crossing two knobs that set the relative turbulence and mixing scales: \(U_0\in\{4,8,16,32\}\,\mathrm{m/s}\), corresponding to \(Re\approx\{1.7\times10^3,\,3.3\times10^3,\,6.7\times10^3,\,1.3\times10^4\}\), and the initial mixing half width \(R\in\{\tfrac{\pi}{4},\,\tfrac{3\pi}{8},\,\tfrac{\pi}{2},\,\tfrac{5\pi}{8}\}\,\mathrm{mm}\). Increasing \(U_0\) shortens the characteristic time scale and intensifies turbulence, while varying \(R\) changes the initial layer thickness.

\paragraph*{Provided data.}
Each snapshot includes temperature \(T\), pressure \(p\), density \(\rho\), velocity components \((u,v,w)\), and nine species mass fractions \(\mathbf{Y}\) (reduced \(\mathrm{H_2}\)-air mechanism). Data are provided on the native \(256\times256\times256\) mesh together with spatial coordinates and a fixed train/validation/test split of 12/2/2 by trajectory.

\subsubsection*{PropHIT$^{3d}_{\Box}$ (3D-regular mesh propagating flame in homogeneous isotropic turbulence).}

\begin{figure}[t!]
  \centering
   \includegraphics[width=11cm]{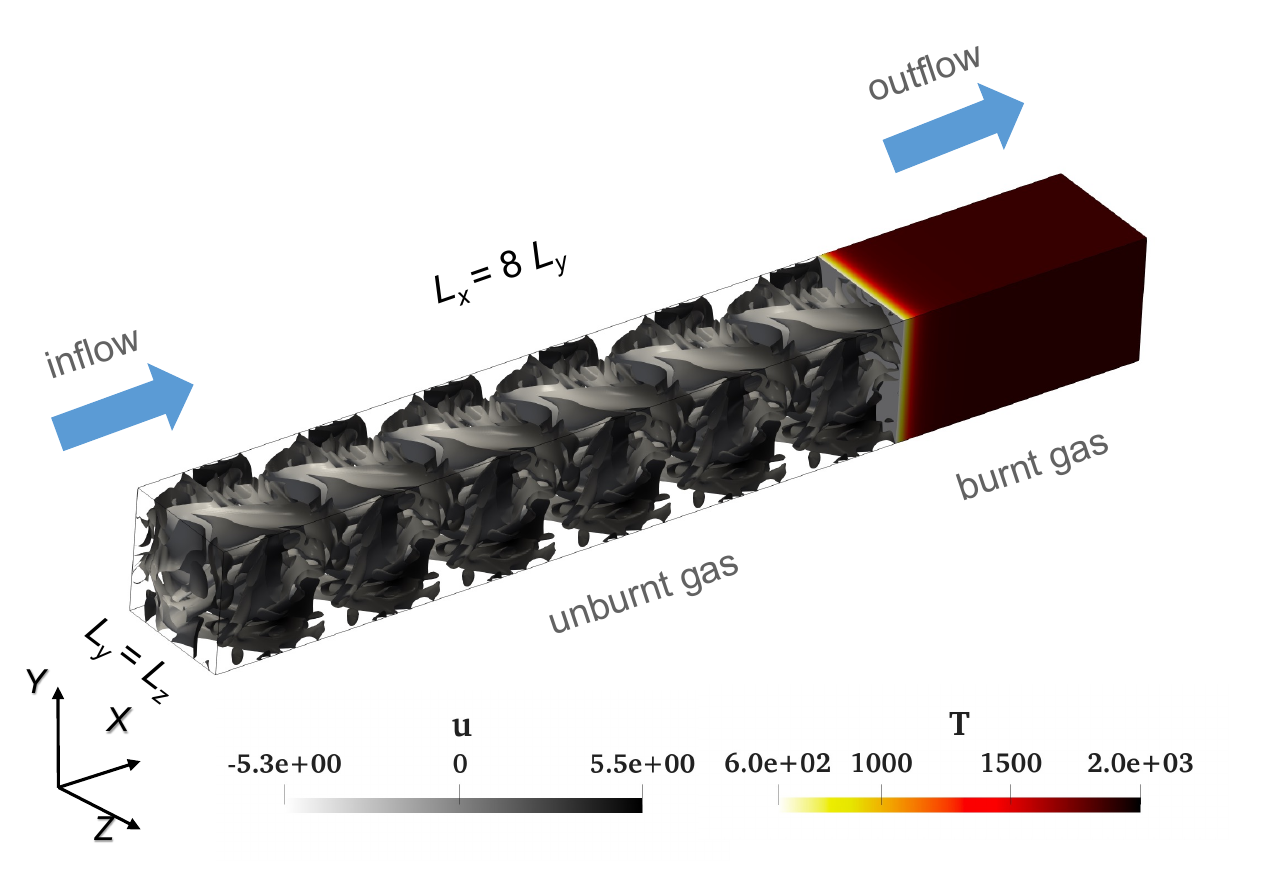}
   \caption{PropHIT computational configurations overlaid with initial velocity and temperature fields.}
   \label{fig:4app}
\end{figure}

\paragraph*{Physical description.}

As shown in Fig.~\ref{fig:4app}, the computational domain is a cuboid box with $L_x \times L_y \times L_z $. $L_y = L_z = 5.3 \delta_L$ (laminar flame thickness), $L_x = 8L_y$. This implies that when we choose different flame thicknesses (e.g., different pressures) for our cases, the computational domain size will change accordingly with the flame thickness variation. The domain is discretized on a uniform grid with $N_x\times N_y\times N_z=12N\times N\times N$ points; we fix $N=128$ for all runs, i.e., $1536\times128\times128$. For initialization, a one-dimensional laminar premixed flame is first computed and its state profiles are mapped onto the 3D box, placing the flame surface near the streamwise outlet; eight pre-evolved HIT boxes are then tiled along the $x$-direction to supply the desired fluctuating velocity field and shorten the transient to statistical stationarity. 

\paragraph*{Simulation method.} 
All simulations are performed with the in-house finite-difference code ASTR \cite{fang2019improved,fang2020turbulence}. Convective terms use a skew-symmetric formulation and diffusive terms a Laplacian form. Spatial derivatives are discretized on the uniform grid with the sixth-order explicit central scheme (ECS6)~\cite{fang2019improved}. Chemical source terms are integrated via a chemistry ODE solver through the Cantera Fortran interface. Time marching employs a three-stage, third-order TVD Runge-Kutta scheme. Numerical resolution is chosen to resolve both turbulence and flame scales, enforcing $k_{\max}\eta \ge 1.5$ for HIT (with $k_{\max}=\pi N/L$) and at least 24 grid points across $\delta_L$, where $k_{\max}=\pi N/L$ and $\eta$ is the Kolmogorov length scale.

\paragraph*{Varied physical parameters.}
We conduct 8 trajectories by crossing two operating pressures \(p\in\{2,\,5\}\,\mathrm{atm}\) with four turbulence-chemistry ratios \(u'/S_L\in\{2,\,5,\,10,\,20\}\). For each setting, the physical domain is tied to the laminar flame thickness \(\delta_L\) via \(L_y=L_z=5.3\,\delta_L\) and \(L_x=8L_y\), so the box size scales with pressure through \(\delta_L\) and \(S_L\) while the mesh is kept fixed.

\paragraph*{Provided data.}

Each snapshot includes temperature $T$, pressure $P$, density $\rho$, velocity components $(u,v,w)$, species mass fractions $\mathbf{Y}$ and heat release rate (HRR) in this case. We also provide a fixed trajectory-level split into training/validation/test sets with 5/1/1 trajectories, respectively.

\subsubsection*{PlanarDet$^{2d}_{\Box}$ (2D-regular mesh planar cellular detonation).}

\begin{figure}[t!]
  \centering
   \includegraphics[width=10cm]{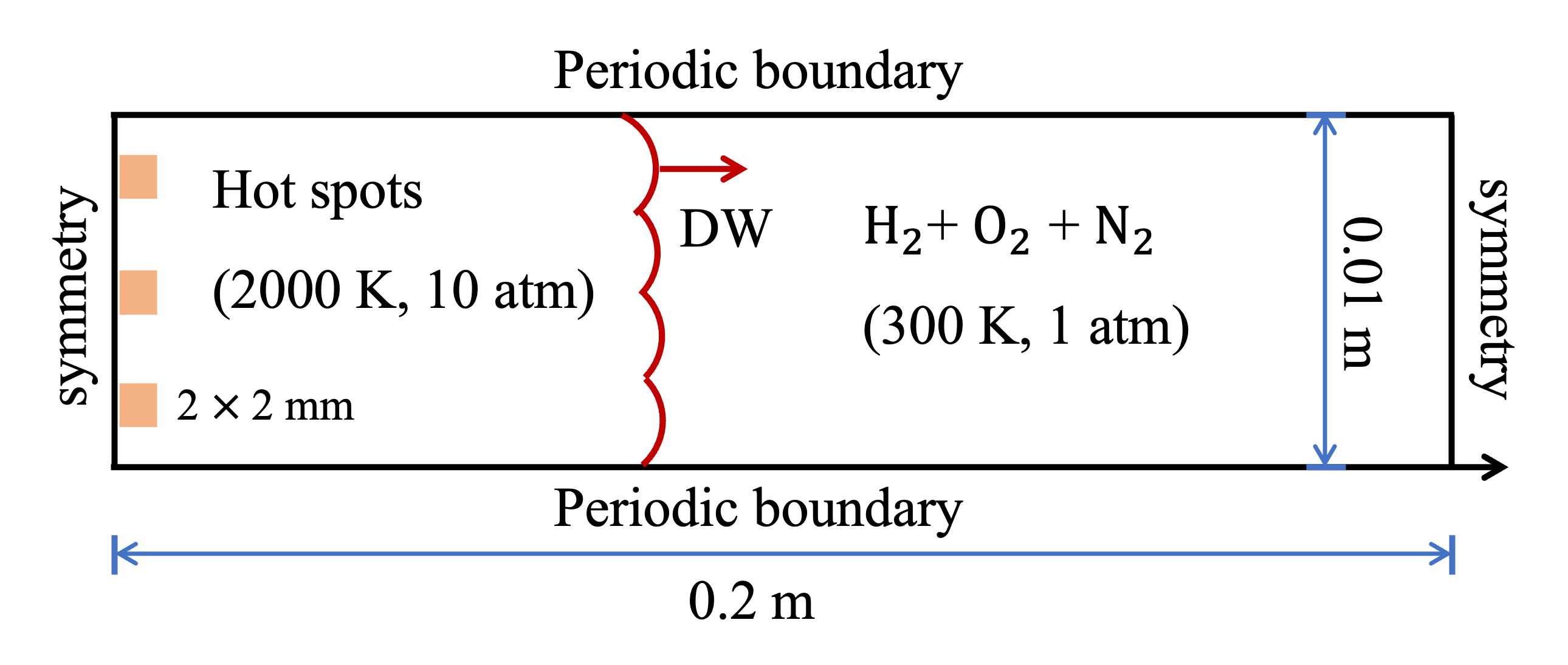}
   \caption{Schematic of the 2D planar detonation configuration.}
   \label{fig:5app}
\end{figure}

\paragraph*{Physical description.}
As shown in Fig.~\ref{fig:5app}, we consider a planar detonation propagating in a 2D channel of size \(0.2\times0.01\,\mathrm{m}^2\), discretized on a uniform Cartesian grid with cell size \(0.025\,\mathrm{mm}\), which has been verified to adequately resolve the main reaction zone. The domain is initially filled with quiescent premixed \(\mathrm{H_2/O_2/N_2}\) at a uniform pressure of \(P_0 = 100\,\mathrm{kPa}\). Near the left side of the domain, three compact high-temperature hot spots are imposed, which act as ignition kernels and rapidly merge into a planar detonation front that propagates in the positive \(x\) direction. Further downstream, the detonation develops a multi-shock interaction region where cellular structures become pronounced. Periodic boundary conditions are imposed on the upper and lower boundaries, while symmetry conditions are applied at the left and right boundaries.

\paragraph*{Simulation method.} Simulations are conducted by the high-speed reacting flow solver implemented within \textit{DeepFlame} and the \emph{“DeepFlame-based implicit DNS setup”} is configured. To initiate the detonation, three hot spots ( $T = 2000$ K and $P = 100$ atm) are introduced adjacent to the left boundary. Fixed timestep $\Delta t=1\times10^{-9}\ \mathrm{s}$ is adopted to ensure stability of simulation. The thermophysical properties and chemical reaction are modeled using a reduced mechanism for $\rm H_2$/Air combustion, consisting of 9 species and 19 reactions~\cite{jachimowski1988analytical}. After the detonation propagation reaches quasi-stable state, snapshots are saved at a uniform interval to present 100 frames per trajectory. The average compute cost to reach stable propagation is 5645 core-hours per case and the collation of snapshots cost another 1008 core-hours per case, for a total of 59877 core-hours across the suite.

\paragraph*{Varied physical parameters.}The suite contains 9 trajectories obtained by crossing three equivalence ratios ($\rm H_2:O_2:N_2 = x:1:7,\  x\in\{1.6,\,2,\,2.4\}$) and three initial temperatures $T_0\in\{290,\,300,\,330\}$ K. Variations in the initial conditions further modulate the stable detonation cell size, thereby broadening operating conditions. 

\paragraph*{Provided data.} Each snapshot includes temperature $T$, pressure $P$, maximum pressure $maxP$, density $\rho$, velocity components $(u,v)$, and species mass fractions $\mathbf{Y}$ for the hydrogen-air mechanism used in this case. Maximum pressure field is recorded to reveal the detonation cellular structure. To control data size, only the spatial domain swept by the detonation wave during the specified time interval is retained. Therefore, data are released on the native $840\times440$ mesh together with spatial coordinates. We also provide a fixed trajectory-level split into training/validation/test sets with 7/1/1 trajectories, respectively.

\subsubsection*{SupCavityFlame$^{2d}_{\triangle}$ (2D-irregular mesh supersonic cavity flame).}

\begin{figure}[t!]
  \centering
   \includegraphics[width=15cm]{ 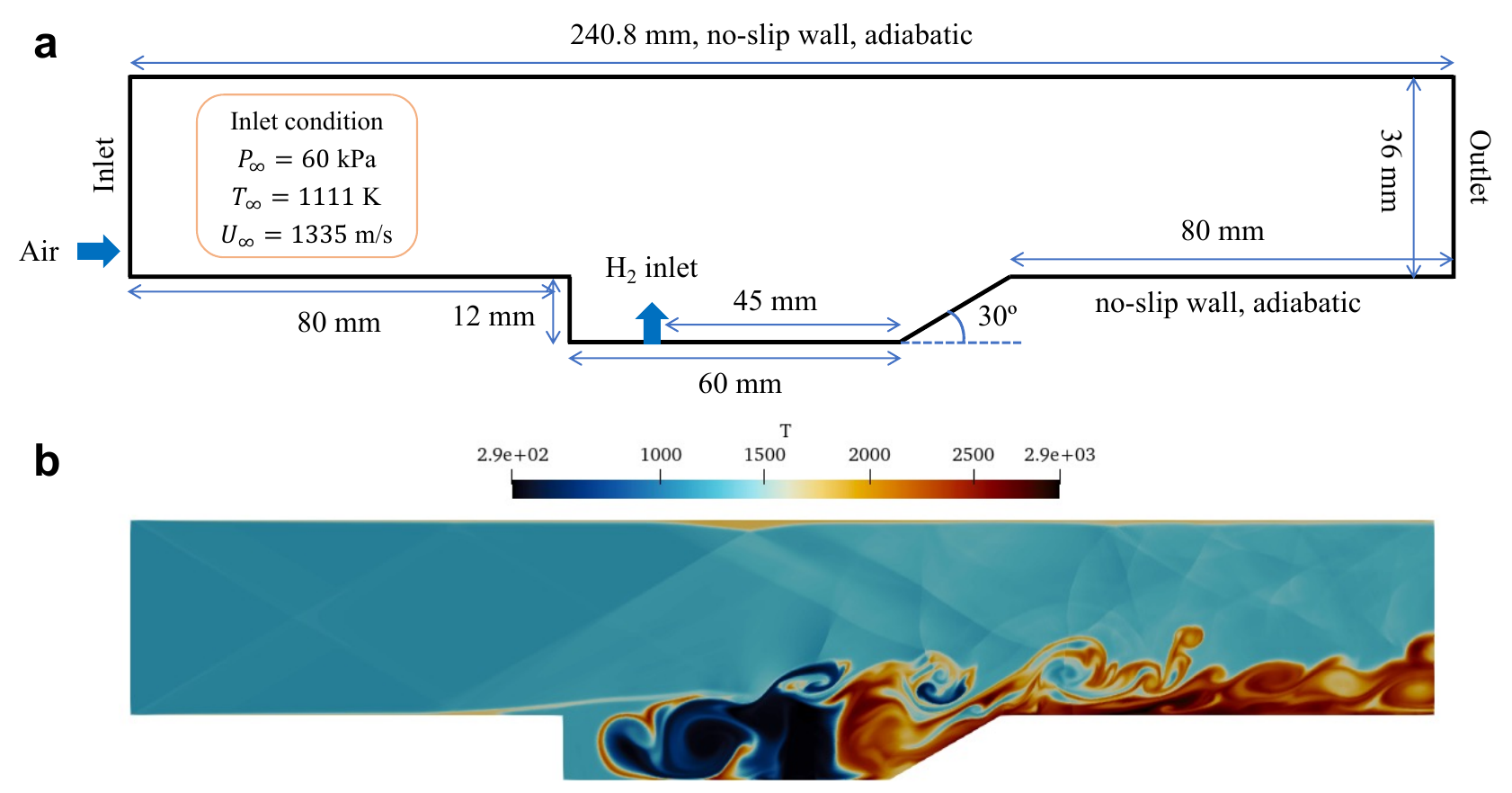}
   \caption{\textbf{Supersonic cavity flame configuration.} \textbf{a}, Schematic of the 2D cavity based combustor. \textbf{b}, Representative instantaneous temperature field.}
   \label{fig:6app}
\end{figure}

\paragraph*{Physical description.}
As shown in Fig.~\ref{fig:6app}, we consider a 2D supersonic ethanol spray combustion case in a cavity-based combustor. The computational domain extends to $240.8\times48\,\mathrm{mm}^2$, with mesh refinement in the cavity region ($1616\times240$ cells) and coarser grids elsewhere ($4816\times540$ cells), adopting dense spacing near walls to resolve boundary layers. The inflow corresponds to heated air at Mach $2.0$, with stagnation temperature $T_0=2000\,\mathrm{K}$, yielding a static temperature of $1111.1\,\mathrm{K}$, static pressure $p=60\,\mathrm{kPa}$, and flow velocity $u\approx1335\,\mathrm{m/s}$. Boundary conditions are specified as zero-gradient for pressure and temperature at walls and outlet, no-slip velocity at walls, and non-reflecting outlet conditions. The fuel is pure hydrogen, injected into the combustor through the nozzle at a fixed velocity. To improve convergence, a staged simulation strategy is adopted: (i) cold-flow simulation without injection, (ii) Flame simulations with hydrogen injection.

\paragraph*{Simulation method.} 
Simulations are conducted by the high-speed reacting flow solver implemented within \textit{DeepFlame} and configured with the \emph{“DeepFlame-based implicit DNS setup”}. The compressible multiphase PDEs are advanced implicitly at each step. Thermophysical properties are evaluated using Cantera with a reduced ethanol mechanism (9 species, 21 reactions)~\cite{millan2018multipurpose}, providing temperature-dependent transport coefficients and chemical rates. Time integration employs a constant step size $\Delta t=5e-09\,\mathrm{s}$, and results are sampled at a fixed cadence to provide trajectory statistics. The average compute cost is 9600 core-hours per case, with a total of 86400 core-hours for the campaign.

\paragraph*{Varied physical parameters.}
The dataset contains nine trajectories generated by the factorial combination of three injection locations (upstream of the cavity floor, downstream of the cavity floor, and on the cavity aft wall) and three injection velocities \(u_{\rm{inject}}\in\{50,\,150,\,300\}\).

\paragraph*{Provided data.}
Each snapshot includes temperature \(T\), pressure \(p\), density \(\rho\), velocity components \((u,v)\), and species mass fractions \(\mathbf{Y}\) for the hydrogen mechanism used in this case. We also provide a fixed trajectory-level split into training/validation/test sets with 7/1/1 trajectories, respectively.

\subsubsection*{ObstacleDet$^{2d}_{\triangle}$ (2D-irregular obstacle attenuated detonation).}

\begin{figure}[t!]
  \centering
   \includegraphics[width=10cm]{ 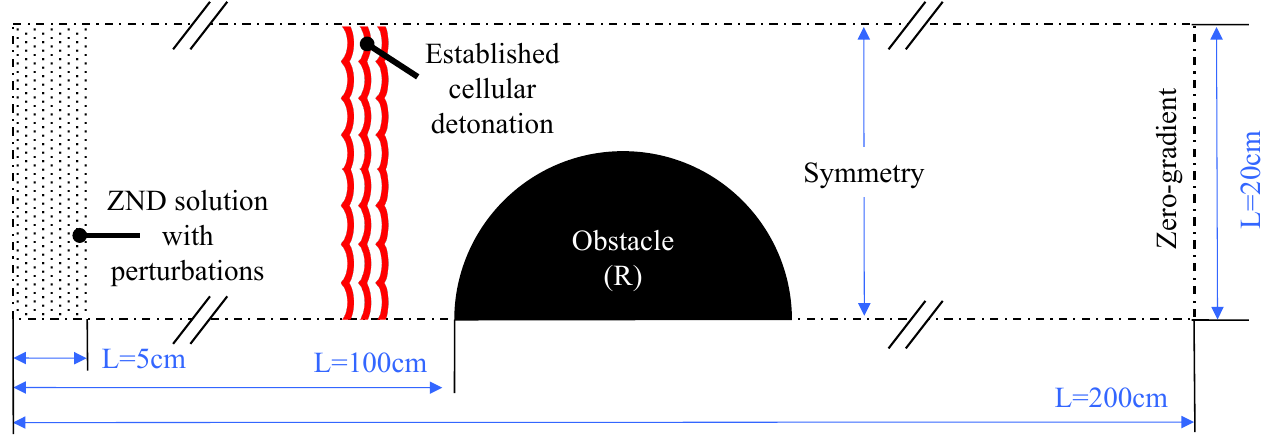}
   \caption{Schematic of the 2D obstacle detonation configuration.}
   \label{fig:7app}
\end{figure}

\paragraph*{Physical description.}
We simulate detonation attenuation and possible re initiation in a 2D channel with a semi cylindrical blockage. As shown in Fig.~\ref{fig:7app}, the computational domain is a straight duct of length \(L_x = 2~\mathrm{m}\) and height \(L_y = 20~\mathrm{cm}\). A semicircular obstacle of radius \(R\) is mounted on the bottom wall at \(x = 1~\mathrm{m}\), defining the blockage ratio \(\mathrm{BR}=2R/L_y\). The domain is filled with a quiescent premixed \(\mathrm{H_2}\)-air mixture. In the first \(L=5~\mathrm{cm}\) near the left boundary, we prescribe a one dimensional Zel’dovich–von
Neumann–Döring (ZND) detonation profile with small perturbations, which rapidly evolves into an established cellular detonation over the next \(L=100~\mathrm{cm}\) before it reaches the obstacle. The wave then propagates downstream and diffracts around the semi cylindrical blockage, leading to detonation attenuation and possible re initiation in the wake region. Symmetry boundary conditions are imposed at the top and bottom channel walls, the obstacle surface is treated as a no slip adiabatic wall, and a zero gradient, non reflecting outlet condition is applied at the right boundary.

\paragraph*{Simulation method.} 
We use an explicit compressible reactive-flow solver tailored for shock-flame problems. Spatial discretization employs a fifth-order weighted essentially non-oscillatory (WENO) scheme together with the Harten-Lax-van Leer with Contact (HLLC) approximate Riemann solver for robust shock capturing on the irregular mesh with the curved semicircular wall. Time integration is third-order strong-stability-preserving Runge-Kutta under a CFL constraint; the chemical source term is advanced with an operator-split explicit Arrhenius update in a calibrated single-step chemical-diffusive model~\cite{lu2021chemical}. Transport coefficients are temperature-dependent, and viscosity, species diffusivities, and thermal conductivity follow standard power-law correlations. Mesh resolution is chosen to resolve the induction/half-reaction zone and to capture cellular structures without excessive numerical diffusion; curved-wall elements are refined near the obstacle to maintain geometric fidelity. Snapshots are written at a uniform cadence on the native mesh, yielding 50 frames per trajectory with a fixed inter-frame interval of \(1.2\times10^{-5}\,\mathrm{s}\) across the six blockage-ratio/activation-energy conditions. Further numerical details in this case can be referred to as~\cite{lai2025reinitiation,houim2016role}.

\paragraph*{Varied physical parameters.}

The suite comprises six trajectories formed by crossing the blockage ratio
\(\mathrm{BR}\in\{0.76,\,0.78,\,0.80\}\) with the (nondimensional) effective activation energy \(\tilde{E}_a\in\{6,\,8\}\) in the chemical-diffusive model. Increasing \(\mathrm{BR}\) strengthens geometric diffraction and promotes decoupling of the lead shock and reaction zone. Raising \(\tilde{E}_a\) lengthens the induction zone and increases sensitivity to temperature and pressure, which raises the threshold for re-initiation. 

\paragraph*{Provided data.}
Each snapshot includes temperature \(T\), pressure \(p\), density \(\rho\), velocity components \((u,v)\), grid coordinates \((C_x,C_y)\), and species mass fractions \(\mathbf{Y}\). Two diagnostics are also provided: a Schlieren proxy (normalized \(\|\nabla \rho\|\)) and spanwise vorticity \(\omega_z\). We also provide a fixed trajectory-level split into training/validation/test sets with 4/1/1 trajectories, respectively.

\subsubsection*{SymmCoaxFlame$^{2d}_{\triangle}$ (2D-irregular coaxial symmetric rocket flame).}

\begin{figure}[t!]
  \centering
   \includegraphics[width=15cm]{ 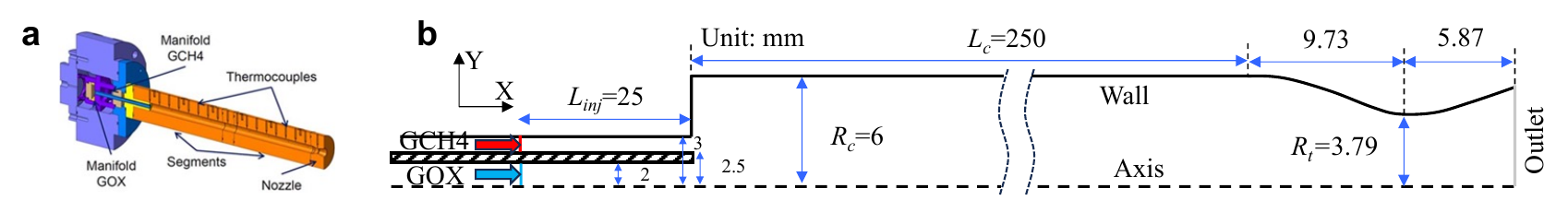}
   \caption{\textbf{Single-element rocket chamber and computational domain.} 
   \textbf{a}, Sketch of the single-element rocket combustor from TUM, showing the injector manifolds, chamber segments, and nozzle. 
   \textbf{b}, Two dimensional axisymmetric computational domain used in this work.}
   \label{fig:8app}
\end{figure}

\paragraph*{Physical description.}
The injector configuration considered here is based on the single-element rocket chamber from Technische Universität München (TUM) (cf. Fig.~\ref{fig:8app}\textbf{a})~\cite{roth2017experimental}. Exploiting the azimuthal symmetry of the geometry and flow, the configuration is reduced to a 2D axisymmetric computational domain, as shown in Fig.~\ref{fig:8app}\textbf{b}. The combustion chamber has an axial length of \(L_c = 250\,\mathrm{mm}\) and a radius of \(R_c = 6\,\mathrm{mm}\). Downstream, a Laval nozzle is attached, with a contraction section of length \(9.73\,\mathrm{mm}\), an expansion section of length \(5.87\,\mathrm{mm}\), and a throat radius \(R_t = 3.79\,\mathrm{mm}\). Upstream of the chamber, the injector tube is extended by \(L_{\mathrm{inj}} = 25\,\mathrm{mm}\) to allow the inflow to develop before entering the chamber. The mesh consists of 294{,}900 cells. Inside the chamber, the grid is uniformly spaced with \(\delta_x = \delta_y = 50\,\mu\mathrm{m}\); a mild axial stretching is applied toward the upstream injector tube and through the Laval nozzle to reduce the overall cell count while preserving resolution in regions of interest. Gaseous methane (GCH\(_4\)) and gaseous oxygen (GOX) are supplied through coaxial inlet boundaries on the left side. All solid walls are treated as adiabatic, no-slip boundaries. A non-reflecting outlet condition is imposed at the nozzle exit to minimize spurious reflections of acoustic and pressure waves. The total propellant mass flow rate \(\dot m_{\mathrm{total}} = \dot m_{\mathrm{CH_4}} + \dot m_{\mathrm{O_2}}\) is kept constant at \(0.062\,\mathrm{kg/s}\). To construct the database, the mixture ratio \(\mathrm{O/F} = \dot m_{\mathrm{O_2}}/\dot m_{\mathrm{CH_4}}\) is varied from 2.0 to 3.5, and the inlet temperatures of CH\(_4\) and O\(_2\) are swept from low-temperature conditions up to near-ambient values.

\paragraph*{Simulation method.} 

Simulations are conducted by the low-mach flow solver implemented within \textit{DeepFlame} and configured with the \emph{“DeepFlame-based implicit DNS setup”}. Adaptive timestep is adopted within the $Co_{max} \le 0.4$ to ensure the stability of the solution procedure, and the typical timestep during simulation is around $2\times 10^{-8}\mathrm{s}$. The thermophysical properties and chemical reaction are modeled using a reduced mechanism for high-pressure methane-oxygen combustion, consisting of 17 species and 44 reactions~\cite{monnier2022simulation}. Typically, the stable flame is established in the shear layer at 1.5 ms, then the solution snapshots are saved every $1\times 10^{-4}\mathrm{s}$ and a total number of 36 snapshots are collected for each trajectory. The average compute cost is 4224 core-hours per case, resulting a total of 50688 core-hours across the suite.

\paragraph*{Varied physical parameters.}

The suite comprises 12 trajectories formed by crossing four mixture ratios
\(O/F\in\{2.0,\,2.5,\,3.0,\,3.5\}\) with three inlet-temperature settings shown in
Table~\ref{tab:inlet-temps}. Changing \(O/F\) alters global stoichiometry and thus
shifts flame temperature, reaction-zone structure, and exhaust composition. Adjusting
the reactant inlet temperatures modifies the thermodynamic state of the propellants,
which affects ignition delay, stabilization, and overall efficiency. Together,
\((O/F,\;T_{\text{in}})\) span the operating envelope used to construct the dataset.

\begin{table}[t!]
  \centering
  \caption{Inlet temperature sets for methane and oxygen.}
  \label{tab:inlet-temps}
  \begin{tabular}{lcc}
    \toprule
    Temperature setting & \(T_{\mathrm{CH_4}}\;[\mathrm{K}]\) & \(T_{\mathrm{O_2}}\;[\mathrm{K}]\) \\
    \midrule
    Low      & 200 & 217 \\
    Medium   & 238 & 260 \\
    High     & 280 & 303 \\
    \bottomrule
  \end{tabular}
\end{table}

\paragraph*{Provided data.}
Each snapshot includes temperature $T$, pressure $P$, density $\rho$, velocity components $(u,v)$, and species mass fractions $\mathbf{Y}$ for the 17 species involved in the methane-oxygen reaction mechanism used in this case. The datasets are released on the native mesh with (294{,}900) cells, together with spatial coordinates and a fixed trajectory-level split of 8/2/2 for training, validation, and test.

\subsubsection*{MultiCoaxFlame$^{3d}_{\triangle}$ (3D-irregular multi-coaxial rocket flame).}

\begin{figure}[t!]
  \centering
   \includegraphics[width=10cm]{ 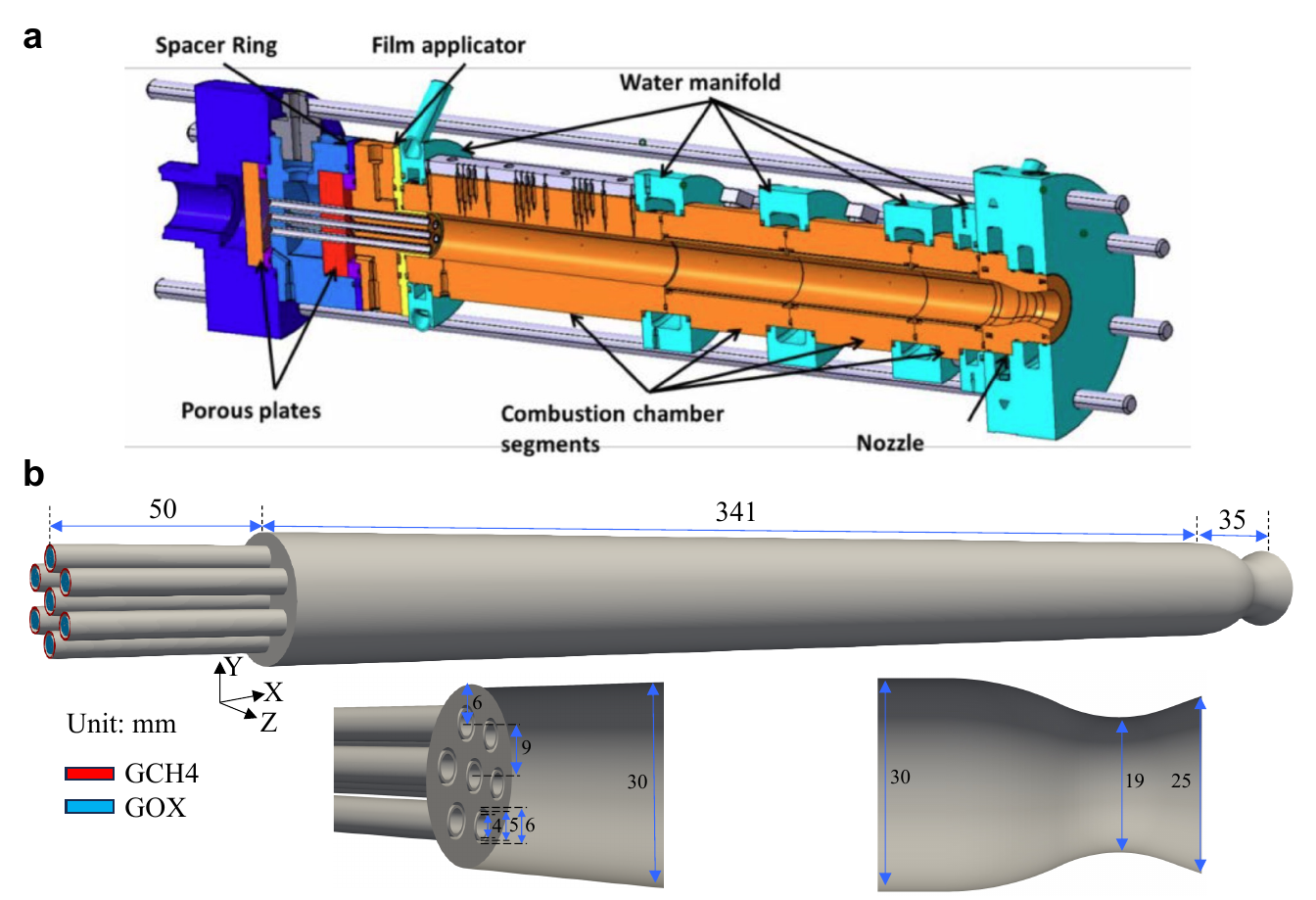}
   \caption{\textbf{Multi-element rocket combustor and computational domain.}
   \textbf{a}, Sketch of the experimental multi element shear coaxial rocket chamber from TUM.
   \textbf{b}, Three dimensional computational domain used in this work.
   }
   \label{fig:9app}
\end{figure}

\paragraph*{Physical description.}
The configuration follows the seven-element shear-coaxial rocket combustor from TUM (cf. Fig.~\ref{fig:9app})~\cite{silvestri2016characterization}. The combustor includes a faceplate with seven equally spaced injectors, a cylindrical chamber of length \(341~\mathrm{mm}\) and diameter \(30~\mathrm{mm}\), and a Laval nozzle of length \(35~\mathrm{mm}\) with contraction ratio \(2.5\). As shown in Fig.~\ref{fig:9app}, the full geometry comprising all injectors, the chamber, and the nozzle is resolved within a 3D irregular mesh containing \(13{,}478{,}444\) cells, of which \(127{,}800\) are prisms and the remainder hexahedra, with local refinement near injector lips, recirculation zones, and the nozzle throat to capture sharp gradients and acceleration. Gaseous oxygen and gaseous methane enter through separate inlets on the left boundary; all solid walls use no-slip adiabatic conditions; the nozzle exit uses a non-reflecting outlet to limit spurious acoustic reflections. Feed temperatures are \(T_{\mathrm{O_2}}=259.4~\mathrm{K}\) and \(T_{\mathrm{CH_4}}=237.6~\mathrm{K}\). Two global mixture ratios \(O/F\) are considered, and for each \(O/F\) the total propellant mass flow \(\dot m_{\mathrm{tot}}=\dot m_{\mathrm{O_2}}+\dot m_{\mathrm{CH_4}}\) is set to represent three thrust settings of 50\%, 75\%, and 100\%, producing six operating cases in total. A representative injector element uses a \(4.0~\mathrm{mm}\) oxygen post inner diameter, \(0.5~\mathrm{mm}\) oxygen post wall thickness, zero recess, and a \(6.0~\mathrm{mm}\) methane annulus hydraulic diameter (see Table~\ref{tab:tum_injector_geom}).

\begin{table}[t!]
  \centering
  \caption{Geometric parameters of a single shear-coaxial element on the TUM faceplate.}
  \label{tab:tum_injector_geom}
  \begin{tabular}{lc}
    \toprule
    {Parameter} & {Value} \\
    \midrule
    Oxygen post inner diameter (GOX core)   & 4.0 mm \\
    Oxygen post wall thickness              & 0.5 mm \\
    Oxygen post recess (setback)            & 0.0 mm \\
    Methane annulus hydraulic diameter      & 6.0 mm \\
    \bottomrule
  \end{tabular}
\end{table}

\paragraph*{Simulation method.} 
Simulations are conducted by the low-mach flow solver implemented within \textit{DeepFlame} and the \emph{“DeepFlame-based implicit DNS setup”} is configured. Adaptive timestep is adopted within the $Co_{max} \le 0.6$ to ensure the stability of the solution procedure, and the typical timestep during simulation is around $5\times 10^{-8}\mathrm{s}$. The thermophysical properties and chemical reaction are modeled using a reduced mechanism for high-pressure methane-oxygen combustion, consisting of 17 species and 44 reactions~\cite{monnier2022simulation}. Typically, the stable flame is established in the shear layer at 1.0 ms, then the solution snapshots are saved every $1\times 10^{-4}\mathrm{s}$ and a total number of 90 snapshots are collected for each trajectory. The average compute cost is 67.78 million core-hours per case, resulting a total of 406 million core-hours across the suite.

\paragraph*{Varied physical parameters.}

The suite contains six trajectories obtained by crossing two mixture ratios \((O/F\in\{2.64,\,3.40\})\) with three thrust levels (summarized in Table~\ref{tab:of_thrust}). Varying \(O/F\) changes the global stoichiometry and thereby influences flame temperature, completeness of combustion, and exhaust composition. Adjusting the thrust level modifies the total propellant mass flow rate, which in turn affects chamber pressure, flow velocity, and the degree of turbulent mixing. Together, \(O/F\) and thrust govern the combustor’s thermal environment, combustion efficiency, and dynamic response, and thus define the operating envelope considered here.

\begin{table}[t!]
  \centering
  \caption{Propellant mass flow rates for the two \(O/F\) settings and three thrust levels. Mass flow rates are in kg\,s\(^{-1}\).}
  \label{tab:of_thrust}
  \begin{tabular}{lcccc}
    \toprule
    Thrust level & \(O/F\) & \(\dot m_{\mathrm{CH_4}}\) & \(\dot m_{\mathrm{O_2}}\) & \(\dot m_{\mathrm{tot}}\) \\
    \midrule
    50\%  & 2.64 & 0.04000 & 0.10550 & 0.14550 \\
    75\%  & 2.64 & 0.06000 & 0.15825 & 0.21825 \\
    100\% & 2.64 & 0.08000 & 0.21100 & 0.29100 \\
    \midrule
    50\%  & 3.40 & 0.03307 & 0.11243 & 0.14550 \\
    75\%  & 3.40 & 0.04960 & 0.16865 & 0.21825 \\
    100\% & 3.40 & 0.06610 & 0.22490 & 0.29100 \\
    \bottomrule
  \end{tabular}
\end{table}

\paragraph*{Provided data.}

Each snapshot includes temperature $T$, pressure $P$, density $\rho$, velocity components $(u,v,w)$, and species mass fractions $\mathbf{Y}$ for the 17 species involved in the methane-oxygen reaction mechanism used in this case. The datasets are released on the native mesh with 13{,}478{,}444 cells, together with spatial coordinates and a fixed trajectory-level split of 3/1/1 for training, validation, and test.

\subsubsection*{PoolFire$^{3d}_{\Box}$ (3D-regular pool fire).}

\begin{figure}[t!]
  \centering
   \includegraphics[width=8cm]{ 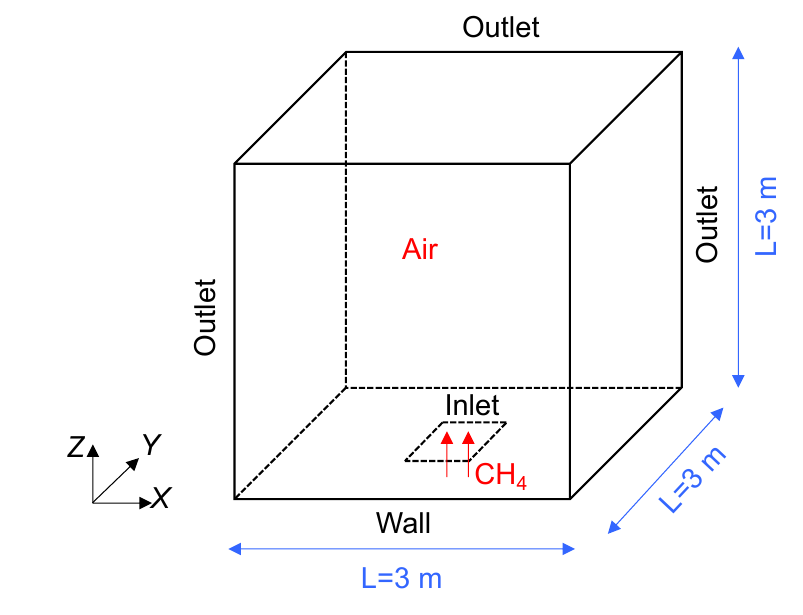}
   \caption{Schematic of the cubical pool-fire configuration.
   }
   \label{fig:10app}
\end{figure}

\paragraph*{Physical description.}
As shown in Fig.~\ref{fig:10app}, the computational domain is a cubic box with dimensions of $3 m \times 3 m \times 3 m$. The bottom of the computational domain is a solid wall, while the sides and the top are open boundaries. At the center of the bottom wall, an inlet is defined through which methane is injected into the domain. At this inlet we prescribe a specified mass-flow condition: the target fuel mass flow rate is set, and the solver consistently accounts for both the advective and diffusive parts of the mass and energy fluxes. The inlet mass flow rate is determined based on the specified heat release rate. For heat release rates of 14, 22, 33, 45, and 58 kW, the corresponding mass flow rates are 0.288, 0.434, 0.660, 0.897, and 1.149 g/s, respectively. The base grid is discretized  with 30 cells in each direction and \textit{snappyHexMesh}, the automatic meshing tool in \textit{OpenFOAM}, is applied for mesh refinement by dividing the computational domain into four regions and refining each individually. For inlet sizes of $0.1 \times 0.1$, $0.3 \times 0.3$, and $0.5 \times 0.5$ m, the total number of cells are 79.3 K, 390.7 K, and 1024.6 K, respectively.

\paragraph*{Simulation method.} 
The simulations are performed using \textit{FireFOAM}, an open source fire simulation code~\cite{WANG20112473} based on the \textit{OpenFOAM} platform. \textit{FireFOAM} is a pressure-based solver in which the discretized Navier-Stokes equations are solved using the finite volume method. The turbulent combustion is modeled using the EddyDissipation Model with a single-step reaction mechanism, under the assumption that chemical reactions occur much faster than the turbulent mixing processes. Large eddy simulation is employed for turbulence, with subgrid-scale stresses represented by the one-equation model, which solves a sub-grid scale kinetic energy equation. Transport properties are modeled using Sutherland’s law for temperature-dependent viscosity, while equal species diffusivities and a unity Lewis number are assumed. Radiative heat transfer is calculated using the Finite Volume Discrete Ordinates Method, with a constant radiant fraction of 20\% adopted for the radiation source term. Time integration is carried out with an adaptive timestep of approximately 0.001 s, constrained by a maximum Courant number of 0.9. Flow field snapshots are saved every 0.02 s for a duration of 2 s after the flow reaches statistical stationary. The average computational cost is 28, 240 and 780 core-hours per case for inlet sizes of $0.1 \times 0.1$, $0.3 \times 0.3$, and $0.5 \times 0.5$ m, respectively, amounting to a total of 5240 core-hours for the full set of simulations.

\paragraph*{Varied physical parameters.}
The suite contains 15 trajectories obtained by crossing five heat release rates (14, 22, 33, 45 and 58 kW) with three inlet sizes ($0.1 \times 0.1$, $0.3 \times 0.3$, and $0.5 \times 0.5$ m). Variations in heat release rate and inlet size significantly affect fire characteristics, including flame height and flame width. We also provide a fixed trajectory-level split into training/validation/test sets with 11/2/2 trajectories, respectively.

\paragraph*{Provided data.}
Each snapshot includes temperature $T$, pressure $P$, density $\rho$, velocity components $(u,v,w)$, and species mass fractions $\mathbf{Y}$ for the single-step methane-air reaction mechanism used in this case. Before sampling, the original flow field is first mapped onto a uniform grid with a resolution of 0.015 m in all directions, and then cropped to retain only the region near the flame, yielding a grid of $80 \times 80 \times 200$ cells in the $x$, $y$, and $z$ directions, respectively.

\subsubsection*{FacadeFire$^{3d}_{\triangle}$ (3D-irregular building facade fire).}

\begin{figure}[t!]
  \centering
   \includegraphics[width=15cm]{ 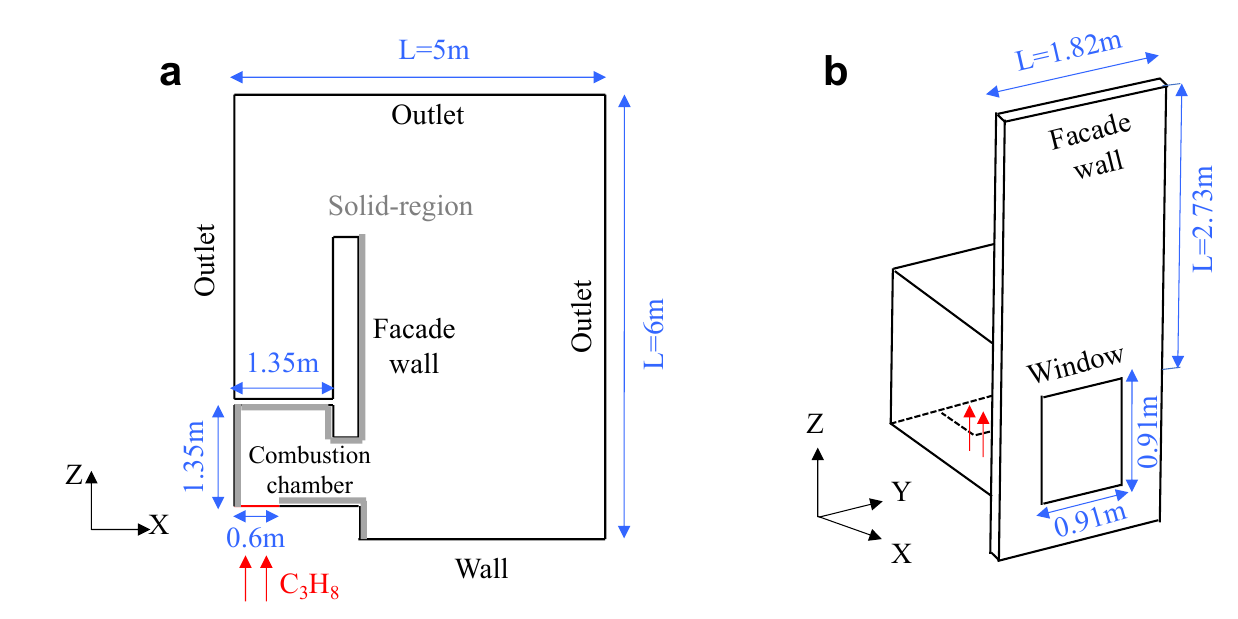}
   \caption{\textbf{Facade fire configuration.}
   \textbf{a}, Side view of the coupled combustion chamber and facade wall in the \(x\text{-}z\) plane. 
   \textbf{b}, Three dimensional view of the facade wall with window opening and the attached chamber; red arrows mark the propane inlet on the chamber floor.
   }
   \label{fig:11app}
\end{figure}

\paragraph*{Physical description.}
As shown in Fig.~\ref{fig:11app}, the computational domain is a rectangular volume of \(5\,\mathrm{m}\times6\,\mathrm{m}\times6\,\mathrm{m}\), comprising two main parts: a cubic combustion chamber with a window opening and an adjacent vertical facade. The chamber interior measures \(1.35\,\mathrm{m}\times1.35\,\mathrm{m}\times1.35\,\mathrm{m}\); the facade panel is \(4.095\,\mathrm{m}\times1.82\,\mathrm{m}\). A square window of \(0.91\,\mathrm{m}\times0.91\,\mathrm{m}\) is located at floor level on the chamber wall. Propane is supplied through a floor inlet of \(0.60\,\mathrm{m}\times0.60\,\mathrm{m}\) adjacent to the rear wall. A base gas-phase grid of \(50\times60\times60\) cells in \((x,y,z)\) is locally refined around the window jet, plume core, and near-wall regions, yielding \(\sim2.67\times10^{6}\) gas cells in total. Solid regions are meshed by extruding the gas-solid interface outward along the surface normal by 20 layers to represent a \(25\,\mathrm{mm}\) ceramic fiber blanket lining the chamber and facade surfaces. At the fuel inlet we impose a specified mass-flow boundary condition (i.e., the inflow velocity and composition are set to deliver the target fuel mass flow; both advective and diffusive contributions are included), using the flow rates corresponding to the representative heat-release levels in Table~\ref{tab:mass flow rate}. The domain bottom is a no-slip wall, while the top and lateral boundaries are open (pressure outlet) to allow natural convection and plume venting.

\begin{table}[t!]
    \centering
    \caption{The heat release rates and corresponding fuel mass flow rates}
    \label{tab:mass flow rate}
    \begin{tabular}{cccccccccc} 
        \toprule
        Heat release rate (kW) & 400 & 500 & 600 & 700 & 800 & 900 & 1000 & 1100 & 1200 \\ 
        \midrule
        Mass flow rate (g/s) & 8.63 & 10.79 & 12.94 & 15.10 & 17.26 & 19.42 & 21.57 & 23.73 & 25.87 \\ 
        \bottomrule
    \end{tabular}
\end{table}

\paragraph*{Simulation method.} 
The simulations are also conducted using \textit{FireFOAM} with Large eddy simulation approach (LES). Turbulent combustion is modeled with a single-step propane-air reaction mechanism using the EddyDissipation Model. Subgrid‑scale turbulence in LES is closed with the dynamic Lagrangian model. Transport properties are evaluated using Sutherland’s law under the assumptions of equal species diffusivities and unity Lewis number. Radiative heat transfer is treated with the Finite Volume Discrete Ordinates Method (FV-DOM) using \texttt{constRadFractionEmission}, which applies a constant radiant fraction (here \(33\%\)) of the local chemical heat release as a radiation source term. Time integration employs an adaptive timestep of approximately $\Delta t=2.5\times10^{-4}\ \mathrm{s}$, constrained by a maximum Courant number of 0.9. Flow field snapshots are saved every 0.02 s for a duration of 2 s after the facade temperature distribution reaches steady state. The 900 kW case is initialized from quiescent conditions at ambient temperature and simulated for 270 s to obtain a stabilized window‑ejected fire plume, requiring about 172,000 core‑hours. The remaining cases are restarted from the stable flow field of the 900 kW case, with each simulation consuming on average 18,000 core‑hours. The total computational cost for the complete suite of simulations is approximately 316,000 core-hours.

\paragraph*{Varied physical parameters.}
The suite comprises 9 trajectories, obtained by varying the heat release rate from 400 kW to 1200 kW in increments of 100 kW. Variations in heat release rate influence the external flame behavior, leading to differences in heat flux and temperature distributions along the façade wall.

\paragraph*{Provided data.}
Sampling is carried out on an irregular mesh. Each snapshot includes temperature $T$, pressure $P$, density $\rho$, velocity components $(u,v,w)$, and species mass fractions $\mathbf{Y}$ for the single-step methane-air reaction mechanism used in this case. We also provide a fixed trajectory-level split into training/validation/test sets with 7/1/1 trajectories, respectively.

\section{Hyperparameter validation and ablations}
\subsection{Selection of structural model parameters}

To ensure fair and reproducible comparison, we adopt a unified hyperparameter-selection strategy across all benchmarks. Instead of case-specific tuning for each neural operator, every model family is evaluated under a small set of \emph{capacity-aligned} presets at three representative scales, S, M, and L, corresponding roughly to $\mathcal{O}(1\text{M})$, $\mathcal{O}(10\text{-}50\text{M})$, and $\mathcal{O}(100\text{-}1000\text{M})$ trainable parameters, respectively. While transformer-style operators typically realize fewer \emph{parameters} at a given tier owing to their substantially larger \emph{activation memory} during self-attention. The concrete configurations for each tier are listed in Tables~\ref{tab:model_size_presets} and~\ref{tab:unstructured_presets}. For regular-grid cases, we evaluate convolutional (CNext), spectral (FNO, FFNO, UNO, CROP, DPOT), transformer-style (FactFormer, Transolver, ONO, GNOT), and pointwise (DeepONet) operators. For irregular-grid cases, we evaluate point-cloud and mesh/graph operators (PointNet, LSM, MGN, GraphUNet, GraphSAGE), along with DeepONet and Transolver under mesh-projected variants. 

Note that no case-specific architectural modifications are introduced; only general structural hyperparameters vary across families in Tables~\ref{tab:model_size_presets} and~\ref{tab:unstructured_presets}: \textbf{(i) width}, the base hidden channel dimension that seeds feature sizes across blocks; \textbf{(ii) layers}, the stacked block depth (or encoder-decoder scale depth for U-shaped operators); \textbf{(iii) modes}, the number of retained Fourier modes per spatial axis for spectral operators; and \textbf{(iv) heads}, the number of attention heads for transformer-style architectures. Not every model instantiates every knob (e.g., convolutional operators do not use \texttt{modes}, and pointwise/graph operators do not use \texttt{heads}); the applicability is summarized directly in the tables. This capacity-aligned design lets us compare architectures at matched representational strength while preserving each family’s intrinsic inductive bias, rather than conflating results with depth/width or spectral-resolution advantages.

\begin{table}[t!]
\centering
\small
\setlength{\tabcolsep}{5pt}
\caption{\textbf{Model-size presets (regular-grid cases only)}. For each family we define S/M/L by capacity; parameter counts (Params) are approximate (millions). \emph{Note:} $^\P$ For \textbf{DeepONet}, \texttt{modes} denotes \texttt{dotDim}.}
\label{tab:model_size_presets}
\begin{tabular}{lllccccr}
\toprule
Family & Dim & Size & width & layers & modes & heads & Params (M) \\
\midrule
\multirow{3}{*}{CNext} & \multirow{3}{*}{2D/3D} & S & 32  & 2 & -- & -- & 0.459 \\
                       &                         & M & 64  & 4 & -- & -- & 28.825 \\
                       &                         & L & 128 & 6 & -- & -- & 1835.948 \\
\midrule
\multirow{3}{*}{FNO}   & \multirow{3}{*}{2D/3D} & S & 64  & -- & 16 & -- & 8.408 \\
                       &                         & M & 128 & -- & 32 & -- & 134.271 \\
                       &                         & L & 256 & -- & 48 & -- & 1208.129 \\
\midrule
\multirow{3}{*}{CROP}  & \multirow{3}{*}{2D/3D} & S & 32  & -- & 16 & -- & 3.955 \\
                       &                         & M & 64  & -- & 32 & -- & 65.099 \\
                       &                         & L & 128 & -- & 48 & -- & 591.735 \\
\midrule
\multirow{3}{*}{UNO}   & \multirow{3}{*}{2D/3D} & S & 10  & -- & 16 & -- & 6.064 \\
                       &                         & M & 32  & -- & 32 & -- & 84.050 \\
                       &                         & L & 64  & -- & 48 & -- & 378.082 \\
\midrule
\multirow{3}{*}{FFNO}  & \multirow{3}{*}{2D/3D} & S & 64  & 2 & 16 & -- & 0.602 \\
                       &                         & M & 128 & 4 & 32 & -- & 8.936 \\
                       &                         & L & 256 & 6 & 48 & -- & 78.692 \\
\midrule
\multirow{3}{*}{DPOT}  & \multirow{3}{*}{2D/3D} & S & 512  & 4 & 16 & -- & 7.914 \\
                       &                         & M & 1024 & 6 & 32 & -- & 36.198 \\
                       &                         & L & 2048 & 8 & 48 & -- & 173.005 \\
\midrule
\multirow{3}{*}{FactFormer} & \multirow{3}{*}{2D} & S & 32  & 2 & -- & 4 & 0.122 \\
                            &                     & M & 64  & 4 & -- & 6 & 1.232 \\
                            &                     & L & 128 & 6 & -- & 8 & 9.214 \\
\midrule
\multirow{3}{*}{Transolver} & \multirow{3}{*}{2D/3D} & S & 32  & 4 & -- & 4 & 0.105 \\
                            &                        & M & 64  & 6 & -- & 6 & 0.563 \\
                            &                        & L & 128 & 8 & -- & 8 & 3.090 \\
\midrule
\multirow{3}{*}{GNOT}  & \multirow{3}{*}{2D} & S & 64  & 4 & -- & 4 & 0.145 \\
                       &                     & M & 96  & 6 & -- & 6 & 0.472 \\
                       &                     & L & 256 & 8 & -- & 8 & 4.363 \\
\midrule
\multirow{3}{*}{ONO}   & \multirow{3}{*}{2D} & S & 64  & 4 & -- & 4 & 0.223 \\
                       &                     & M & 128 & 6 & -- & 6 & 1.231 \\
                       &                     & L & 256 & 8 & -- & 8 & 6.417 \\
\midrule
\multirow{3}{*}{DeepONet$^\P$} & \multirow{3}{*}{2D/3D} & S & 32  & 2 & 32  & -- & 0.016 \\                                                   &                        & M & 64  & 4 & 64  & -- & 0.080 \\
  &                        & L & 128 & 6 & 128 & -- & 0.382 \\
\bottomrule
\end{tabular}
\end{table}

\begin{table}[t!]
\centering
\small
\setlength{\tabcolsep}{6pt}
\caption{\textbf{Model-size presets for \emph{irregular-grid} cases}. 
Each family has three capacity presets (S/M/L) defined by \texttt{width}/\texttt{layers}/\texttt{modes}/\texttt{heads}; “Params (M)” are approximate. 
\emph{Notes:} $^\P$ For \textbf{DeepONet}, \texttt{modes} denotes \texttt{dotDim}. 
$^{\dagger}$ Shares the same presets as in the regular-grid table.}
\label{tab:unstructured_presets}
\begin{tabular}{lcccccc}
\toprule
Model & Size & width & layers & modes & heads & Params (M) \\
\midrule
\multirow{3}{*}{DeepONet$^\P$$^{\dagger}$}
 & S & 32  & 2 & 32  & -- & 0.026 \\
 & M & 64  & 4 & 64  & -- & 0.118 \\
 & L & 128 & 6 & 128 & -- & 0.532 \\
\midrule
\multirow{3}{*}{Transolver$^{\dagger}$}
 & S & 32  & 4 & -- & 4 & 0.036 \\
 & M & 64  & 6 & -- & 6 & 0.188 \\
 & L & 128 & 8 & -- & 8 & 0.980 \\
\midrule
\multirow{3}{*}{PointNet}
 & S & 32  & -- & -- & -- & 0.921 \\
 & M & 64  & -- & -- & -- & 3.672 \\
 & L & 96  & -- & -- & -- & 8.254 \\
\midrule
\multirow{3}{*}{LSM}
 & S & 32  & -- & 16 & 4 & 6.976 \\
 & M & 60  & -- & 24 & 6 & 33.532 \\
 & L & 96  & -- & 32 & 8 & 118.718 \\
\midrule
\multirow{3}{*}{MGN}
 & S & 32  & 3 & -- & -- & 0.046 \\
 & M & 64  & 4 & -- & -- & 0.224 \\
 & L & 96  & 5 & -- & -- & 0.601 \\
\midrule
\multirow{3}{*}{GraphUNet}
 & S & 32  & 3 & -- & -- & 0.061 \\
 & M & 64  & 4 & -- & -- & 0.895 \\
 & L & 96  & 5 & -- & -- & 7.911 \\
\midrule
\multirow{3}{*}{GraphSAGE}
 & S & 32  & 3 & -- & -- & 0.015 \\
 & M & 64  & 4 & -- & -- & 0.064 \\
 & L & 96  & 5 & -- & -- & 0.158 \\
\bottomrule
\end{tabular}
\end{table}

\subsection{Training schedule and model selection}
For each \emph{(case, model family, capacity tier)}, we perform a lightweight sweep over the maximum learning rate $\texttt{max\_lr}\!\in\!\{10^{-2},\,10^{-3},\,10^{-4}\}$ under a \texttt{OneCycleLR} schedule. Training is conducted either as single-step autoregressive forecasting or two-step rollout. \emph{Graph-based operators} are evaluated only under single-step rollout due to their significantly higher per-iteration cost, and \emph{DeepONet} is trained in a time-conditioned formulation without recurrent rollout. The mini-batch size is set to the largest configuration that fits on a single A800--80G GPU. For each \emph{(case, model)}, we select the configuration achieving the lowest \texttt{val\_error} on the held-out validation split.

\subsection{Case-wise hyperparameter selection and final performance}

We now report, for each benchmarked case, the outcome of the capacity-aligned sweeps and the final selected configuration. Table~\ref{tab:c3_overview} provides a per-case summary of the best-performing \emph{(family, tier)} under our unified protocol, together with the corresponding training knobs (\texttt{iters}, $\texttt{max\_lr}$), rollout setting, and the main test metrics. The primary objective $\overline{\mathcal{L}}$ (\texttt{test\_error}, evaluated in the encoder-normalized space) is used for selection. The reported inference time reflects end-to-end forward evaluation.

\begin{table}[t!]
\centering
\small
\setlength{\tabcolsep}{5.5pt}
\caption{\textbf{Per-case best configurations and summary metrics.}
For each benchmark case we list the selected model \emph{family} and \emph{capacity tier} (S/M/L), training setup (rollout type, $\texttt{max\_lr}$, batch size), the primary objective $\overline{\mathcal{L}}$ (\texttt{test\_error} in the encoder-normalized space) and the end-to-end inference time.}
\label{tab:c3_overview}
\resizebox{\linewidth}{!}{%
\begin{tabular}{l l l c c c c c c c c}
\toprule
Case & Family & Tier & Rollout & \texttt{max\_lr} & Batch &
train\_error & val\_error & test\_error &Corr.\\
\midrule
IgnitHIT        & FFNO      & M & 2-step & $1{\times}10^{-3}$ & 26 & 0.51785 & 1.8635 & 1.8739  & 0.97357 \\
EvolveJet           & FFNO     & L & 2-step & $1{\times}10^{-3}$ & 12 & 0.53881 & 2.4981 & 0.97919  & 0.95078  \\
ReactTGV        & FFNO      & M & 2-step & $1{\times}10^{-4}$ &  1 & 	1.4813 & 	1.8366 & 18.446  & 0.89590 \\
PropHIT   & DPOT      & M & 1-step & $1{\times}10^{-3}$ &  5 & 78.383 & 136.14	 & 49.856  & 0.41330 \\
PlanarDet  & FFNO       & M & 2-step & $1{\times}10^{-3}$ &  7 & 13.399 & 12.577 & 9.5377  & 0.99144 \\
SupCavityFlame  & LSM       & L & 2-step & $1{\times}10^{-3}$ &  7 & 27.487 & 	36.600 & 33.851  & 0.75046 \\
ObstacleDet  & DeepONet       & M & 1-step & $1{\times}10^{-3}$ &  4 & 84.087 & 60.735 & 	57.137  & 0.96460 \\
SymmCoaxFlame  & DeepONet       & L & 1-step & $1{\times}10^{-3}$ &  2 & 27.354 & 27.818 & 29.564  & 	0.79584 \\
MultiCoaxFlame  & DeepONet       & M & 1-step & $1{\times}10^{-3}$ &  2 & 27.646 & 25.961 & 31.841 & 0.70615 \\
PoolFire  & DeepONet   & S & 1-step & $1{\times}10^{-3}$ &  6 & 26.715 & 11.051 & 23.242  & 0.76791 \\
FacadeFire  & 	DeepONet       & L & 1-step & $1{\times}10^{-3}$ &  7 & 28.477 & 	35.691 & 24.957  & 0.94302 \\
\bottomrule
\end{tabular}}
\end{table}

Following the overview, we provide \emph{per-case sweep tables} that summarize, for each benchmarked case, the outcome of the capacity-aligned searches across model families. Each table reports, for every \emph{model family}, the single best configuration (``top-1 within family'') selected over all capacity tiers (S/M/L) and training knobs, together with the chosen \emph{tier} (capacity), the training setup, and the resulting \texttt{val\_error} and \texttt{test\_error}. Within each table, the row achieving the lowest \texttt{test\_error} is \textbf{boldfaced}; the best validation entry is also \underline{underlined} for reference. This layout makes it straightforward to (i) compare families at matched representational capacity and (ii) inspect sensitivity to the training configuration across operators.

Specifically, we report the capacity-aligned selections for each case in Tables~\ref{tab:2dplanardet}–\ref{tab:U3drocket}. For the 2D regular cases, the best configurations for \emph{PlanarDet}, \emph{EvolveJet}, and \emph{IgnitHIT} are summarized in Tables~\ref{tab:2dplanardet}–\ref{tab:2dHIT}, where spectral/Fourier operators consistently achieve the best performance. The 3D regular cases, \emph{PropHIT}, \emph{PoolFire}, and \emph{ReactTGV}, are reported in Tables~\ref{tab:3dHomoTurb}–\ref{tab:3dTGV}. Irregular-mesh scenarios, including \emph{ObstacleDet}, \emph{SupCavityFlame}, \emph{SymmCoaxFlame}, \emph{FacadeFire}, and \emph{MultiCoaxFlame}, are collected in Tables~\ref{tab:U2dAttenDet}–\ref{tab:U3drocket}, where DeepONet most frequently attains the best performance among graph and point-based competitors.

\begin{table}[t!]
\centering
\small
\setlength{\tabcolsep}{10pt}
\caption{\textbf{PlanarDet Selection.}
Among all candidates, FFNO achieved the lowest
\texttt{test\_error} under 2-step rollout with $\texttt{max\_lr}=10^{-3}$ and batch $=7$ }
\label{tab:2dplanardet}
\begin{tabular}{l c c c c c c c c c c}
\toprule
model & params & infertime &Corr.& train\_error & val\_error & test\_error \\
\midrule
CNext & 28.826 & 0.53689 & 0.89513 & 10.241 & 10.954 & 25.134 \\
CROP & 591.74 & 0.57267 & 0.95567 & 15.337 & 31.644 & 24.026 \\
DPOT & 41.071 & 0.79316 & 0.98468 & 8.7231 & 16.275 & 12.187 \\
DeepONet & 0.0175 & 0.12732 & 0.78803 & 51.233 & 34.156 & 31.197 \\
FFNO & 8.9367 & 0.36485 & 0.99144 & 13.399 & 12.577 & 9.5377 \\
FNO & 134.27 & 0.13129 & 0.89918 & 40.389 & 27.911 & 27.104 \\
FactFormer & 9.2147 & 1.1678 & 0.98904 & 11.364 & 15.974 & 11.836 \\
GNOT & 4.3633 & 0.32454 & 0.60514 & 149.77 & 73.912 & 98.392 \\
ONO & 0.22286 & 4.7004 & 0.53047 & 94.133 & 50.452 & 93.993 \\
Transolver & 0.56308 & 0.48899 & 0.83098 & 125.72 & 53.032 & 39.074 \\
UNO & 2093.6 & 1.4269 & 0.91812 & 28.632 & 34.318 & 34.412 \\
\bottomrule
\end{tabular}
\end{table}

\begin{table}[t!]
\centering
\footnotesize
\setlength{\tabcolsep}{10pt}
\caption{\textbf{EvolveJet Selection.}
Among all candidates, FFNO achieved the lowest
\texttt{test\_error} under 2-step rollout with $\texttt{max\_lr}=10^{-3}$ and batch $=12$ }
\label{tab:2dEvojet}
\begin{tabular}{l c c c c c c c c c c}
\toprule
model & params & infertime &Corr.& train\_error & val\_error & test\_error \\
\midrule
CNext & 28.856 & 0.46541 & 0.92558 & 1.0552 & 4.2489 & 2.5876 \\
CROP & 65.107 & 0.38328 & 0.84948 & 1.6608 & 3.6993 & 3.4078 \\
DPOT & 9.1718 & 0.51783 & 0.87707 & 2.0989 & 5.3195 & 3.3614 \\
DeepONet & 0.83127 & 0.23463 & 0.62205 & 7.8538 & 28.788 & 18.331 \\
FFNO & 78.703 & 0.27537 & 0.95078 & 0.53881 & 2.4981 & 0.97919 \\
FNO & 1208.1 & 0.11177 & 0.90397 & 0.95752 & 7.2116 & 2.9696 \\
FactFormer & 9.2213 & 0.87939 & 0.92762 & 0.56707 & 2.148208 & 1.5445 \\
GNOT & 4.3833 & 0.25356 & 0.85813 & 5.0318 & 12.567 & 7.4699 \\
ONO & 1.2481 & 2.8938 & 0.37446 & 75.289 & 88.589 & 70.962 \\
Transolver & 3.1003 & 0.46534 & 0.50621 & 71.499 & 77.775 & 56.099 \\
UNO & 235.05 & 0.77753 & 0.86088 & 4.1644 & 5.3127 & 4.0897 \\
\bottomrule
\end{tabular}
\end{table}

\begin{table}[t!]
\centering
\footnotesize
\setlength{\tabcolsep}{10pt}
\caption{\textbf{IgnitHIT Selection.}
Among all candidates, FFNO achieved the lowest
\texttt{test\_error} under 2-step rollout with $\texttt{max\_lr}=10^{-3}$ and batch $=26$ }
\label{tab:2dHIT}
\begin{tabular}{l l l c c c c c c c c}
\toprule
model & params & infertime &Corr.& train\_error & val\_error & test\_error \\
\midrule
CNext & 0.45857 & 0.43863 & 0.96087 & 2.4666 & 3.1737 & 2.7122 \\
CROP & 65.099 & 0.77965 & 0.94906 & 0.20289 & 5.2445 & 4.3738 \\
DPOT & 36.198 & 0.66759 & 0.94193 & 1.4654 & 6.1487 & 5.9008 \\
DeepONet & 0.38196 & 0.07768 & 0.68603 & 39.627 & 48.539 & 48.248 \\
FFNO & 8.9365 & 0.15041 & 0.97357 & 0.51785 & 1.8635 & 1.8739 \\
FNO & 8.4078 & 0.05901 & 0.92489 & 5.4989 & 8.4531 & 6.9115 \\
FactFormer & 9.2144 & 0.66352 & 0.95699 & 0.61568 & 2.7411 & 2.9051 \\
GNOT & 4.3625 & 0.20302 & 0.92712 & 8.1788 & 12.939 & 13.381 \\
ONO & 6.4165 & 5.2328 & 0.78758 & 38.208 & 46.319 & 39.596 \\
Transolver & 3.0899 & 0.53799 & 0.86846 & 14.861 & 23.251 & 13.925 \\
UNO & 378.08 & 0.67071 & 0.93944 & 1.0307 & 4.8818 & 5.5917 \\
\bottomrule
\end{tabular}
\end{table}

\begin{table}[t!]
\centering
\footnotesize
\setlength{\tabcolsep}{10pt}
\caption{\textbf{PropHIT Selection.}
Among all candidates, DPOT achieved the lowest
\texttt{test\_error} under 1-step rollout with $\texttt{max\_lr}=10^{-3}$ and batch $=5$ }
\label{tab:3dHomoTurb}
\begin{tabular}{l c c c c c c}
\toprule
model & params & infertime &Corr.& train\_error & val\_error & test\_error \\
\midrule
CNext      & 0.68037  & 0.38360  & 0.44553  & 85.627   & 137.11   & 64.965   \\
CROP       & 134.24   & 0.40130  & 0.25470  & 31.045   & 181.97   & 82.875   \\
DPOT       & 70.345   & 0.76649  & 0.41330  & 78.383   & 136.14   & 49.856   \\
DeepONet   & 0.017517 & 4.0054   & 0.12463  & 68.799   & 159.50   & 68.479   \\
FFNO       & 0.86395  & 0.19029  & 0.30569  & 68.196   & 150.14   & 69.034   \\
FNO        & 268.45   & 0.19550  & 0.24404  & 120.18   & 193.72   & 91.666   \\
Transolver & 0.28115  & 0.42644  & 0.39358  & 141.37   & 149.70   & 61.347   \\
UNO        & 73.668   & 0.99340  & 0.21644  & 47.453   & 159.42   & 83.604   \\
\bottomrule
\end{tabular}
\end{table}

\begin{table}[t!]
\centering
\footnotesize
\setlength{\tabcolsep}{10pt}
\caption{\textbf{PoolFire Selection.}
Among all candidates, DeepONet achieved the lowest
\texttt{test\_error} under 1-step rollout with $\texttt{max\_lr}=10^{-3}$ and batch $=6$ }
\label{tab:3dPoolfire}
\begin{tabular}{l c c c c c c}
\toprule
model & params & infertime &Corr.& train\_error & val\_error & test\_error \\
\midrule
CNext      & 35.564   & 0.47173  & 0.62052  & 23.913   & 14.599   & 29.449   \\
CROP       & 134.24   & 0.35685  & 0.79213  & 3.9315   & 14.232   & 27.962   \\
DPOT       & 68.243   & 0.42149  & 0.92327  & 12.318   & 12.007   & 25.935   \\
DeepONet   & 0.013161 & 2.6151   & 0.76791  & 26.715   & 11.051   & 23.242   \\
FFNO       & 13.130   & 0.23960  & 0.84658  & 14.566   & 11.817   & 24.691   \\
FNO        & 268.45   & 0.11157  & 0.88325  & 6.6713   & 12.729   & 25.324   \\
Transolver & 1.4491   & 0.45051  & 0.69612  & 33.178   & 11.919   & 29.182   \\
UNO        & 77.501   & 0.77081  & 0.92682  & 24.372   & 15.211   & 29.186   \\
\bottomrule
\end{tabular}
\end{table}

\begin{table}[t!]
\centering
\footnotesize
\setlength{\tabcolsep}{10pt}
\caption{\textbf{ReactTGV Selection.}
Among all candidates, FFNO achieved the lowest
\texttt{test\_error} under 2-step rollout with $\texttt{max\_lr}=10^{-3}$ and batch $=1$ }
\label{tab:3dTGV}
\begin{tabular}{l c c c c c c}
\toprule
model & params & infertime &Corr.& train\_error & val\_error & test\_error \\
\midrule
CNext      & 35.571   & 0.69053  & 0.87934  & 5.9302   & 4.1637   & 32.223   \\
CROP       & 134.24   & 0.47286  & 0.43991  & 18.558   & 25.671   & 65.092   \\
DPOT       & 240.19   & 0.86313  & 0.74489  & 2.8530   & 6.2201   & 38.761   \\
DeepONet   & 0.075979 & 7.0975   & 0.23663  & 29.313   & 31.192   & 65.633   \\
FFNO       & 13.131   & 0.37042  & 0.89590  & 1.4813   & 1.8366   & 18.446   \\
FNO        & 268.45   & 0.19001  & 0.28214  & 2.3010   & 24.082   & 69.544   \\
Transolver & 0.28096  & 0.68584  & 0.11141  & 100.03   & 80.459   & 148.39   \\
UNO        & 94.508   & 1.1188   & 0.37187  & 15.606   & 11.308   & 63.151   \\
\bottomrule
\end{tabular}
\end{table}

\begin{table}[t!]
\centering
\footnotesize
\setlength{\tabcolsep}{10pt}
\caption{\textbf{ObstacleDet Selection.}
Among all candidates, DeepONet achieved the lowest
\texttt{test\_error} under 1-step rollout with $\texttt{max\_lr}=10^{-3}$ and batch $=4$ }
\label{tab:U2dAttenDet}
\begin{tabular}{l c c c c c c}
\toprule
model & params & infertime &Corr.& train\_error & val\_error & test\_error \\
\midrule
DeepONet  & 0.050565 & 0.26593  & 0.96460  & 84.087   & 60.735   & 57.137   \\
GraphSAGE & 0.013413 & 0.22779  & 0.50415  & 208.21   & 219.62   & 203.10   \\
GraphUNet & 0.89082  & 22.953   & 0.93926  & 85.434   & 134.14   & 81.314   \\
LSM       & 33.528   & 0.97887  & 0.97671  & 88.534   & 54.179   & 60.311   \\
MGN       & 0.045061 & 0.18434  & 0.49494  & 193.54   & 237.55   & 185.01   \\
PointNet  & 0.91847  & 0.40075  & 0.87323  & 147.76   & 168.84   & 135.38   \\
Transolver& 0.18496  & 0.40264  & 0.90047  & 150.76   & 126.05   & 131.29   \\
\bottomrule
\end{tabular}
\end{table}

\begin{table}[t!]
\centering
\footnotesize
\setlength{\tabcolsep}{10pt}
\caption{\textbf{SupCavityFlame Selection.}
Among all candidates, LSM achieved the lowest
\texttt{test\_error} under 2-step rollout with $\texttt{max\_lr}=10^{-3}$ and batch $=7$ }
\label{tab:U2dCavity}
\begin{tabular}{l c c c c c c}
\toprule
model & params & infertime &Corr.& train\_error & val\_error & test\_error \\
\midrule
DeepONet  & 0.016396 & 0.21795  & 0.79271  & 26.823  & 40.111  & 34.423  \\
GraphSAGE & 0.014316 & 0.13373  & 0.71465  & 37.306  & 50.671  & 43.665  \\
GraphUNet & 0.060140 & 5.0960   & 0.73988  & 37.224  & 41.139  & 43.698  \\
LSM       & 118.71   & 0.57813  & 0.75046  & 27.487  & 36.600  & 33.851  \\
MGN       & 0.045516 & 0.084094 & 0.74568  & 32.840  & 45.797  & 38.098  \\
PointNet  & 0.91937  & 0.13939  & 0.76024  & 33.678  & 47.484  & 37.366  \\
Transolver& 0.18631  & 0.18574  & 0.75247  & 30.625  & 40.065  & 37.939  \\
\bottomrule
\end{tabular}
\end{table}

\begin{table}[t!]
\centering
\footnotesize
\setlength{\tabcolsep}{10pt}
\caption{\textbf{SymmCoaxFlame Selection.}
Among all candidates, DeepONet achieved the lowest
\texttt{test\_error} under 1-step rollout with $\texttt{max\_lr}=10^{-3}$ and batch $=2$ }
\label{tab:U2dRocket}
\begin{tabular}{l c c c c c c}
\toprule
model & params & infertime &Corr.& train\_error & val\_error & test\_error \\
\midrule
DeepONet   & 0.53173  & 0.32102  & 0.79584  & 27.354  & 27.818  & 29.564  \\
GraphSAGE  & 0.15756  & 0.36232  & 0.23413  & 123.84  & 108.81  & 262.30  \\
GraphUNet  & 0.89493  & 32.650   & 0.74323  & 51.304  & 44.092  & 108.75  \\
LSM        & 118.72   & 0.79837  & 0.71576  & 29.637  & 27.304  & 35.544  \\
MGN        & 0.60137  & 0.38023  & 0.47392  & 118.88  & 122.95  & 241.24  \\
PointNet   & 3.6720   & 0.77203  & 0.61491  & 88.038  & 68.667  & 85.111  \\
Transolver & 0.18805  & 0.30262  & 0.68579  & 46.059  & 38.529  & 47.192  \\
\bottomrule
\end{tabular}
\end{table}

\begin{table}[t!]
\centering
\footnotesize
\setlength{\tabcolsep}{10pt}
\caption{\textbf{PoolFire Selection.}
Among all candidates, DeepONet achieved the lowest
\texttt{test\_error} under 1-step rollout with $\texttt{max\_lr}=10^{-3}$ and batch $=7$ }
\label{tab:U3dPopfire}
\begin{tabular}{l c c c c c c}
\toprule
model & params & infertime &Corr.& train\_error & val\_error & test\_error \\
\midrule
DeepONet & 0.33217 & 0.21195 & 0.94302 & 28.477 & 35.691 & 24.957 \\
GraphSAGE & 0.06081 & 0.13408 & 0.90851 & 42.819 & 46.395 & 34.703 \\
GraphUNet & 7.9067 & 13.201 & 0.93435 & 31.615 & 38.452 & 28.142 \\
MGN & 0.22253 & 0.10148& 0.90434 & 37.788 & 43.608 & 31.507 \\
PointNet & 3.6691 & 0.20629 & 0.92898 & 34.827 & 40.202 & 29.798 \\
Transolver & 0.035193 & 0.15695 & 0.94152 & 31.791 & 36.741 & 26.387 \\
\bottomrule
\end{tabular}
\end{table}

\begin{table}[t!]
\centering
\footnotesize
\setlength{\tabcolsep}{10pt}
\caption{\textbf{MultiCoaxFlame Selection.}
Among all candidates, DeepONet achieved the lowest
\texttt{test\_error} under 1-step rollout with $\texttt{max\_lr}=10^{-3}$ and batch $=3$ }
\label{tab:U3drocket}
\begin{tabular}{l c c c c c c}
\toprule
model & params & infertime &Corr.& train\_error & val\_error & test\_error \\
\midrule
DeepONet & 0.12245 & 0.25198 & 0.70615& 27.646& 25.691& 31.841\\
GraphSAGE & 0.01561 & 0.18663 & 0.02383& 121.46& 119.33& 136.89\\
GraphUNet & 0.89532 & 14.353 & 0.45859& 99.223& 98.823& 107.02\\
MGN & 0.04619 & 0.12844 & 0.35686& 99.316& 97.224& 111.17\\
PointNet & 0.92073 & 0.20587& 0.35798& 77.516& 73.888& 86.455\\
Transolver & 0.18837 & 0.26799& 0.67749& 62.787& 60.162& 68.507\\
\bottomrule
\end{tabular}
\end{table}

\section{Supplementary Videos}

\noindent\textbf{Supplementary Video 1 | \emph{PropHIT}.}  
Time-resolved 3D visualization of the \emph{PropHIT} case. The video shows volume and isosurface renderings of vorticity magnitude and heat release rate, highlighting the interaction between the turbulent flow and the flame surface. Two planar slices further display 2D projections of the heat release rate and velocity fields.

\medskip
\noindent\textbf{Supplementary Video 2 | \emph{MultiCoaxFlame}.}  
Volume rendering of the temperature field for the \emph{MultiCoaxFlame} case, capturing the ignition, development and downstream evolution of the reacting coaxial jets in three dimensions.

\medskip
\noindent\textbf{Supplementary Video 3 | \emph{FacadeFire}.}  
Visualization of the \emph{FacadeFire} case, showing the growth of the pool fire, attachment and spread along the facade, and the formation of large-scale plumes and smoke.

\end{document}